\documentclass{amsart}
\usepackage{amsmath}

\usepackage{graphicx}
\usepackage[table]{xcolor}

\usepackage{cancel}	
\usepackage{algorithm}
\usepackage{algorithmic}
\usepackage{enumitem}
\usepackage{tikz}
\usepackage{booktabs}
\usepackage{arydshln}
\usepackage{threeparttable}
\usepackage{colortbl}
\usepackage{adjustbox}
\usetikzlibrary{calc}
\usetikzlibrary{positioning}
\usetikzlibrary{decorations.pathreplacing,calc}

\newcommand{\vect}[1]{\boldsymbol{#1}}

\newcommand{\cgr}{\cellcolor{gray!20}}
\renewcommand{\phi}{\varphi}
\newcommand{\bvarphi}{{\boldsymbol{\varphi}}}
\usepackage{multirow}

\theoremstyle{definition}

\theoremstyle{remark}

\usepackage{url}
\usepackage{natbib}

\begin{document}
\bibliographystyle{mystyle}

\title{FineMorphs: affine-diffeomorphic sequences for regression}

\author{Michele Lohr}
\address{Department of Applied Mathematics and Statistics, The Johns Hopkins University, Baltimore, Maryland 21218}
%\curraddr{Department of Mathematics and Statistics,
%Case Western Reserve University, Cleveland, Ohio 43403}
\email{mlohr@cis.jhu.edu}

\author{Laurent Younes}
\address{Department of Applied Mathematics and Statistics, Center for Imaging Science, Mathematical Institute for Data Science and Kavli Neuroscience Discovery Institute, The Johns Hopkins University, Baltimore, Maryland 21218}
{}
\email{younes@cis.jhu.edu}

%%    General info
%\subjclass[2000]{Primary 54C40, 14E20; Secondary 46E25, 20C20}
%
%\date{January 1, 2001 and, in revised form, June 22, 2001.}


\begin{abstract}
A multivariate regression model of affine and diffeomorphic transformation sequences---FineMorphs---is presented. Leveraging concepts from shape analysis, model states are optimally ``reshaped” by diffeomorphisms generated by smooth %complete 
vector fields during learning.  Affine transformations and vector fields are optimized within an optimal control setting, and the model can naturally reduce (or increase) dimensionality and adapt to large datasets via suboptimal vector fields.
An existence proof of solution and necessary conditions for optimality for the model are derived.  
% Additionally, the model framework naturally adapts to address large datasets with high dimensionality.   
% An existence proof of solution for the resulting variational problem
% , optimizing over the affine transformations and vector fields, we provide an existence proof of solution 
% and necessary conditions for optimality. 
Experimental results on real datasets from the UCI repository are presented, with favorable results  in comparison with state-of-the-art in the literature and densely-connected neural networks in TensorFlow.
%, and a robustness to out-of-distribution experiments,  
%This paper is a sample prepared to illustrate the use of the American
%Mathematical Society's \LaTeX{} document class \texttt{amsart} and
%publication-specific variants of that class for AMS-\LaTeX{} version 2.  
\end{abstract}

\maketitle

\section{Introduction}
We present FineMorphs---an affine-diffeomorphic sequence model for multivariate regression. 
Our approach combines arbitrary sequences of affine and diffeomorphic transformations with a training algorithm using concepts from optimal control.
%, the mathematical framework underpinning backpropagation in neural networks (NNs) \citep{lecun-88}.  
Predictors, estimated responses, and states in between are transformed or ``reshaped" via diffeomorphisms of their respective ambient spaces, in an optimal way to facilitate learning.   

Recall that diffeomorphisms of an open subset $M$ of a Euclidean space $\mathbb R^d$ (where we will typically take $M= \mathbb R^d$) are one-to-one, invertible, $C^1$ transformations mapping $M$ onto itself that have $C^1$ inverse. (If $C^1$ is replaced by $C^0$, one speaks of homeomorphisms.) Because diffeomorphisms form a group, arbitrary large deformations can be generated via the composition of many small ones, making them natural objects to utilize within a feed forward setting. In the limit of infinite compositions of transformations that differ infinitesimally from the identity, one finds the classical representation of diffeomorphisms 
as flows associated to ordinary differential equations (ODEs). 
 %Diffeomorphisms  are generated by ODEs associated with
 % In shape analysis, 
 % the ODEs are associated with time-dependent vector fields in reproducing kernel Hilbert spaces (RKHSs).   
%  Through these roots in shape analysis and kernels, this approach demonstrates similarities in architecture and low- to high-level feature extraction 
%  performance 
%  with traditional deep convolutional neural networks (CNNs), but with the advantages of 
% smooth 
% invertibility and ``infinite depth."

Several papers have recently explored the possibility of using homeomorphic or diffeomorphic transformations within feed-forward machine learning models. Discrete invertible versions of the ResNet architecture \citep{he2016deep} were proposed as ``normalizing flows'' in \citet{rezende2015variational} (see \citet{kobyzev2020normalizing} for a recent review), and extended to a time-continuous form in \citet{chen2019neural,rousseau2019residual,dupont2019augmented}. Continuous-time optimal control as a learning principle was proposed in \cite{weinan2017proposal,owhadi2020ideas,ganaba2021deep}.  Applications of deep residual neural networks (NNs) to the large deformation diffeomorphic metric mapping (LDDMM) framework of shape analysis have recently been explored \citep{arguillere:hal-03413643,wu2023neurepdiff} as well as sub-Riemannian landmark matching as time-continuous NNs  \citep{jansson2022subriemannian}.
% Reference by Kevrekidis \citep{doi:10.1080/00986449208936084}.

A direct formalization of the diffeomorphic learning approach was proposed in \citet{younes2019diffeomorphic}. While most flow-based learning approaches build dynamical systems that are adapted to NN implementations, diffeomorphic learning is presented as a non-parametric penalized regression problem, parametrized by a diffeomorphism of the data space. The penalty is specified as a  Riemannian metric on the diffeomorphism group, in a framework directly inspired from shape analysis \citep{younes2010shapes}. When applied to finite training data, the method reduces to a finite-, albeit large-, dimensional problem involving reproducing kernels (see Section \ref{sec:trick}). Shape analysis methods were also introduced for dimensionality reduction in \citet{walder2009diffeomorphic}. Similar models were used combined with a shooting formulation for the comparison of geodesics in \citet{vialard2020shooting}.  

In this paper, we provide three extensions to the approach in \citet{younes2019diffeomorphic}, with existence proof of solution 
and derivation of necessary conditions for optimality.  First, we extend the single diffeomorphic layer sequence approach to arbitrary affine-diffeomorphic sequences, providing a natural framework for automated data scaling as well as dimensionality reduction.
% through the additional affine layers.  
Second, we extend the model from classification to regression.  In particular, we consider vector regression predictors of the form 
\begin{equation}
\label{eq:regression.intro}
    x\in \mathbb{R}^{d_X}\mapsto A_m\circ\phi_m\circ A_{m-1}\circ\cdots\circ\phi_1\circ A_0(x)\in\mathbb{R}^{d_Y},
    %\nonumber
\end{equation}
where $A_q,$ $q=0,\dots,m,$ are affine transformations from $\mathbb{R}^{{d}_q}$ to $\mathbb{R}^{{d}_{q+1}},$ and $\phi_q,$ $q=1,\dots,m,$ are diffeomorphisms on $\mathbb{R}^{d_q}.$ 
In this model, a $d_Y$-dimensional output variable is predicted by the transformation of a $d_X$-dimensional input through an arbitrary number and order of arbitrary affine and diffeomorphic transformations.  
% We provide a proof of the existence of a solution to the resulting variational problem and derive necessary conditions for optimality.
Third, we extend the approach 
% to sub-Riemannian 
to include a more general sub-optimal vector fields setting 
% \citep{younes:hal-02386227} \citep{holm2004reduced} 
to train diffeomorphisms on a subset of the training data, providing a natural framework for dataset (and model) reduction in the case of very large datasets.  
% Necessary conditions for optimality are also provided for this setting.
% to allow for data subset training in the case of very large datasets.    
% Additionally, we extend the model framework to include sub-optimal vector fields to allow for data subset training in the case of very large datasets.
Combined with a GPU implementation, this allows for experiments on datasets beyond smaller-sized, simulated datasets to real-world data with larger, more realistic dimensions and sizes.  

We test our diffeomorphic regression models on real datasets from the UCI repository 
\citep{Dua:2019}, 
% \citep{dua2019uci},
with favorable results in comparison with  the literature and with densely-connected NNs (DNNs) in TensorFlow \citep{45166}.    
% We note a potential parallel between our models and CNNs in 
We note improved performance with  multiple sequential diffeomorphic modules with decreasing kernel sizes as well as a robustness of our models to ``out-of-distribution" testing.  
For the largest dataset in our experiments, in both dimensionality and number size, our model reduces dimensionality through affine transformations and reduces number through sub-optimal vector fields, with a significant decrease in run-time and good predictive results in comparison with the literature and DNNs.  
% provides a natural framework for dimensionality reduction through the affine transformations and for dataset (and model) reduction through the sub-optimal vector fields approach, with a significant decrease in run-time and with good predictive results in comparison with the literature and DNNs.

\section*{Notation}
For our multivariate regression setting, $X : \Omega \to \mathbb{R}^{d_X}$ is the predictor variable and $Y: \Omega \to \mathbb{R}^{d_Y}$ is the response. The training dataset is denoted 
\[
{\mathcal{T}_0 = (x_1,y_1,\dots,x_N,y_N).}
\]
The training predictors are $\vect{x} = (x_1,\dots,x_N) \in (\mathbb{R}^{d_X})^N$ and training responses are $\vect{y} = (y_1,\dots,y_N)\in(\mathbb{R}^{d_Y})^N$.  
We define the operator $\iota_{j}: \mathbb{R}^{d} \to \mathbb{R}^{d+j}$, where $\iota_j(x)$  appends   $j$ zero coordinates to $x$,  and the operator $\pi_j: \mathbb{R}^{d} \to \mathbb{R}^{d-j}$, where $\pi_j(x)$ removes the last $j$ coordinates from $x.$ 
For matrix notation, if $k,l$ are two integers, $\mathcal M_{k,l}(\mathbb{R})$ is the space of all $k\times l$ real matrices, reducing to $\mathcal M_{k}(\mathbb{R})$ for square $k\times k$ real matrices.  The $d\times d$ identity matrix is denoted $\mathrm{I}_{d}.$ When applied to vectors and matrices, the norm $\|\cdot\|$ is the Euclidean and Frobenius norm, respectively.  For time-dependent vector fields
\begin{align}
v: \mathbb{R}\times\mathbb{R}^d&\rightarrow \mathbb{R}^d \nonumber \\
                         (t,x)&\mapsto v(t,x) \nonumber
\end{align}
we will denote by $v$ the mapping 
${t\mapsto v(t),}$ where $v(t)$ is the time-indexed vector field $x\mapsto v(t,x).$  In particular, the time-dependent vector fields $v$ in the Bochner spaces $L^2(I,V)$ will represent the mapping   
\[
t\in I\mapsto v(t)\in V,
\]
where $V$ is a Hilbert space.

\section{Model}\label{sec:model}
We consider the following regression model approximating $Y$ by $f(X)$, in which we complete \eqref{eq:regression.intro} by possibly padding zeros in input and removing coordinates in output,
\begin{equation}
\label{eq:regression.notation0}
    f: x\in \mathbb{R}^{d_X}\mapsto\pi_r\left(A_m\circ\phi_m\circ A_{m-1}\circ\cdots\circ\phi_1\circ A_0(\iota_{s}(x))\right)\in\mathbb{R}^{d_Y}.
    \nonumber
\end{equation}
Here, $\iota_s$ pads the input with $s$ zeros so that ${d_0 = d_X + s,}$ and $\pi_r$ removes the last $r$ coordinates from the model output so that $d_{m+1}=d_Y + r$.  
Advantages of adding ``dummy" dimensions are discussed in Section \ref{sec:addeddims}.  In contrast to the single affine layered approach of standard linear regression, this model alternates $m+1$ affine transformations and $m$ diffeomorphic layers,
denoted as A and D modules, respectively, 
starting and ending with affine modules.  For affine modules A$_q,$ ${q=0,\dots,m,}$ the corresponding affine transformations are  
\[
{A}_q : x\in\mathbb{R}^{{d}_q} \mapsto M_q x + b_q \in \mathbb{R}^{{d}_{q+1}},
\]
where ${M_q\in\mathcal M_{d_{q+1},d_{q}}(\mathbb{R}),}$ ${b_q\in\mathbb{R}^{d_{q+1}}.}$
For diffeomorphic modules D$_q,$ $q=1,\dots,m,$ the corresponding diffeomorphisms and their domains are $\phi_q$ and $\mathbb{R}^{d_q},$ respectively. 

\begin{figure}[h!]
	\begin{center}
		\begin{tikzpicture}[
		auto,
		blockD/.style={
			rectangle,
			draw=black,
			thick,
			fill=green!20,
			text width=1.em,
			align=center,
			rounded corners,
			minimum height=4em
		},
		blockA/.style={
			rectangle,
			draw=black,
			thick,
			fill=blue!20,
			text width=1.em,
			align=center,
			rounded corners,
			minimum height=4em
		},
		line/.style={
			draw,thick,
			-latex',
			shorten >=2pt
		},
		cloud/.style={
			draw=red,
			thick,
			ellipse,
			fill=red!20,
			minimum height=1em
		}
		]
		\draw (1.75,0) node (A) {$X$};
		\path 
		(3,0) node[blockA] (B) {A}
		(4.25,0) node (C) {$Y$} ;
		\draw[thick,->] (A.east) -- (B.west) node[midway, above] {};
		\draw[thick,->] (B.east) -- (C.west) node[midway, above] {};
		
		\draw (1.25,-2) node (A) {$X$};
		\path 
		(2.5,-2) node[blockD] (B) {D$_1$}
		(3.5,-2) node[blockA] (C) {A$_1$}
		(4.75,-2) node (D) {$Y$} ;
		\draw[thick,->] (A.east) -- (B.west) node[midway, above] {$\iota_s$};
		\draw[thick,->] (B.east) -- (C.west) node[midway, above] {};
		\draw[thick,->] (C.east) -- (D.west) node[midway, above] {$\pi_r$};
		
		\draw (1.25,-4) node (A) {$X$};
		\path 
		(2.5,-4) node[blockA] (B) {A$_0$}
		(3.5,-4) node[blockD] (C) {D$_1$}
		(4.75,-4) node (D) {$Y$} ;
		\draw[thick,->] (A.east) -- (B.west) node[midway, above] {$\iota_s$};
		\draw[thick,->] (B.east) -- (C.west) node[midway, above] {};
		\draw[thick,->] (C.east) -- (D.west) node[midway, above] {$\pi_r$};
		
		\draw (0.25,-6) node (A) {$X$};
		\path 
		(1.5,-6) node[blockD] (B) {D$_1$}
		(2.5,-6) node[blockA] (C) {A$_1$}
		(3.5,-6) node[blockD] (D) {D$_2$}
		(4.5,-6) node[blockA] (E) {A$_2$}
		(5.75,-6) node (F) {$Y$} ;
		\draw[thick,->] (A.east) -- (B.west) node[midway, above] {$\iota_s$};
		\draw[thick,->] (B.east) -- (C.west) node[midway, above] {};
		\draw[thick,->] (C.east) -- (D.west) node[midway, above] {};
		\draw[thick,->] (D.east) -- (E.west) node[midway, above] {};
		\draw[thick,->] (E.east) -- (F.west) node[midway, above] {$\pi_r$};
		
		\draw (-0.25,-8) node (A) {$X$};
		\path 
		(1,-8) node[blockA] (B) {A$_0$}
		(2,-8) node[blockD] (C) {D$_1$}
		(3,-8) node[blockD] (D) {D$_2$}
		(4,-8) node[blockA] (E) {A$_2$}
		(5,-8) node[blockA] (F) {A$_3$}
		(6.25,-8) node (G) {$Y$} ;
		\draw[thick,->] (A.east) -- (B.west) node[midway, above] {$\iota_s$};
		\draw[thick,->] (B.east) -- (C.west) node[midway, above] {};
		\draw[thick,->] (C.east) -- (D.west) node[midway, above] {};
		\draw[thick,->] (D.east) -- (E.west) node[midway, above] {};
		\draw[thick,->] (E.east) -- (F.west) node[midway, above] {};
		\draw[thick,->] (F.east) -- (G.west) node[midway, above] {$\pi_r$};
		\end{tikzpicture}
	\end{center}
	\caption{\label{fig:models0} Standard linear regression (top) followed by four example transformation sequences that can be operated by the FineMorphs model, with naming convention (from top to bottom): A, DA, AD, DADA or (DA)$^2,$ and ADDAA or AD$^2$A$^2$ (A: affine module; D: diffeomorphic module).  Identity modules are omitted.}
\end{figure}
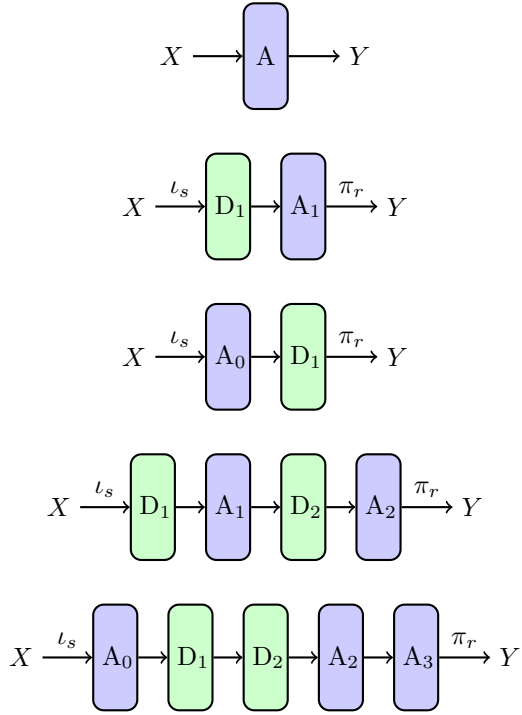

The values of $n$ and $r$, and the internal dimensions $d_1, \ldots, d_m$ are parts of the design of the model, i.e., they are user-specified. Given them, the dimensions of the linear operators are uniquely determined, and so are the spaces on which the diffeomorphisms operate. Any module in a sequence with identical input and output dimensions can be set to  the identity map, $\mathrm{id}$,  which allows for simple definitions of submodels from an initial sequence of modules (obviously, one wants to keep at least one A module and at least one D module free to optimize by the system).  
%The integers $n,r\ge0$ are user-specified and used to determine the input dimension of the first module $d_0$ and the output dimension of the last module $d_{m+1},$ respectively.  The user-specified dimensions $d_1,\dots,d_{m}$ of the inner modules are then modified accordingly.  
The flexibility of assigning module dimensions as well as arbitrary modules to the identity generalizes our model from a simple and fixed alternating sequence to an arbitrary sequence of arbitrary affine and diffeomorphic transformations.  In this setting, affine modules can provide not only useful data scaling prior to diffeomorphic transforms but also a natural approach to dimensionality reduction or increase.  
In the following, the naming convention for sequences includes only non-identity modules, e.g., the sequence of modules A$_0,$ D$_1,$ A$_1,$ D$_2,$ A$_2,$ D$_3,$ and A$_3,$ where A$_1$ and D$_3$ are identities, is denoted ADDAA. 
% For sequence names containing repetitious module or module subsequence elements, we further adopt a simplified notation superscripting the repetition, e.g., 
For sequence names containing repetitive module or module subsequence elements, we further adopt a simplified notation superscripting the repetition, e.g., ADDAA can be expressed as AD$^2$A$^2,$ and sequence ADAD$\cdots$A with x sequential AD module pairs before the final A can be denoted as (AD)${\mbox{\textsuperscript{x}}}$A. Several sequence examples are illustrated in Figure \ref{fig:models0}, including the smallest possible sequences that can be represented in our model, DA and AD.

\section{Objective Function}\label{sec:objective}

Learning is implemented by minimizing the objective function 
\begin{equation}
\label{eq:obj1}
\sum_{q=1}^{m} d_{V_q}(\mathrm{id},\phi_q)^2 + \lambda \sum_{q=0}^{m}U_q({A}_q) + \frac{1}{\sigma^2}\sum_{k=1}^N \Gamma_k(\pi_r({A}_m\circ\phi_m\circ{A}_{m-1}\circ\cdots\circ\phi_1\circ{A}_0(\iota_s(x_k)))) \nonumber
\end{equation}
over $\phi_1,\dots,\phi_m,{A}_0,\dots,{A}_m.$ The objective function
combines an optimal deformation cost $d_{V_q},$ an affine cost $U_q,$ and a standard loss function or endpoint cost $\Gamma_k.$ In our setting, $d_{V_q}$ is a Riemannian distance in a group of diffeomorphisms of $\mathbb{R}^{d_q}$ described in Section \ref{sec:distances}, $U_q$ is a ridge regularization function
\begin{equation}
U_q({A}) =  \left\| M \right\|^2 = \mathrm{trace}(M^TM) \nonumber
\end{equation} 
for affine transformations ${A}:\mathbb{R}^{d_q}\rightarrow\mathbb{R}^{d_{q+1}}$ of the form
${A}(x) = M x + b,$ ${M\in\mathcal M_{d_{q+1},d_{q}}(\mathbb{R}),}$ ${b\in\mathbb{R}^{d_{q+1}},}$ and $\Gamma_k$ is a squared error loss function 
\begin{equation}
\Gamma_k(\cdot) = \left\| y_k - (\cdot)\right\| ^2 \nonumber
\end{equation} 
for comparison of experimental responses with model predictions.

\section{Distance over Diffeomorphisms}\label{sec:distances}

Spaces of diffeomorphisms are defined as follows. Let $\mathrm{\textbf{B}}_p=C_0^p(\mathbb{R}^d,\mathbb{R}^d)$ denote the space of $C^p$ vector fields on $\mathbb{R}^d$ that tend to zero (together with their first $p$ derivatives) at infinity.  This is a Banach space for the norm
\begin{equation}
\left\| f\right\|_{p,\infty} = \max_{0\le k\le p} \left\| d^kf\right\|_{\infty}, \nonumber
\end{equation}
where $\left\|\,\cdot\,\right\|_{\infty}$ denotes the usual supremum norm.
Let $V$ denote a Hilbert space of vector fields on $\mathbb{R}^d,$
continuously embedded in $\mathrm{\textbf{B}}_p$ for some $p\ge 1,$ 
% A Hilbert space $V$ of vector fields ${v: \mathbb{R}^{d} \to \mathbb{R}^{d}}$ is admissible if it is continuously embedded in $B_p$ for some $p\ge 1,$ 
so that there exists a $C>0$ such that
\begin{equation}
\left\| f \right\|_{p,\infty} \le C\left\|f\right\|_V,\nonumber%\quad \forall f\in V\nonumber
\end{equation}
for all $f\in V,$ where $\left\|\cdot\right\|_V$ is the Hilbert norm on $V$ with inner product $\langle\cdot,\cdot{\rangle}_V.$  

% by time-dependent vector fields ${v\in L^2([0,1],V)},$ where $v(t)\in V,$ $t\in [0,1],$ with finite kinetic energy 
% \begin{equation}
% \int_0^1 \|v(t)\|_V^2 dt < \infty.\nonumber
% \end{equation}

Diffeomorphisms can be generated as flows of ODEs associated with time-dependent elements of $V.$ 
% Let ${v\in L^2([0,1],V)}$ if and only if 
% ${v : t\in[0,1] \mapsto v(t)\in V}$
% % $v(t)\in V$ for $t\in [0,1]$ and 
% with finite kinetic energy 
% \begin{equation}
% \int_0^1 \|v(t)\|_V^2 dt < \infty.\nonumber
% \end{equation}
Let $\mathcal{H}$ denote the Hilbert space %of time-dependent vector fields 
${L^2([0,1],V)}$ of 
time-dependent vector fields, so that % ${v=(v(t),t\in[0,1])}$
% with norm $\|\cdot\|_{\mathcal{H}}$ defined by
% \[
% \|v\|_{\mathcal{H}}^2 = \int_0^1 \|v(t,\cdot)\|_V^2 dt%\quad \forall v\in V
% \]
% and associated inner product $\langle\cdot,\cdot{\rangle}_{\mathcal{H}}$.
% and let ${v\in \mathcal{H}}$ if and only if
${v\in \mathcal{H}}$, if and only if
${v(t)\in V}$ for $t\in [0,1]$, $v$ is measurable  
% ${v : t\in[0,1] \mapsto v(t)\in V}$
% $v(t)\in V$ for $t\in [0,1]$ and 
and %with finite total kinetic energy 
\begin{equation}
\|v\|_{\mathcal{H}}^2 = \int_0^1 \|v(t)\|_V^2 dt < \infty,\nonumber
\end{equation}
where $\|\cdot\|_{\mathcal{H}}$ denotes the norm on $\mathcal{H}$ with inner product $\langle\cdot,\cdot{\rangle}_{\mathcal{H}}$.
 % where $\mathcal{H}$ is equivalently the set of time-dependent vector fields $v=(v(t),t\in[0,1])$ such that 
 % ${v : t\in[0,1] \mapsto v(t)\in V}$ 
Then the ODE 
\[
\partial_t y(t) = v(t)(y(t)) 
\]
has a unique solution over $[0,1]$
given any initial condition $y(0)=x.$
% given any initial condition $y_0\in\mathbb{R}^d.$ 
The flow of the ODE is the function
\[
\bvarphi_v: (t,x)\mapsto y(t),
\]
where $y(t)$ is the solution starting at $x,$ after $t$ units of time. 
This function is the unique flow of $\mathbb{R}^d$-diffeomorphisms  satisfying the dynamical system 
\begin{align}
\partial_t\bvarphi_v(t,x)&=v(t)(\bvarphi_v(t,x)) \nonumber \\
\bvarphi_v(0,x)&=x  \nonumber
\end{align}
over $t\in[0,1].$
We will often 
% use the notation ${\bvarphi_v(t)(\cdot) = \bvarphi_v(t,\cdot)},$  
write  $\bvarphi_v(t)$ for the time-indexed function $x\mapsto\bvarphi_v(t,x)$  satisfying 
\begin{align}
\partial_t\bvarphi_v(t)&=v(t)\circ\bvarphi_v(t),\quad t\in[0,1] \nonumber \\
\bvarphi_v(0)&=\mbox{id}.  \nonumber
\end{align}
The set of diffeomorphisms that can be generated in such a way forms a group denoted $\mathrm{Diff}_V,$ such that a flow path associated with some $v\in V$ is a curve on $\mathrm{Diff}_V$.  
% between given endpoints at $t=0$ and $t=1,$ is a flow path associated with some ${v\in V}$
% between points $\bvarphi_v(0)=\mathrm{id}$ and $\bvarphi_v(1)=\psi\in\mathrm{Diff}_V$ 
% is a flow path defined by some ${v\in V},$ 
Let $\frac{1}{2}\| v(t) \|_V^2$ denote 
the kinetic energy associated with the flow's velocity at time $t$ along this curve.
% the total kinetic energy of our system at time $t$ along this curve.  
Given $\psi\in\mathrm{Diff}_V$, we define the optimal deformation cost from $\mathrm{id}$ to $\psi$ as the minimal %total 
kinetic energy 
% along a curve 
among all curves  
between $\mathrm{id}$ and $\psi$ on $\mathrm{Diff}_V$, i.e., the minimum of $\int_0^1\| v(t)\|_V^2dt$
over all 
${v\in \mathcal{H}}$ 
such that $\bvarphi_v(1)=\psi.$  
% This results from the equivalence of minimizing the curve length 
% \[
% \int_0^1 \| \partial_t \bvarphi_v(t) \|_{\bvarphi_v(t)}dt
% \]
% between $\bvarphi(0)=\mathrm{id}$ and $\bvarphi(1)=\psi$ over all $v\in\mathcal{H}$
% with minimizing 
% \[
% \int_0^1 \| \partial_t \bvarphi_v(t) \|_{\bvarphi_v(t)}^2dt
% \]
% over all such curves, and rephrasing curve length in terms of flow velocity
% \[
% \int_0^1 \| v(t) \|_{V}dt.
% \]
A right-invariant distance $d_V(\cdot,\cdot)$ can then be defined on $\mathrm{Diff}_V$.   
Given $\psi,\psi'\in\mathrm{Diff}_V,$ ${d_V(\psi,\psi') = d_V(\mathrm{id}, \psi'\circ \psi^{-1})}$ and
\[
d_V(\mathrm{id}, \psi)^2 = \min_{v\in \mathcal{H}}\left\{\int_0^1 \|v(t)\|_V^2 dt: \bvarphi_v(1)=\psi\right\}\,.
\]
% as the 
% minimal path between two points on $\mathrm{Diff}_V.$ 

% sum of the 
%   minimum total kinetic energy, 
% over all 
% % ${v\in L^2([0,1],V)},$ 
% ${v\in \mathcal{H}},$ 
% required to generate a diffeomorphism flow between two points on $\mathrm{Diff}_V.$  

% To more explicitly express 
% the time-indexed elements of $V$ in ${v\in\mathcal{H},}$ 
% we introduce the small abuse of notation ${v(t)(\cdot) = v(t,\cdot)},$  where $\mathcal{H}$ is equivalently the set of time-dependent vector fields $v=(v(t),t\in[0,1])$ such that 
%  ${v : t\in[0,1] \mapsto v(t)\in V}$ and 
% \begin{equation}
% \|v\|_{\mathcal{H}}^2 = \int_0^1 \|v(t)\|_V^2 dt < \infty.\nonumber
% \end{equation}  
% Similarly, we will often use the notation ${\bvarphi_v(t)(\cdot) = \bvarphi_v(t,\cdot)},$  writing  $\bvarphi_v(t)$ for the time-indexed function $x\mapsto\bvarphi_v(t,x)$  satisfying 
% \begin{align}
% \partial_t\bvarphi_v(t)&=v(t)\circ\bvarphi_v(t),\quad t\in[0,1] \nonumber \\
% \bvarphi_v(0)&=\mbox{id}.  \nonumber
% \end{align}

In our setting of $m$ distinct D modules, we assume for each D$_q$ module the corresponding Hilbert space $V_q$ of vector fields 
on $\mathbb{R}^{d_q}$
and Hilbert space ${L^2([0,1],V_q)}$ denoted $\mathcal{H}_q,$
and let the time-dependent vector fields ${v_q\in \mathcal{H}_q}$ 
generate the corresponding $\mathrm{Diff}_{V_q}$ space of diffeomorphisms.  
Our optimal deformation cost can then be expressed in terms of the vector fields as
\[
\sum_{q=1}^m d_{V_q}(\mathrm{id}, \phi_q)^2 = \sum_{q=1}^m\min_{v_q\in \mathcal{H}_q}\left\{\int_0^1 \|v_q(t)\|_{V_q}^2 dt: \bvarphi_{v_q}(1)=\phi_q\right\},
\]
and the objective function becomes 
\begin{align}
\begin{split}\label{eq:obj2}
&\sum_{q=1}^m\int_0^1 \|v_q(t)\|_{V_q}^2 dt + \lambda \sum_{q=0}^{m}U_q({A}_q)  \\
&\qquad\quad\quad+ \frac{1}{\sigma^2}\sum_{k=1}^N \Gamma_k(\pi_r({A}_m\circ\bvarphi_{v_m}(1)\circ{A}_{m-1}\circ\cdots\circ\bvarphi_{v_1}(1)\circ{A}_0(\iota_s(x_k))))
\end{split}
\end{align}
minimized
% over $v_1(\cdot),\dots,v_m(\cdot),$ and ${A}_0,\dots,{A}_m,$  
over ${A}_0,\dots,{A}_m,$ and $v_q\in \mathcal{H}_q,$ $q=1,\dots,m,$
such that $\bvarphi_{v_q}(t)$ satisfies  
\begin{align}
\partial_t\bvarphi_{v_q}(t)&=v_q(t)\circ\bvarphi_{v_q}(t),\quad t\in[0,1] \nonumber \\
\bvarphi_{v_q}(0)&=\mbox{id}.\nonumber
\end{align}
%with initial condition $\bvarphi_q(0)=\mbox{id}.$  
When the norms on the RKHS's are translation invariant, a minimizer of this objective function 
% in $v_1(\cdot),\dots,v_m(\cdot)$ 
% $v_1,\dots,v_m$ 
% with $v_q\in L^2([0,1],V_q)$ always exists for fixed ${A}_0,\dots,{A}_m$ 
always exists. This is demonstrated in Appendix \ref{sec:existence}. 
%under mild regularity conditions on the dependency of $\Gamma$ with respect to $\mathcal{T}_0$ (continuity in $x_1,\dots,x_N$ suffices). 

\section{Forward States}\label{sec:forward}

We define forward states between modules 
as $\xi^0,\zeta^1,\xi^1,\zeta^2,\dots,\xi^m,\zeta^{m+1},$ 
as shown in Figure \ref{fig:models1}, with model input $\xi_k^0$ and model output $\zeta_k^{m+1}.$
The forward states 
\[
\xi_k^q=\bvarphi_{v_q}(1)(\zeta_k^q),\quad q=1,\dots,m
\]
and
% are the output of the corresponding D$_q$ modules and  
% \[
% \xi_k^0=\iota_n(x_k),
% \]
% and 
\[
\zeta_k^{q+1}={A}_{q}(\xi_k^{q}),\quad q=0,\dots,m
\]
are the outputs of the corresponding D$_{q}$ and A$_{q}$ modules, respectively, with initialization
\[
\xi_k^0=\iota_s(x_k).
\]
Let 
\[
z_k^q(t)=\bvarphi_{v_q}(t)(\zeta_k^q)
\]
represent the time-dependent state in $\mathbb{R}^{d_q}$ of module D$_q,$ and denote the array of $N$ states as ${\vect{z}^q(\cdot) = (z_1^q(\cdot),\dots,z_N^q(\cdot)).}$  

\section{Kernel Reduction}\label{sec:trick}

The assumptions in Section \ref{sec:distances} imply that $V_1,\dots,V_m$ are vector-valued RKHSs \citep{aronszajn1950theory,wahba1990spline,joshi2000landmark,miller2002metrics,vaillant2004statistics,micchelli2005learning}.  By Riesz’s representation theorem, each $V_q$ has an associated matrix-valued kernel function
%The kernel associated with $V_q$  is a matrix-valued function
\begin{equation}
    K_q:\mathbb{R}^{d_q}\times\mathbb{R}^{d_q}\rightarrow \mathcal{M}_{d_q}(\mathbb{R})\nonumber
\end{equation}
that reproduces every function in $V_q.$  More precisely, 
for every $y,a\in\mathbb{R}^{d_q}$, there exists a unique element $K_q(\cdot,y)a$ of $V_q$ such that
\[
K_q(\cdot,y)a:x\in\mathbb{R}^{d_q}\mapsto K_q(x,y)a %\quad \forall x\in\mathbb{R}^{d_q},
\]
and
\[
 \langle K_q(\cdot,y)a,f{\rangle}_{V_q} = a^T f(y) %\quad \forall f\in V_q
\]
for all $f\in V_q$.
These properties imply 
\[
\langle K_q(\cdot,x)a,K_q(\cdot,y)b{\rangle}_{V_q} = a^T K_q(x,y)b
\]
and thus symmetry, $K_q(y,x)=K_q(x,y)^T$,
and positive semi-definiteness
for all $x,y,a,b\in \mathbb{R}^{d_q}.$
Conversely, by the Moore–Aronszajn theorem, any matrix-valued kernel that is symmetric and positive semi-definite induces the corresponding vector-valued RKHS of functions reproducible by this kernel.

An RKHS argument similar to the kernel trick used in standard kernel methods can reduce the dimension of our problem as follows.
% Our endpoint cost is a function of 
% the $N$ responses in ${\mathcal{T}_0,}$ 
% the trajectories of the $N$ predictors in ${\mathcal{T}_0}$ through each D$_q,$ propagated by 
% \[
% \partial_t z_k^q(t) = v_q(t)(z_k^q(t)),\quad k=1,\dots,N,\nonumber
% \]
% and ${A}_0,\dots,{A}_m.$  
% The dependence of our endpoint cost on each $v_q$ is through the $N$ endpoints propagated by
% \begin{equation}\label{eqn:sys_0}
% \partial_t z_k^q(t) = v_q(t)\circ z_k^q(t),\quad k=1,\dots,N.\nonumber
% \end{equation}
% The vector fields minimizing this cost are regularized by the RKHS norm $\|\cdot\|_{V_q}$ on their respective spaces $V_q.$
% Each vector field $v_q(t)$ is regularized by the RKHS norm $\|\cdot\|_{V_q}$ on its respective space $V_q.$
The dependence of our endpoint cost on each vector field $v_q(t)$ is through the $N$  trajectories 
\[
\partial_t z_k^q(t) = v_q(t)(z_k^q(t)),\quad k=1,\dots,N\nonumber
\]
generating the $N$ corresponding endpoints $\zeta_1^{m+1},\dots,\zeta_N^{m+1}.$
The vector fields minimizing this cost are regularized by the RKHS norm $\|\cdot\|_{V_q}$ on their respective spaces $V_q.$
By the  representer theorem, % \citep{Scholkopf2002},  
these minimizers   
must then take the form 
\[
v_q(t)(\cdot) = \sum_{l=1}^{N}K_q(\cdot,z_l^q(t))a_l^q(t),
\]
% at each point in time $t,$ 
where 
$\vect{a}^q(\cdot)=(a_1^q(\cdot),\dots,a_N^q(\cdot))$ are the unknown time-dependent vectors in $\mathbb{R}^{d_q}$ to be determined. 
In this reduced representation, our objective function
\begin{align}
\begin{split}\label{eq:obj3}
&\sum_{q=1}^{m} \int_{0}^{1}\sum_{k,l=1}^{N}a_k^q(t)^TK_q(z_k^q(t),z_l^q(t))a_l^q(t)dt \\
&\qquad\qquad\qquad\qquad\qquad\quad + \lambda \sum_{q=0}^{m}U_q({A}_q) + \frac{1}{\sigma^2}\sum_{k=1}^N \Gamma_k(\pi_r(\zeta_k^{m+1})) \nonumber
\end{split}
\end{align}
is minimized  
over $\vect{a}^1(\cdot),\dots,\vect{a}^m(\cdot),$ ${A}_0,\dots,{A}_m,$ subject to the system of trajectories and initial conditions
\begin{align}
    \partial_t z_k^q(t) &= \sum_{l=1}^{N}K_q(z_k^q(t),z_l^q(t))a_l^q(t)\nonumber \\
    z_k^q(0)&=\zeta_k^q={A}_{q-1}(\xi_k^{q-1}) \nonumber \\
    \xi_k^q&=z_k^q(1) \nonumber
\end{align}
and initialization $\xi_k^0=\iota_s(x_k).$ 
Our learning problem can now be solved as an optimal control problem with a finite dimensional %state space.
control space.
% over the finite dimensional control space $\vect{a}^1,\dots,\vect{a}^m$ and learning parameters ${A}_0,\dots,{A}_m$ for the dynamical system  

\section{Optimal Control}\label{sec:control}

% In optimal control theory, one seeks the optimal control that 
An optimal control steers the state of a system from a given initial state to a final state while optimizing an objective function, typically a running cost and an endpoint cost to be minimized. 
% Our learning problem can be similarly cast, as it
Our learning problem can be solved in an optimal control framework, as we 
seek the optimal deformations (control) and affine parameters for our system of trajectories and initial conditions such that a deformation (running) cost and a learning (endpoint) cost are minimized.

Assuming existence of solutions, the Pontryagin Maximum Principle (PMP) \citep{hocking1991optimal,macki2012introduction} provides necessary conditions for optimality in optimal control settings.  By the PMP, an optimal control and trajectory must also solve a Hamiltonian system with a corresponding costate and a stationarity condition.  We derive the PMP for our model within the Lagrangian variational framework in 
% Appendix \ref{sec:PMPoptimal}, 
Appendix \ref{sec:PMP}, 
with the resulting solutions as follows. 

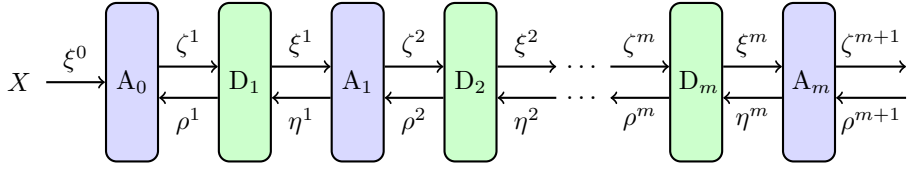
\begin{figure}[h!]
	\begin{center}
		\begin{tikzpicture}[
		auto,
		block_blank/.style={
			rectangle,
			draw=white,
			thick,
			fill=white!20,
			text width=1.3em,
			align=center,
			rounded corners,
			minimum height=6em
		},
		block_D/.style={
			rectangle,
			draw=black,
			thick,
			fill=green!20,
			text width=1.3em,
			align=center,
			rounded corners,
			minimum height=6em
		},
		block_A/.style={
			rectangle,
			draw=black,
			thick,
			fill=blue!15,
			text width=1.3em,
			align=center,
			rounded corners,
			minimum height=6em
		},
		line/.style={
			draw,thick,
			-latex',
			shorten >=2pt
		},
		cloud/.style={
			draw=red,
			thick,
			ellipse,
			fill=red!20,
			minimum height=1em
		}
		]

		\draw (0.,0) node[block_blank] (A) {$X$};
		\path 
		(1.5,0) node[block_A] (B) {$\text{A}_0$}
		(3.,0) node[block_D] (C) {$\text{D}_1$}
		(4.5,0) node[block_A] (D) {$\text{A}_1$}
		(6.,0) node[block_D] (E) {$\text{D}_2$}
		(7.5,0) node[block_blank] (F) {\parbox{1cm}{$\dots$ $\dots$}}
		(9.,0) node[block_D] (G) {$\text{D}_m$}
		(10.5,0) node[block_A] (H) {$\text{A}_m$}
		(12.15,0) node[block_blank] (I) {};%$G(Y,\zeta^{m+1},\\\vect{a}^1(\cdot),\dots,{A}^0)$} ;
		\draw[thick,->] (A.east) -- (B.west) node[midway, above] {$\xi^0$};
		\draw[thick,->] (B.30) -- (C.150) node[midway, above] {$\zeta^1$};
		\draw[thick,<-] (B.-30) -- (C.210) node[midway, below] {$\rho^1$};
		\draw[thick,->] (C.30) -- (D.150) node[midway, above] {$\xi^1$};
		\draw[thick,<-] (C.-30) -- (D.210) node[midway, below] {$\eta^1$};
		\draw[thick,->] (D.30) -- (E.150) node[midway, above] {$\zeta^2$};
		\draw[thick,<-] (D.-30) -- (E.210) node[midway, below] {$\rho^2$};
		\draw[thick,->] (E.30) -- (F.150) node[midway, above] {$\xi^2$};
		\draw[thick,<-] (E.-30) -- (F.210) node[midway, below] {$\eta^2$};
		\draw[thick,->] (F.30) -- (G.150) node[midway, above] {$\zeta^m$};
		\draw[thick,<-] (F.-30) -- (G.210) node[midway, below] {$\rho^m$};
		\draw[thick,->] (G.30) -- (H.150) node[midway, above] {$\xi^m$};
		\draw[thick,<-] (G.-30) -- (H.210) node[midway, below] {$\eta^m$};
		\draw[thick,->] (H.30) -- (I.150) node[midway, above] {$\zeta^{m+1}$};
		\draw[thick,<-] (H.-30) -- (I.210) node[midway, below] {$\rho^{m+1}$};
		\end{tikzpicture}
	\end{center}
	\caption{\label{fig:models1} General model of alternating A and D modules.  Forward states are ${\xi^0,\zeta^1,\xi^1,\zeta^2,\dots,\xi^m,\zeta^{m+1},}$ with ${\xi_k^q=\bvarphi_{v_q}(1)(\zeta_k^q),}$ ${\zeta_k^{q+1}={A}_{q}(\xi_k^{q}),}$ and initialization ${\xi^0_k = \iota_s(x_k).}$ Backpropagation states are ${\rho^1,\eta^1,\rho^2,\dots,\eta^m,\rho^{m+1},}$ with ${\eta_k^q=M_q^T\rho_k^{q+1},}$ ${\rho_k^q= \mathcal{F}_q(\eta_k^q),}$ and initialization ${\rho^{m+1}_k=-\frac{1}{\sigma^2}\iota_r(\nabla\Gamma_k(\pi_r(\zeta^{m+1}_k))).}$}
\end{figure}

First define backpropagation states between modules as $\rho^1,\eta^1,\rho^2,\dots,\eta^m,\rho^{m+1},$ as shown in Figure \ref{fig:models1}, where 
\[
\eta_k^q=M_q^T\rho_k^{q+1},\quad q=m,\dots,1
\]
and
\[
\rho_k^q= \mathcal{F}_q(\eta_k^q),\quad q=m,\dots,1
\]
are states propagating back from corresponding A$_q$ and D$_q$ modules, respectively, with initialization
\[
\rho^{m+1}_k=-\frac{1}{\sigma^2}\iota_r(\nabla\Gamma_k(\pi_r(\zeta^{m+1}_k))).
\]
$\mathcal{F}_q(\eta_k^q)$ is obtained by solving the ODEs 
\begin{equation}
\begin{cases}
\partial_t z_k^q(t) = \sum\limits_{l=1}^{N}K_q(z_k^q(t),z_l^q(t))a_l^q(t),\quad z_k^q(0)=\zeta_k^q \\
\partial_t p_k^q(t)=-\sum\limits_{l=1}^{N}\nabla_1 K_q(z_k^q(t),z_l^q(t))(p_k^q(t)^Ta_l^q(t)+a_k^q(t)^Tp_l^q(t) \\
\phantom{\partial_t p_k^q(t)} \qquad\qquad\qquad\qquad\qquad\qquad\qquad\qquad\quad -2a_k^q(t)^Ta_l^q(t)),\quad p_k^q(1)=\eta_k^q \nonumber
\end{cases}   
\end{equation}
for the state $z_k^q(t)$ and costate $p_k^q(t)$ of module D$_q,$ and assigning $\mathcal{F}_q(\eta_k^q)=p_k^q(0).$ Note the states $z_k^q(t)$ are calculated on the forward pass of the model and cached for the backpropagation pass. 

Let $G$ denote our objective function.  The gradients for determining our optimal control parameters $\vect{a}^1(\cdot),\dots,\vect{a}^m(\cdot)$ and affine parameters ${A}_0,\dots,{A}_m$ are then 
\begin{align}
&\partial_{a_k^q(t)}G = \sum_{l=1}^N K_q(z_k^q(t),z_l^q(t))(2a_l^q(t)-p_l^q(t)),\quad q=1,\dots,m \nonumber\\
&\partial_{M_q}G = \lambda\partial_{M_q}U_q({A}_q) - \sum_{k=1}^N \rho_k^{q+1} {\xi_k^q}^T,\quad q=0,\dots,m \nonumber \\
&\partial_{b_q}G =  - \sum_{k=1}^N \rho_k^{q+1},\quad q=0,\dots,m, \nonumber  
% &\partial_{b_q}G = \lambda\partial_{b_q}U_q({A}_q) - \sum_{k=1}^N \rho_k^{q+1},\quad q=0,\dots,m, \nonumber  
\end{align}
which can be used in gradient descent methods as the directions in which to step the current parameters to minimize the objective function.  Once the parameters 
% or ``weights" 
are updated, another forward pass through our model is run, recalculating the forward states and objective function, followed by backpropagation, recalculating the backpropagation states and gradients. 
 The parameters are then updated again, and the cycle repeated, until a sufficient minimum in the objective function or total gradient is achieved. 
% another forward pass through our model is run to update the forward states and objective function followed by backpropagation to update the backpropagation states, gradients, and then parameters again, repeating until a sufficient minimum in the objective function or total gradient is achieved. 

\section{Subset Training}\label{sec:distill}

Large datasets and large models are time and resource prohibitive in many machine learning tasks. Our model can be naturally adapted to large datasets, in an approach that suggests both model compression and dataset condensation.  
Extending our optimal vector fields model to the more general ``sub-Riemannian" or sub-optimal vector fields approach, 
% The approach trains the diffeomorphisms on a subset of the training data, 
we can train the diffeomorphisms on a subset of the training data, 
which simultaneously decreases the number of model parameters.  During learning, the lower-complexity diffeomorphisms are applied to the entire training dataset for analysis in the endpoint cost.  Similar approximations were introduced in shape analysis (see \citet{younes:hal-02386227} for a review and references) and in \citet{walder2009diffeomorphic,vialard2020shooting}.

We choose a training data subset of size $N_S\le N$ and, without loss of generality, renumber the training data such that its first $N_S$ elements coincide with this subset.  Then the sub-optimal vector fields notation is %$v_q(t)$ at each $t\in[0,1]$ are  
\[
v_q(t)(\cdot) = \sum_{l=1}^{N_S}K_q(\cdot, z_l^q(t))a_l^q(t),%\quad k=1,\dots,N,
\]
% where $\vect{u}_S=( u_i\in\vect{u}, i\in S)$
where ${(z_1^q(\cdot),\dots,z_{N_S}^q(\cdot))}$ and ${\vect{a}^q(\cdot)=(a_1^q(\cdot),\dots,a_{N_S}^q(\cdot))}$ are the states corresponding to this subset and the control parameters, respectively.  
The resulting objective function
\begin{align}
\begin{split}\label{eq:obj4}
&\sum_{q=1}^{m} \int_{0}^{1}\sum_{k,l=1}^{N_S}a_k^q(t)^TK_q(z_k^q(t),z_l^q(t))a_l^q(t)dt \\
&\qquad\qquad\qquad\qquad\qquad\quad + \lambda \sum_{q=0}^{m}U_q({A}_q) + \frac{1}{\sigma^2}\sum_{k=1}^N \Gamma_k(\pi_r(\zeta_k^{m+1})),
\end{split}
\end{align}
is minimized  
over $\vect{a}^1(\cdot),\dots,\vect{a}^m(\cdot),$ ${A}_0,\dots,{A}_m,$ subject to the system of trajectories 
\begin{equation}
    \partial_t z_k^q(t) = \sum_{l=1}^{N_S}K_q(z_k^q(t),z_l^q(t))a_l^q(t),\quad k=1,\dots,N\nonumber
\end{equation}
with the same initial conditions and initialization as in the optimal vector fields case. 
% initial conditions $z_k^q(0)=\zeta_k^q={A}_{q-1}(\xi_k^{q-1}),$ $\xi_k^q=z_k^q(1),$ and initialization ${\xi_k^0=\iota_n(x_k).}$ 
The existence of a minimizer of this objective function 
is demonstrated in Appendix \ref{sec:existence}.  The PMP is derived in Appendix \ref{sec:PMP}, resulting in more general expressions for the costate trajectories and gradients for the optimal control.

\section{Dummy Dimensions}\label{sec:addeddims}

\begin{figure}[!htb]
	\includegraphics[height=5.9cm,trim={1cm 5cm 16cm 5cm},clip]{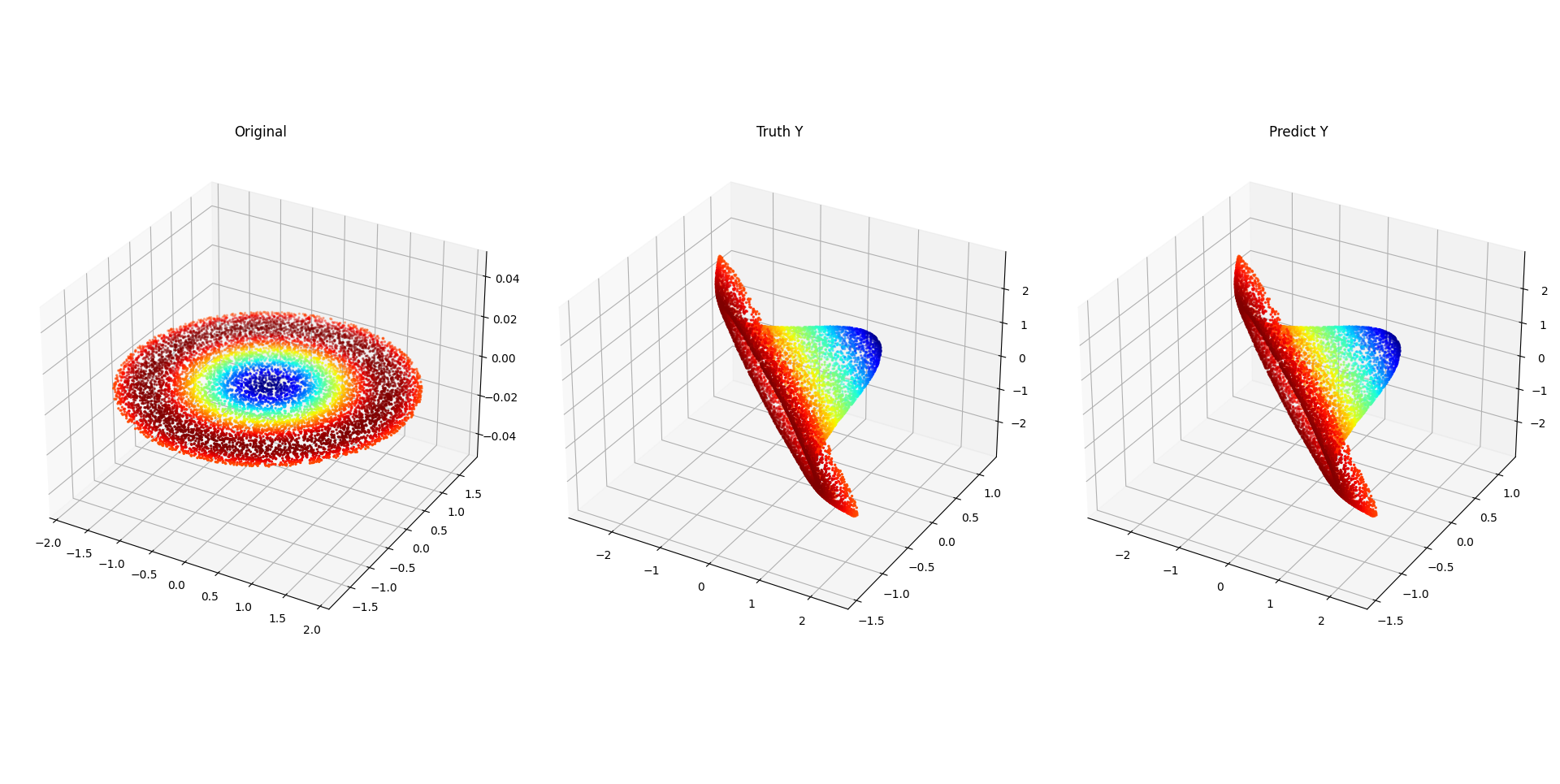}
 %                                trim={<left> <lower> <right> <upper>}
	\caption{Two-dimensional Rings dataset (left) with a  linear representation of the color-coded true responses (right) following a diffeomorphism on the domain with one added dummy dimension.}
	\label{fig:Rings}
\end{figure}

Adding ``dummy" dimensions to a dataset provides two benefits in our setting \citep{younes2019diffeomorphic,dupont2019augmented}.  First, in cases where a diffeomorphism of the given domain cannot 
reshape the data to within an affine transformation of the true responses 
for successful regression---or is too costly to do so---adding dimensions can provide a more viable or less costly pathway for the diffeomorphism.  An example is illustrated with the two-dimensional Rings on the left in Figure \ref{fig:Rings}, where the data point locations and colors represent the predictors and true responses, respectively.  Zero padding the predictors with one additional dimension then applying 
 % several iterations of 
 our model\footnote{The baseline ADA model described in Section \ref{sec:experiments}.} leads to a 
 % straightforward diffeomorphic linear separation of the data, 
%fast 
linear representation of the true responses by a 
simple diffeomorphism of the predictors
 as shown on the right.  
Second, the construction of our diffeomorphisms is predicated on the assumption of data non-redundancy.  In cases where predictors may be redundant, e.g., in real datasets, one can initialize the extra dimensions with random number values small enough to break the symmetry without impacting data structure.
% Second, our model is predicated on the uniqueness of diffeomorphic flow.  In cases of redundant predictors with different responses, as can be found in real datasets due to low precision or errors in data collection, this uniqueness is preserved by initializing the extra dimensions with random number values small enough to break the symmetry without impacting data structure.

\section{Implementation}\label{sec:implementation}
The model is implemented in Python using a dynamic programming approach, with 
objective function
% \begin{equation}
% \sum\limits_{q=1}^{m} \frac{1}{T_q}\sum\limits_{i=0}^{T_q-1}\sum\limits_{k,l=1}^{N}a_k^q(i/T_q)^TK_q(z_k^q(i/T_q),z_l^q(i/T_q))a_l^q(i/T_q) 
% + \lambda \sum\limits_{q=0}^{m}U_q({A}_q) + \frac{1}{\sigma^2}\sum\limits_{k=1}^N \Gamma_k(\pi_r(\zeta_k^{m+1}))\nonumber
% \end{equation}
\begin{align}
\sum\limits_{q=1}^{m} \frac{1}{T_q}\sum\limits_{i=0}^{T_q-1}\sum\limits_{k,l=1}^{N}a_k^q(i/T_q)^TK_q(z_k^q(i/T_q)&,z_l^q(i/T_q))a_l^q(i/T_q) \nonumber \\
& + \lambda \sum\limits_{q=0}^{m}U_q({A}_q) + \frac{1}{\sigma^2}\sum\limits_{k=1}^N \Gamma_k(\pi_r(\zeta_k^{m+1}))\nonumber
\end{align}
minimized over %$A_q:x\in \mathbb{R}^{d_q}\mapsto M_q(x)+b_q\in \mathbb{R}^{d_{q+1}}$ 
% \begin{enumerate}[label=(\roman*)] % [label=(\Alph*)] % [label=(\alph*)]
%     \item $\vect{a}^q(t/T_q) = (a_1^q(t/T_q),\dots,a_N^q(t/T_q))\in (\mathbb{R}^{d_q})^N,\quad t = 0,\dots,T_q-1,\quad q=1,\dots,m$  
%     \item ${M_q\in\mathcal M_{d_{q+1},d_{q}}(\mathbb{R}),}$ $b_q\in\mathbb{R}^{d_{q+1}},\quad q=0,\dots,m$
% 		% \item $\theta\in\mathbb{R}^{(\tilde{d}_1+1)\times \tilde{d}_2}$
% \end{enumerate}
\begin{equation}
    \vect{a}^q(i/T_q) = (a_1^q(i/T_q),\dots,a_N^q(i/T_q))\in (\mathbb{R}^{d_q})^N,\quad i = 0,\dots,T_q-1,\quad q=1,\dots,m \nonumber
\end{equation}
and ${A}_0,\dots,{A}_m,$ i.e.,
\[
M_q\in\mathcal M_{d_{q+1},d_{q}}(\mathbb{R}),\mbox{ }b_q\in\mathbb{R}^{d_{q+1}},\quad q=0,\dots,m,
\]
subject to 
\[
z_k^q((i+1)/T_q) = z_k^q(i/T_q) + \frac{1}{T_q}\sum_{l=1}^{N}K_q(z_k^q(i/T_q),z_l^q(i/T_q))a_l^q(i/T_q) %,\quad i = 0,\dots,T_q-1,\quad q=1,\dots,m 
\] 
with $z_k^q(0)=\zeta_k^q={A}_{q-1}(\xi_k^{q-1}),$ $\xi_k^q=z_k^q(1),$ 
and initialization $\xi_k^0=\iota_s(x_k).$ 
The model parameters are 
initialized as 
\begin{enumerate}[label=(\roman*)]
 \item $\vect{a}^q(i/T_q)=0\in(\mathbb{R}^{d_q})^N,\quad i = 0,\dots,T_q-1,\quad q=1,\dots,m$
% \item $\vect{a}=0\in\mathbb{R}^{\tilde{d}_2\times N \times T_q}$
\item $M_q\sim\mathcal{N}(0,0.01^2)\in\mathcal{M}_{d_{q+1},d_q}(\mathbb{R}),$ $b_q=0\in\mathbb{R}^{d_{q+1}},\quad q=0,\dots,m.$		
% \item $\theta_0=$ ridge regression solution of $\mbox{\textbf{T}}$ with $\lambda$
\end{enumerate}
We include an option to speed up kernel computations using PyKeOps  \citep{charlier2021kernel} with user-specified precision and GPUs.  Our optimization algorithms are gradient descent methods implemented with line search. %, restarting in the event of a bad direction in the line search.  For the case of PyKeOps kernel computations, the gradient descent restart uses float64 precision for these computations, then resets back to the user-specified precision.  

To run the model, the user specifies an arbitrary sequence and number of non-identity A and D modules, dimension parameters $s,$ $r,$ and $d_1,\dots,d_{m},$ ridge regularization weight $\lambda,$ and optimization algorithm parameters for gradient descent, including stopping thresholds and maximum number of iterations. 
For each D$_q$ module, the user specifies the kernel type 
% and parameters, including width $h_q,$ 
and the number of discretized time points $T_q$ for state and costate propagation and the control variables.  For each kernel $K_q,$ the algorithm assumes a default kernel width $h_q$ of 0.5, as the affine module preceding D$_q$ automatically scales and adapts its input to the kernel width of the subsequent D$_q.$
%as the input data is standardized to unit variance and the A$_0$ module then scales and adapts the data to this kernel width of the subsequent D$_1$ diffeomorphism.  
Input and output dimension assignments for each module in the sequence are automated by our algorithm based on $d_X,$ $d_Y,$ $s,$ $r,$ and the inner module dimensions provided by the user. 
The normalization factor $\sigma$ of the error term is determined by our model as a function of the training data and initial training iterations, as described in Section \ref{sec:normfactor}.

\section{Data Preprocessing}

Prior to training, the $\vect{x}$ and $\vect{y}$ training data in $\mathcal{T}_0$ are standardized to 
zero mean and unit variance by subtracting their respective means, ${\mu_X\in\mathbb{R}^{d_X}}$ and ${\mu_Y\in\mathbb{R}^{d_Y},}$ and dividing by their 
respective standard deviations, ${\sigma_X\in\mathbb{R}^{d_X}}$ and ${\sigma_Y\in\mathbb{R}^{d_Y}.}$  The test predictors are standardized using the 
standardization parameters of the training predictors, $\mu_X$ and $\sigma_X.$  For $s>0,$  $s$ extra dimensions are then appended to the training and test predictors by $N$ vector draws from $\mathcal{N}(0,0.01^2)\in\mathbb{R}^s$ and $N_{\mathit{test}}$ zero vectors $0\in\mathbb{R}^s,$ respectively, where $N_{\mathit{test}}$ is the number of data points in the test set.
% For the case of $n>0,$ added dimensions are appended to the standardized training and test inputs by vector draws from $\mathcal{N}(0,0.01^2)\in\mathbb{R}^n$ and zero vectors $0\in\mathbb{R}^n,$ respectively.
% The additional $X$ dimension $n$ is initialized as ${\sim\mathcal{N}(0,0.01^2)\in\mathbb{R}^{N \times n}.}$   

\section{Normalization Factor and Model Training}\label{sec:normfactor}

To determine an optimal penalty for endpoint matching errors, the $\sigma$ normalization factor 
of the error term is calculated by the model as follows.  
% is first initialized as a function of the (unappended) standardized 
% training data, 
For each data point $x_{i}$ in the unappended, standardized %training data 
$\mathcal{T}_0,$  $(y_{j} - y_{i}), j\in J_i$ is linearly regressed on ${(x_j - x_{i}), j\in J_i},$ where $J_{i}$ indexes the $k$ nearest neighbors of $x_{i}$ for
${k=\min \left\{ 2 d_X+1,\lfloor \frac{N}{5} \rfloor \right\}.}$ This regression (without intercept) estimates
``gradients'' $g_{i}\in\mathcal M_{d_Y,d_X}(\mathbb{R})$, with residuals
%are applied to the nearest neighbor (indexed $j_{i}(1)$) of $x_{i}$ to calculate the respective errors $\sigma_{i}$  
% in 
% \[
% y_j = y_{i_0} + g_{i_0} (x_j-x_{i_0}) + \sigma_{i_0} 
% \]
\[
r_{ji} = (y_j - y_{i}) - g_{i} (x_j-x_{i}), j\in J_i
\]
and mean square error
\[
\sigma^2_{\mathrm{MSE}} = \frac1{Nkd_Y} \sum_{i=1}^N\sum_{j\in J_i} \|r_{ji}\|^2.
\]
The initial $\sigma$ is set to
% \[
% \sigma = N^{\frac{1}{4}}\max \left\{ \left\| \sqrt{\sigma^2_{\mathrm{MSE}}}\oslash \sigma_Y\right\|_1, 0.1  \right\},
% \]
\[
\sigma^2 = N^{\frac{1}{2}}\max \left\{\sqrt{\sigma^2_{\mathrm{MSE}}}/2, 0.01  \right\},
\]
% \[
% \sigma = N^{\frac{1}{4}}\max \left\{ \sqrt{\sigma^2_{\mathrm{MSE}}}\cdot \sigma_Y^{-1}, 0.1  \right\},
% \]
% where $\sigma^2_{\mathrm{MSE}}$ is
% the mean-square error (MSE) scaled by $\frac{1}{2}\sigma_Y^2$
% \[
% \sigma^2_{\mathrm{MSE}} = \frac{1}{2}\sigma_Y^2 \odot \frac{1}{N}\sum_{i=1}^N \sigma_{i}^2.
% \]

Training begins with the initialized model parameters $\vect{a}^1(\cdot),\dots,\vect{a}^m(\cdot),$ ${A}_0,\dots,{A}_m,$ the initial $\sigma,$ and the appended, standardized training data.
% and lower stopping threshold for objective function,
%$\vect{a}^1,\dots,\vect{a}^m,{A}_0,\dots,{A}_m,$ 
The model iteratively decreases $\sigma$ until the training MSE 
%(with standardization removed)
\[
\frac{1}{N}\sum_{k=1}^{N}\Gamma_k(\sigma_Y
\odot\pi_r(\zeta_{k}^{m+1})+\mu_Y),
\]
using the unstandardized experimental responses $y_k,$ 
% removed from % the model outputs and 
% the experimental responses ${y_k\mapsto \sigma_Y\odot y_k + \mu_Y,}$
is less than %training error threshold 
\[
\max \left\{ \sigma^2_{\mathrm{MSE}} ,0.01\right\}
\]
% a data-dependent upper bound.
or a maximum number of model loops are reached.
Using the final value for $\sigma$ and parameters $\vect{a}^1(\cdot),\dots,\vect{a}^m(\cdot),{A}_0,\dots,{A}_m$ initialized to their final values in this step, a final training loop through the model is executed to complete training.

\section{Evaluation Metric}

The diffeomorphisms and affine transformations learned on the training set are applied to the corresponding test set for performance analysis. Specifically, the test predictors are forward propagated through the model, transformed in turn 
by the learned affine transformations of each A$_q$ and the vector fields of each D$_q,$ the latter functions of the learned $\vect{a}^q(\cdot)$ and cached $\vect{z}^q(\cdot).$   The  evaluation metric is  root-MSE (RMSE) between the model outputs $\zeta_{k,\mathit{test}}^{m+1}$ and the test experimental responses 
\[
\sqrt{\frac{1}{N_{\mathit{test}}}\sum_{k=1}^{N_{\mathit{test}}}\Gamma_k(\sigma_Y
\odot\pi_r(\zeta_{k,\mathit{test}}^{m+1})+\mu_Y)},
\]
which we will denote test RMSE.

\section{Baseline Experiments}\label{sec:experiments}

\begin{figure}[h!]
	\begin{center}
		\begin{tikzpicture}[
		auto,
		blockD/.style={
			rectangle,
			draw=black,
			thick,
			fill=green!20,
			text width=1.em,
			align=center,
			rounded corners,
			minimum height=4em
		},
		blockA/.style={
			rectangle,
			draw=black,
			thick,
			fill=blue!20,
			text width=1.em,
			align=center,
			rounded corners,
			minimum height=4em
		},
		line/.style={
			draw,thick,
			-latex',
			shorten >=2pt
		},
		cloud/.style={
			draw=red,
			thick,
			ellipse,
			fill=red!20,
			minimum height=1em
		}
		]

		\draw (0.75,0) node (A) {$X$};
		\path 
		(2,0) node[blockA] (B) {A$_0$}
		(3,0) node[blockD] (C) {D$_1$}
		(4,0) node[blockA] (D) {A$_1$}
		(5.25,0) node (E) {$Y$} ;
		\draw[thick,->] (A.east) -- (B.west) node[midway, above] {$\iota_1$};
		\draw[thick,->] (B.east) -- (C.west) node[midway, above] {};
		\draw[thick,->] (C.east) -- (D.west) node[midway, above] {};
		\draw[thick,->] (D.east) -- (E.west) node[midway, above] {$\pi_0$};
		
		\end{tikzpicture}
	\end{center}
	\caption{\label{fig:models2} ADA transformation sequence used in the experiments.}
\end{figure}
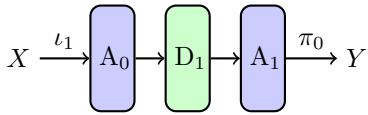

While DA and AD are the smallest possible sequences that can be represented in our model, the AD sequence is not as practical for regression purposes, and the DA sequence requires the user to specify a data-specific kernel width for the D$_1$ diffeomorphism.  Therefore, we consider the ADA model, which is the sequence case for $m=1$ and no identity modules, as our simplest regression model sequence, and we choose this baseline model for our experiments, as shown in Figure \ref{fig:models2}.  Additionally, we choose the simplest reasonable values for our model parameters.  
We set $\lambda=1$ and assign dimensions ${s=1,}$ ${r = 0,}$ and ${d_1=d_X+s,}$ ensuring the dummy dimension added to the dataset is carried through module A$_0$ to the diffeomorphism in D$_1.$ 
For module D$_1,$ we set ${T_1=10}$, and we construct a matrix-valued kernel from the scalar 
%3\textsuperscript{rd}-order Laplacian kernel 
Mat\'ern kernel and the identity matrix $\mathrm{I}_{d_1}$ \citep{younes2019diffeomorphic}.  In particular, 
\[
K_1(x,y)=\left(1+u+0.4u^2+\frac{1}{15}u^3\right)e^{-u}\mathrm{I}_{d_1},\quad u = \frac{|y-x|}{h_1}
\]
with default kernel width $h_1=0.5$.
The optimization algorithm is the limited-memory Broyden–Fletcher–Goldfarb–Shanno algorithm (L-BFGS) with Wolfe conditions on the line search.  Early stopping, typically used to prevent overfitting, is avoided by setting the
maximum number of gradient descent iterations large enough 
to ensure numerical convergence.
% i.e., no early stopping to avoid overfitting.   
% The maximum number of gradient descent iterations is set to 2000.

Our model is tested on nine UCI datasets---{\it Concrete}, {\it Energy}, {\it Kin8nm}, {\it Naval}, {\it Power}, {\it Protein}, {\it Wine Red}, {\it Yacht}, and {\it Year}---with standard splits\footnote{https://github.com/yaringal/DropoutUncertaintyExps/tree/master/UCI\_Datasets} originally generated for the experiments in \cite{pmlr-v37-hernandez-lobatoc15} and gap splits\footnote{https://github.com/cambridge-mlg/DUN/tree/master/experiments/data/UCI\_for\_sharing} generated by \cite{Foong2019InBetweenUI}. Datasets are split into training and test sets by uniform subsampling for the standard splits and by a custom split assigning ``outer regions" to the training sets and ``middle regions" to the test sets for the gap splits.
For the standard splits, 
20 randomized train-test splits (90\% train, 10\% test) of each dataset are provided, with the exception of the larger {\it Protein} (5 splits) and {\it Year} (1 split) datasets. Note that the {\it Year} standard split is not provided in the standard splits repositories, so we assume it follows the single split (90\% train, 10\% test)  guideline\footnote{https://archive.ics.uci.edu/ml/datasets/yearpredictionmsd} provided for that dataset in the UCI respository.    
For the gap splits, $d_X$ train-test splits of each dataset are provided, each split corresponding to one of the $d_X$ dimensions of that dataset.  These splits are generated by sorting the data points in increasing order in the dimension of interest, then assigning the middle third to the test set and the outer two-thirds to the training set.  The {\it Year} dataset is not included in the gap splits repository or experiments.  For each multiple split experiment, the evaluation metric is test RMSE averaged over all splits with standard error. 

The total number of data points $N_T$ prior to splitting and the dimensions $d_X$ and $d_Y$ of each provided dataset are listed in Tables \ref{tab:experiments_standard_gap1} and \ref{tab:experiments_standard_gap2}.  
Note that although two of the original datasets---{\it Energy} and {\it Naval}---have response dimension ${d_Y=2,}$ all provided standard and gap splits have $d_Y=1.$   
In the {\it Year} dataset (${N_T=515345,}$ ${d_X=90}$) experiment, to make it computationally tractable, we set $d_1 = 10$  to reduce dimensionality and train the diffeomorphisms on a training data subset ($N_S= 1000$) selected from the training data as the initial $N_S$ cluster seeds for $k$-means clustering according to the $k$-means++ algorithm.
Kernel computations are performed using PyKeOps in all experiments. %, with float64 precision in all experiments except those involving the larger {\it Protein} (float32) and {\it Year} (float32) datasets.

We implement standard ridge regression (A) and five DNNs for performance comparison with our model. Ridge regression is implemented in Python with regularization weight ${\lambda=1.}$ The DNN models, implemented in TensorFlow and denoted DNN-x, ${\mathrm{x}=1,2,3,5,10}$, consist of x sequential densely-connected hidden layers with ReLU activation and layer sizes listed in Table \ref{tab:TF}, followed by a densely-connected output layer.  In TensorFlow, we use the Adam optimizer \citep{https://doi.org/10.48550/arxiv.1412.6980}, MSE loss, and 400 training epochs or number of complete passes through the training datasets.  Default values are assumed for all other TensorFlow parameters, including learning rate of 0.001, batch size of 32, and no validation split of the data. 
The A and DNN models are trained and tested on the standardized datasets, and the standardization is removed from the model outputs for performance analysis. 
\begin{table}[ht!] 
	\caption{Hidden Layer Sizes of Densely-Connected Neural Networks.}
	\label{tab:TF}
		\centering\small
\begin{adjustbox}{center}	
		\begin{tabular}{@{}>{\bfseries}l*{10}{r}@{}}
			&\multicolumn{10}{@{}c@{}}{\bfseries Hidden Layer}\\
		    \addlinespace[1pt]
    		\cline{2-11}
    		\addlinespace[3pt]
			 Model  & \textbf{1}& \textbf{2} & \textbf{3}  & \textbf{4} & \textbf{5} & \textbf{6}& \textbf{7} & \textbf{8}  & \textbf{9} & \textbf{10}   \\
			\midrule%[0.75pt]
            DNN-1& 64 & \phantom{256} & \phantom{256} & \phantom{256} & \phantom{256} & \phantom{256} & \phantom{256} & \phantom{256} & \phantom{256} & \phantom{256} \\
            DNN-2& 128 & 64 & \phantom{256} & \phantom{256} & \phantom{256} & \phantom{256} & \phantom{256} & \phantom{256} & \phantom{256} & \phantom{256} \\
            DNN-3& 256 & 128 & 64 & \phantom{256} & \phantom{256} & \phantom{256} & \phantom{256} & \phantom{256} & \phantom{256} & \phantom{256} \\
            DNN-5& 256 & 128 & 64 & 32 & 16 & \phantom{256} & \phantom{256} & \phantom{256} & \phantom{256} & \phantom{256} \\
            DNN-10&	256 & 128 & 64 & 32 & 16 & 16 & 8 & 8 & 4 & 4\\
			\bottomrule%[2pt]
		\end{tabular}
	\end{adjustbox}
\end{table}

Performance of our ADA model is compared in Tables \ref{tab:experiments_standard_gap1} and \ref{tab:experiments_standard_gap2} 
with the A and DNN models and with RMSE 
experimental results found in the literature using the same standard splits and gap splits.  
The literature results in those tables are the top performing models from each literature reference in Table \ref{tab:LiteratureResults} that conducted experiments on the same standard splits and gap splits.
A comprehensive list of all Table \ref{tab:LiteratureResults} results is found in Appendix Tables \ref{tab:all_literature_experiments_standard1} and \ref{tab:all_literature_experiments_standard2} for standard split experiments and Tables \ref{tab:all_literature_experiments_gap1} and \ref{tab:all_literature_experiments_gap2} for gap split experiments.  Gray shading in Tables \ref{tab:all_literature_experiments_standard1} and \ref{tab:all_literature_experiments_standard2} indicates experiments using standard splits that are different 
% or potentially different \citep{ober2019benchmarking}
from those used in our experiments 
but generated following the 
training-test
protocol from \citet{pmlr-v37-hernandez-lobatoc15}.
The literature models include Bayesian deep learning techniques such as 
variational inference (VI); backpropagation (BP) and probabilistic BP (PBP) for Bayesian NNs (BNNs);
% with one hidden layer in the network; BP-x and PBP-x for x hidden layers ($\mathrm{x}>1$); 
Monte Carlo dropout run in a timed setting (Dropout-TS or Dropout), to convergence (Dropout-C), and with grid hyperparameter tuning (Dropout-G); BNNs with variational matrix Gaussian posteriors (VMG) and horseshoe priors (HS-BNN); and PBP with the matrix variate Gaussian distribution (PBP-MV).   
Additional models are Bayes by backprop (BBB); stochastic, low-rank, approximate natural-gradient (SLANG) method; % for variational inference in large, deep models; 
variations of the neural linear (NL) model: 
% in which a neural network extracts features from the input to be used as basis functions for Bayesian linear regression,
maximum a posteriori (MAP) estimation NL (MAP NL), regularized NL (Reg NL), Bayesian noise (BN) NL by marginal likelihood maximization (BN(ML) NL) and by Bayesian optimization (BO) (BN(BO) NL);
% where x is the number of hidden layers in the network; 
depth uncertainty network (DUN) 
% a probabilistic model that treats the depth of a neural network as a random variable over which to perform inference
with multi-layer perceptron (MLP) architecture (DUN (MLP)); deep ensembles (Ensemble); Gaussian mean field VI (MFVI); vanilla NNs (SGD); and distributional regression by negative log-likelihood (NLL) with alternative loss formulation ($\beta\mathrm{-NLL}$) ($\mathcal{L}_{\beta\mathrm{-NLL}}$), ``moment matching" (MM) ($\mathcal{L}_{\mathrm{MM}}$), MSE loss ($\mathcal{L}_{\mathrm{MSE}}$), Student's t-distribution (Student-t), and 
different variance priors and variational inference (xVAMP, xVAMP*, VBEM, VBEM*). An integer ``-x" appended to a model name denotes x hidden layers in the network.
All presented literature results involve some form of hyperparameter tuning,
typically by BO or a grid approach, using a portion of each training set as a validation set. 

\begin{table}[h!] 
	\caption{Literature Models Tested on Standard Splits (S), Gap Splits (G), and Different Standard Splits (D).}
	\label{tab:LiteratureResults}
	\begin{threeparttable}
    \centering\small
  
    \begin{adjustbox}{center}	
        \begin{tabular}{lll}
        \multicolumn{1}{l}{\textbf{Models}} & \multicolumn{1}{l}{\textbf{Splits}} & \multicolumn{1}{l}{\textbf{Reference}}\\
			\toprule
            VI, BP, PBP & S & \citet{pmlr-v37-hernandez-lobatoc15}\\
            Dropout-TS & S &
            \citet{pmlr-v48-gal16} \\
			VMG & D & \citet{pmlr-v48-louizos16}\\  
            HS-BNN & D & \citet{2017arXiv170510388G}\\ 
            PBP-MV & D & \citet{pmlr-v54-sun17b}\\  
            Dropout-C, Dropout-G & S & \citet{https://doi.org/10.48550/arxiv.1811.09385} \\
            BBB, SLANG & S & \citet{NEURIPS2018_d3157f2f} \\
            MAP, MAP NL, Reg NL,   & \multirow{2}{*}{D,G} & \multirow{2}{*}{\citet{ober2019benchmarking}}\\
            BN(ML) NL, BN(BO) NL & & \\
            DUN, DUN (MLP), Dropout,  & \multirow{2}{*}{S,G} & \multirow{2}{*}{\citet{NEURIPS2020_781877bd}} \\
            Ensemble, MFVI, SGD & & \\
            $\mathcal{L}_{\beta\mathrm{-NLL}},$ $\mathcal{L}_{\mathrm{MM}},$ 
            $\mathcal{L}_{\mathrm{MSE}},$ Student-t,  &  \multirow{2}{*}{S,D} & \multirow{2}{*}{\citet{Seitzer2022Pitfalls}} \\
            xVAMP, xVAMP*, VBEM, VBEM* & & \\
			\bottomrule
		\end{tabular}
	\end{adjustbox}
    \end{threeparttable}
\end{table}

For consistency in performance comparison, we convert the standard deviation results in \citet{2017arXiv170510388G}, \citet{NEURIPS2020_781877bd}, and \citet{Seitzer2022Pitfalls} to standard errors and use the standard error representation of the results in \citet{pmlr-v48-gal16} found in \citet{https://doi.org/10.48550/arxiv.1811.09385}.   
Due to size, the larger {\it Protein} and {\it Year} datasets are not analyzed in some of the literature references.  
\citet{pmlr-v48-louizos16} and \citet{pmlr-v54-sun17b} generate their own standard splits, following the training-test protocol from \citet{pmlr-v37-hernandez-lobatoc15}, and randomly generate the {\it Year} data split. 
\citet{Seitzer2022Pitfalls} also generate their own standard splits for the {\it Energy} and {\it Naval} datasets (maintaining the original response dimensions of $d_Y=2$) and use the standard splits from \citet{pmlr-v37-hernandez-lobatoc15} for the rest of the datasets. 
In
\citet{2017arXiv170510388G} and
\citet{ober2019benchmarking}, %use the gap splits generated by \cite{Foong2019InBetweenUI} but 
it is unclear if the standard splits are those used in \citet{pmlr-v37-hernandez-lobatoc15} or if they are generated by the authors following that training-test protocol, thus these results are shaded in gray in Tables \ref{tab:all_literature_experiments_standard1} and \ref{tab:all_literature_experiments_standard2}.
All literature results are provided in 2-digit decimal precision, with the exception of 3-digit decimal precision in \citet{pmlr-v37-hernandez-lobatoc15}, \citet{NEURIPS2020_781877bd} and the {\it Kin8nm} and {\it Wine Red} analysis in \citet{Seitzer2022Pitfalls} and 4-digit decimal precision for the {\it Naval} analysis in \citet{Seitzer2022Pitfalls}. 

\begin{figure}[hbt!] % [height=5.4cm,trim={3.5cm 1.3cm 2.8cm 1.6cm},clip]
	\includegraphics[height=5.6cm,trim={3.5cm 1.2cm 2.8cm 1.7cm},clip]{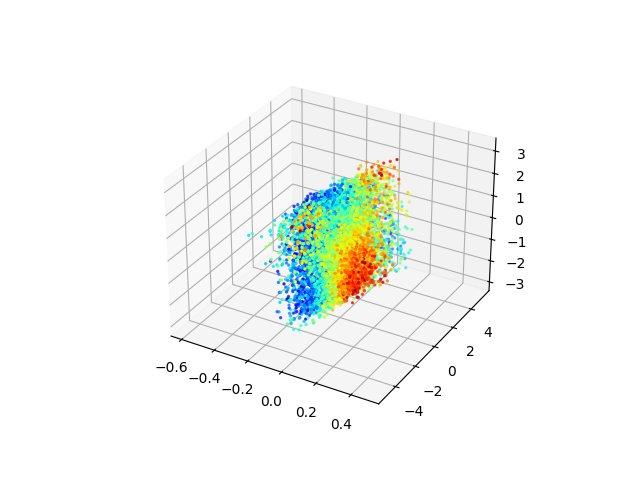} % trim = {left, bottom, right, top}
 \includegraphics[height=5.6cm,trim={3.5cm 1.2cm 2.8cm 1.7cm},clip]{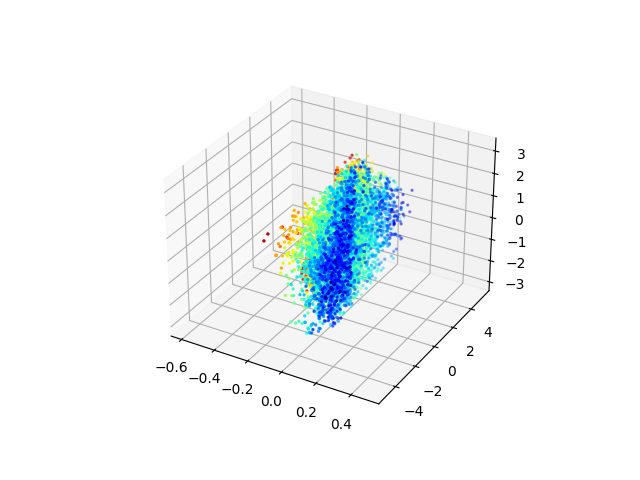} 
  \includegraphics[height=5.6cm,trim={3.5cm 1.2cm 2.8cm 1.7cm},clip] {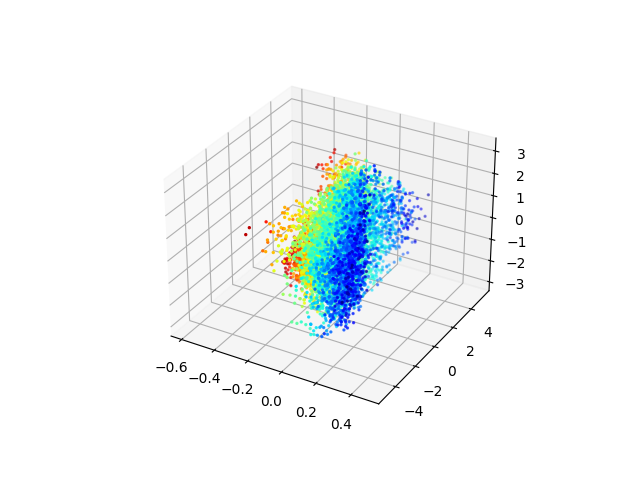}  %
    \includegraphics[height=5.6cm,trim={3.5cm 1.2cm 2.8cm 1.7cm},clip]{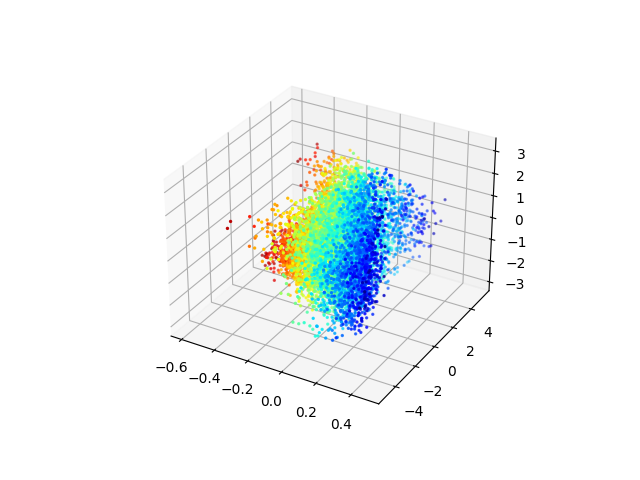}
    \includegraphics[height=5.6cm,trim={3.5cm 1.2cm 2.8cm 1.7cm},clip]{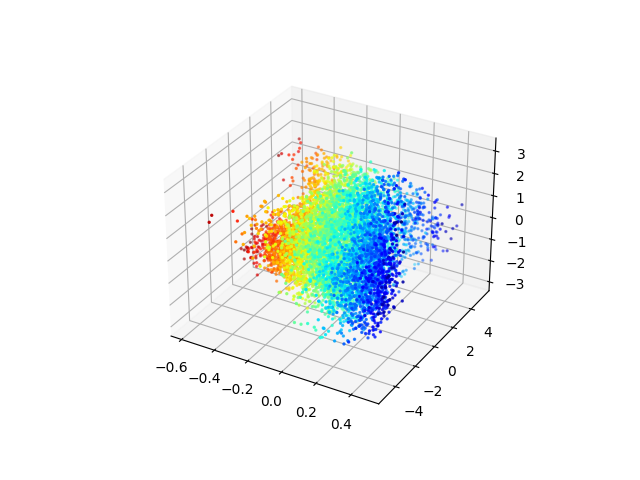}  %
  \includegraphics[height=5.6cm,trim={3.5cm 1.2cm 2.8cm 1.7cm},clip]{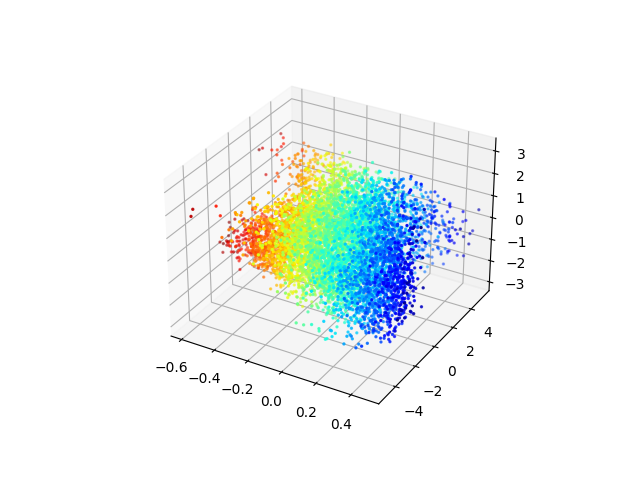}
	\caption{Reshaped sequence  of a {\it Kin8nm} standard training split through module D$_1.$ Each plot corresponds to the first three principal components of $\vect{z}^1(t)$ at a fixed time $t,$ with the true responses color coded.  Starting at top and viewing left to right, $t=0,$ 0.2, 0.4, 0.6, 0.8, and 1, respectively.} 
	\label{fig:Kin8nm}
\end{figure}

\begin{figure}[!htb] % [height=6.3cm,trim={3.5cm 1cm 2.8cm 1cm},clip]
	\includegraphics[height=5.6cm,trim={3.5cm 1.2cm 2.8cm 1.7cm},clip] {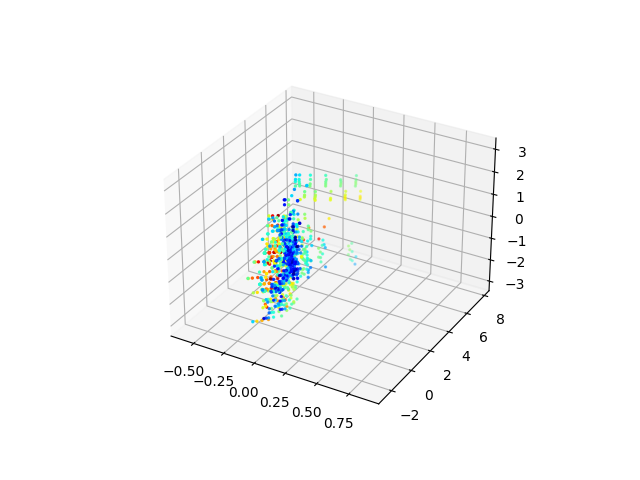}
 \includegraphics[height=5.6cm,trim={3.5cm 1.2cm 2.8cm 1.7cm},clip]{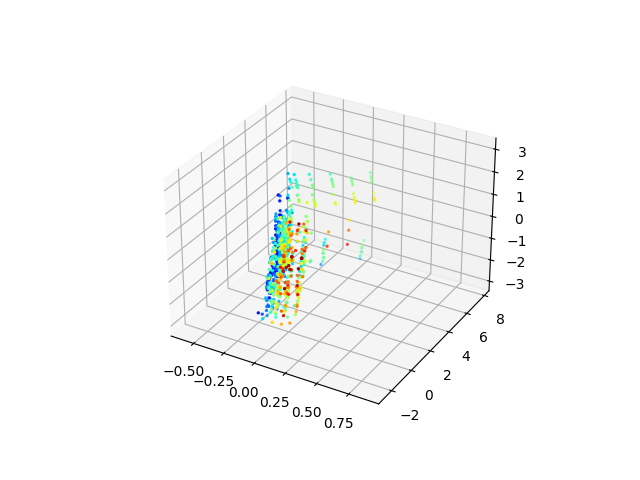}
  \includegraphics[height=5.6cm,trim={3.5cm 1.2cm 2.8cm 1.7cm},clip]{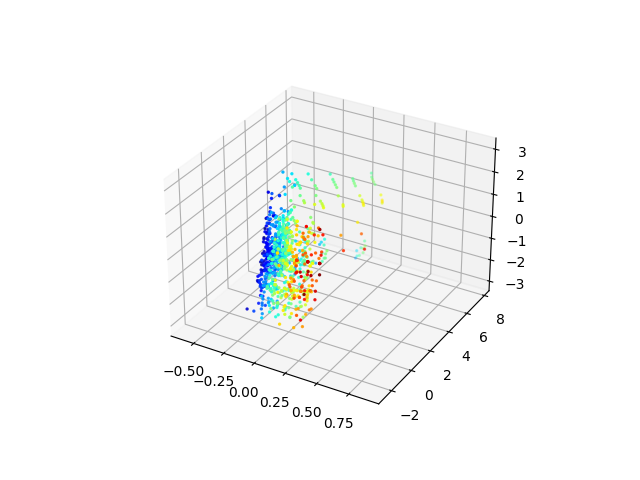}
    \includegraphics[height=5.6cm,trim={3.5cm 1.2cm 2.8cm 1.7cm},clip]{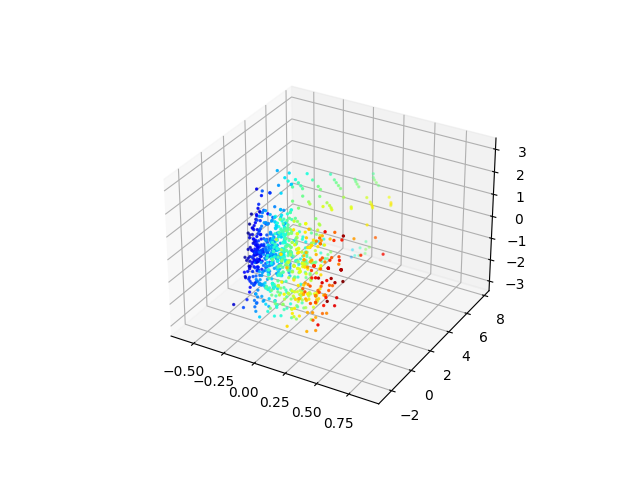}
        \includegraphics[height=5.6cm,trim={3.5cm 1.2cm 2.8cm 1.7cm},clip]{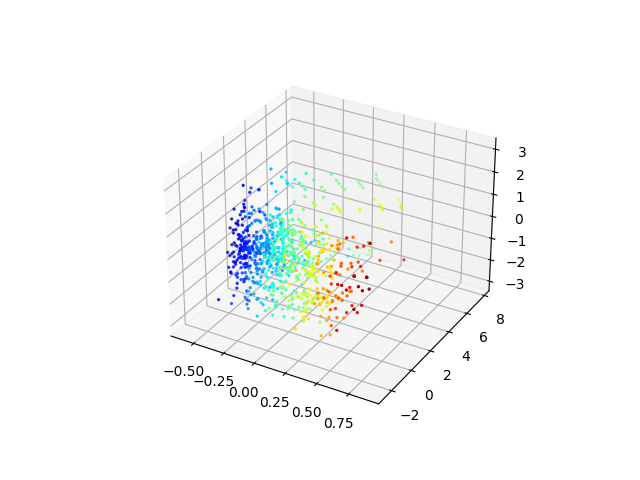}
  \includegraphics[height=5.6cm,trim={3.5cm 1.2cm 2.8cm 1.7cm},clip]{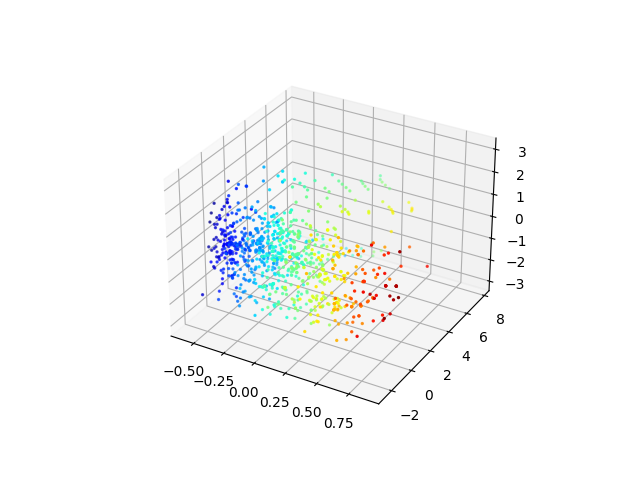}
	\caption{Reshaped sequence  of a {\it Concrete} standard training split through module D$_1.$ Each plot corresponds to the first three principal components of $\vect{z}^1(t)$ at a fixed time $t,$ with the true responses color coded.  Starting at top and viewing left to right, $t=0,$ 0.2, 0.4, 0.6, 0.8, and 1, respectively.}
	\label{fig:Concrete}
\end{figure}

\begin{figure}[!htb] % [height=6.3cm,trim={3.5cm 1cm 2.74cm 1cm},clip]
	\includegraphics[height=5.6cm,trim={3.5cm 1.1cm 2.74cm 1.7cm},clip]{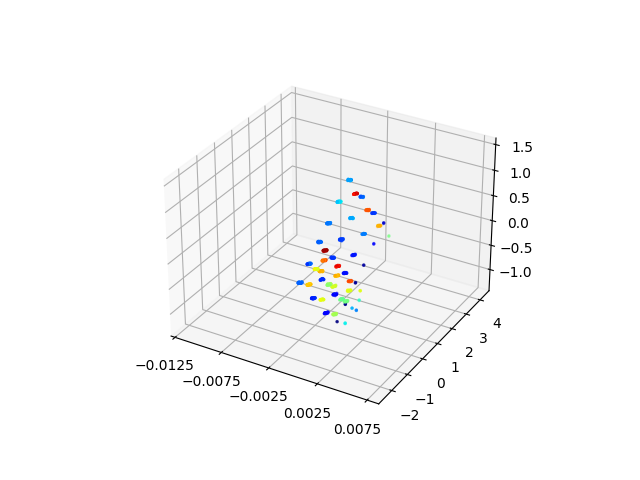} % trim = {left, bottom, right, top}
 \includegraphics[height=5.6cm,trim={3.5cm 1.1cm 2.74cm 1.7cm},clip]{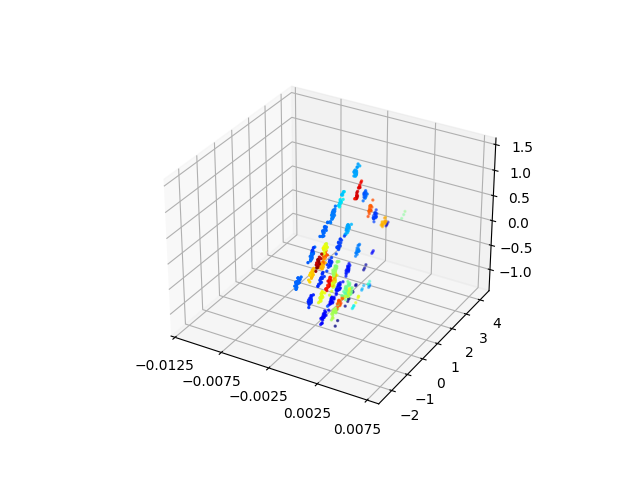}
  \includegraphics[height=5.6cm,trim={3.5cm 1.1cm 2.74cm 1.7cm},clip]{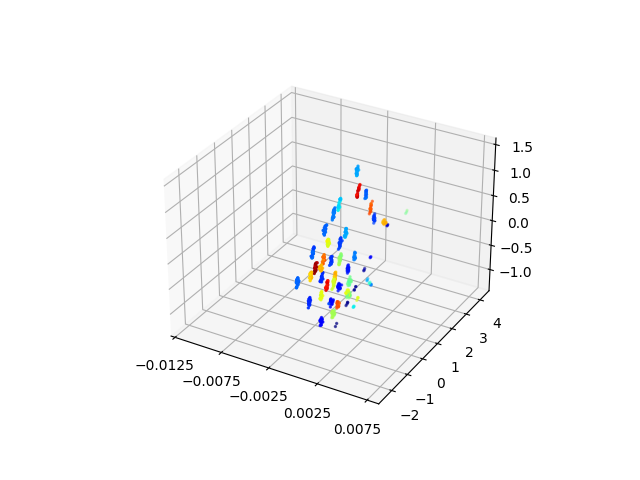}
    \includegraphics[height=5.6cm,trim={3.5cm 1.1cm 2.74cm 1.7cm},clip]{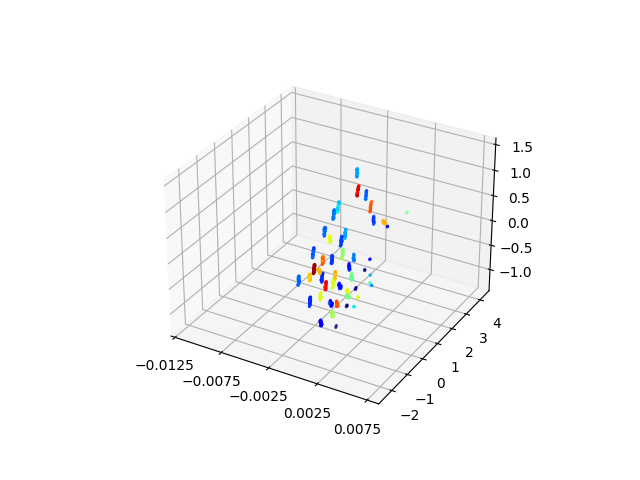}
        \includegraphics[height=5.6cm,trim={3.5cm 1.1cm 2.74cm 1.7cm},clip]{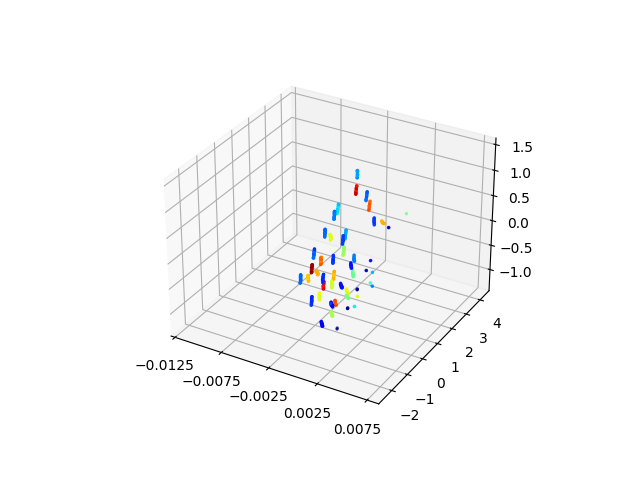}
  \includegraphics[height=5.6cm,trim={3.5cm 1.1cm 2.74cm 1.7cm},clip]{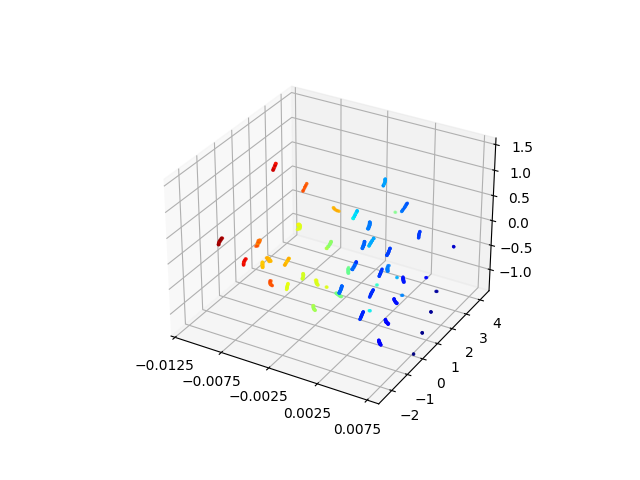}
	\caption{Reshaped sequence  of an {\it Energy} standard training split through module D$_1.$ Each plot corresponds to the first three principal components of $\vect{z}^1(t)$ at a fixed time $t,$ with the true responses color coded.  Starting at top and viewing left to right, $t=0,$ 0.2, 0.4, 0.6, 0.8, and 1, respectively.} 
	\label{fig:Energy}
\end{figure}

The lowest average test RMSE in each standard splits column and each gap splits column in Tables \ref{tab:experiments_standard_gap1} and \ref{tab:experiments_standard_gap2} is bolded, determined in the {\it Kin8nm} and {\it Naval} standard split columns and the {\it Kin8nm} and {\it Wine Red} gap split columns by a comparison of results in higher decimal precision.
Result values in these four columns from sources with only 2-digit decimal representation that cannot be confirmed as lower or higher than these lowest values are bolded as well.  Examples of final reshaped sequences through module D$_1$ of standard training splits of {\it Kin8nm}, {\it Concrete}, and {\it Energy} are illustrated in Figures \ref{fig:Kin8nm}, \ref{fig:Concrete}, and \ref{fig:Energy}, respectively. 
In each figure plot, data point locations represent the first three principal components of $\vect{z}^1(t)$ at a fixed time $t,$ and color coding represents the true responses.  Each figure contains six plots, corresponding to $t=0,$ 0.2, 0.4, 0.6, 0.8, and 1, respectively.

%%%%%%%%%%%%%%%%%%%%%%%%%%%%%%%%%%%%%

\clearpage 

\setcounter{table}{0}
\renewcommand{\thetable}{3\Alph{table}}%\arabic{table}}
\begin{table}[hb!] 
	\caption{Average test RMSE $\pm$ 1 standard error (best values in bold).}
	\label{tab:experiments_standard_gap1}
	\begin{threeparttable}
	\centering\small

	\begin{adjustbox}{center}	
		\begin{tabular}{@{}>{\bfseries}l*{5}{r}@{}}
			&\multicolumn{5}{@{}c@{}}{\bfseries UCI Standard Splits (Top) and Gap Splits (Bottom)}\\
		    \addlinespace[1pt]
    		\cline{2-6}
    		\addlinespace[3pt]
			 & \textbf{\textit{Concrete}}& \textbf{\textit{Energy}} & \textbf{\textit{Kin8nm}}  & \textbf{\textit{Naval}} & \textbf{\textit{Power}} \\ %& \textbf{Protein}& \textbf{Wine Red} & \textbf{Yacht}  & \textbf{Year}   \\
			       & $N_T=1030$ & $N_T=768$ & $N_T=8192$ & $N_T=11934$ & $N_T=9568$ \\ %& $N_T=45730$ & $N_T=1599$ & $N_T=308$ & $N_T=515345$ \\
			       & $d_X=8$ & $d_X=8$ & $d_X=8$ & $d_X=16$ & $d_X=4$ \\ %& $d_X=9$ & $d_X=11$ & $d_X=6$ & $d_X=90$  \\
			 Model  & $d_Y=1$ & $d_Y=1$ & $d_Y=1$ & $d_Y=1$ & $d_Y=1$ \\ % & $d_Y=1$ & $d_Y=1$ & $d_Y=1$ & $d_Y=1$  \\
			\midrule%[0.75pt]
			ADA     &$4.93 \pm 0.13$  & $0.50 \pm 0.01$ & \resizebox{0.62in}{\height}{$\mathbf{0.07 \pm 0.00}$}\tnote{1}  & \resizebox{0.62in}{\height}{$\mathbf{0.00 \pm 0.00}$}\tnote{1} & \resizebox{0.62in}{\height}{$\mathbf{3.36 \pm 0.05}$} \\ %& $3.33 \pm 0.05$ & \resizebox{0.62in}{\height}{$\mathbf{0.59 \pm 0.01}$} & $0.72 \pm 0.06$ & \resizebox{0.60in}{\height}{$\mathbf{8.84 \pm \mbox{NA}}$}\\  
			A       &$10.31 \pm 0.14$  & $3.06 \pm 0.05$ & $0.20 \pm 0.00$  & $0.01 \pm 0.00$ & $4.61 \pm 0.03$ \\ %& $5.21 \pm 0.02$ & $0.65 \pm 0.01$ & $8.95 \pm 0.27$ & $9.51\pm\mbox{NA}$  \\ 
			DNN-1    &$5.02 \pm 0.14$  & $0.53 \pm 0.01$ & $0.08 \pm 0.00$  & $0.00 \pm 0.00$ & $3.98 \pm 0.04$ \\% & $4.35 \pm 0.04$ & $0.66 \pm 0.01$ & $0.95 \pm 0.07$ & $8.96\pm\mbox{NA}$ \\
			DNN-2    &$4.47 \pm 0.13$  & $0.51 \pm 0.01$ & $0.08 \pm 0.00$  & $0.00 \pm 0.00$ & $3.70 \pm 0.04$ \\ %& $3.79 \pm 0.02$ & $0.73 \pm 0.02$ & $0.92 \pm 0.08$ & $9.78\pm\mbox{NA}$ \\
			DNN-3    &\resizebox{0.62in}{\height}{$\mathbf{4.46 \pm 0.12}$}  & \resizebox{0.62in}{\height}{$\mathbf{0.43 \pm 0.02}$} & $0.08 \pm 0.00$  & $0.00 \pm 0.00$ & $3.63 \pm 0.05$ \\ %& $3.83 \pm 0.03$ & $0.66 \pm 0.02$ & $1.27 \pm 0.15$ & $10.73\pm\mbox{NA}$ \\
			DNN-5    &$4.71 \pm 0.15$  & $0.46 \pm 0.02$ & $0.08 \pm 0.00$  & $0.00 \pm 0.00$ & $3.67 \pm 0.05$ \\ %& $3.70 \pm 0.01$ & $0.64 \pm 0.01$ & $1.22 \pm 0.12$ & $10.20\pm\mbox{NA}$\\
			DNN-10   &$4.64 \pm 0.14$  & $0.54 \pm 0.08$ & $0.08 \pm 0.00$  & $0.01 \pm 0.00$ & $3.59 \pm 0.04$ \\ %& $4.73 \pm 0.52$ & $0.70 \pm 0.02$ & $3.32 \pm 1.02$ & $10.31\pm\mbox{NA}$\\
% 			\midrule[0.05pt] % Hernandez-Lobato & Adams, 2015: 
% 			&\citet{pmlr-v37-hernandez-lobatoc15}
			% VI      &$7.13 \pm 0.12$  & $2.65 \pm 0.08$ & $0.10 \pm 0.00$  & $0.01 \pm 0.00$ & $4.33 \pm 0.04$ & $4.84 \pm 0.03$ & $0.65 \pm 0.01$ & $6.89 \pm 0.67$ & $9.03 \pm \mbox{NA}$\\
			% BP      &$5.98 \pm 0.22$  & $1.10 \pm 0.07$ & $0.09 \pm 0.00$  & $0.00 \pm 0.00$ & $4.18 \pm 0.04$ & $4.54 \pm 0.03$ & $0.65 \pm 0.01$ & $1.18 \pm 0.16$ & $8.93\pm\mbox{NA}$\\
			% BP-2      &$5.40 \pm 0.13$  & $0.68 \pm 0.04$ & $0.07 \pm 0.00$  & $0.00 \pm 0.00$ & $4.22 \pm 0.07$ & $4.19 \pm 0.03$ & $0.65 \pm 0.01$ & $1.54 \pm 0.19$ & $8.98\pm\mbox{NA}$\\
			BP-3      &$5.57 \pm 0.13$  & $0.63 \pm 0.03$ & $0.07 \pm 0.00$  & $0.00 \pm 0.00$ & $4.11 \pm 0.04$ \\ %& $4.01 \pm 0.03$ & $0.65 \pm 0.01$ & $1.11 \pm 0.09$ & $8.93\pm\mbox{NA}$\\
			BP-4      &$5.53 \pm 0.14$  & $0.67 \pm 0.03$ & $0.07 \pm 0.00$  & $0.00 \pm 0.00$ & $4.18 \pm 0.06$ \\ %& $3.96 \pm 0.01$ & $0.65 \pm 0.02$ & $1.27 \pm 0.13$ & $9.05\pm\mbox{NA}$\\
			% PBP     &$5.67 \pm 0.09$  & $1.80 \pm 0.05$ & $0.10 \pm 0.00$  & $0.01 \pm 0.00$ & $4.12 \pm 0.03$ & $4.73 \pm 0.01$ & $0.64 \pm 0.01$ & $1.02 \pm 0.05$ & $8.88\pm\mbox{NA}$\\
			PBP-2     &$5.24 \pm 0.12$  & $0.90 \pm 0.05$ & $0.07 \pm 0.00$  & $0.00 \pm 0.00$ & $4.03 \pm 0.03$ \\ %& $4.25 \pm 0.02$ & $0.64 \pm 0.01$ & $0.85 \pm 0.05$ & $8.92\pm\mbox{NA}$\\
			PBP-3     &$5.73 \pm 0.11$  & $1.24 \pm 0.06$ & $0.07 \pm 0.00$  & $0.01 \pm 0.00$ & $4.07 \pm 0.04$ \\ %& $4.09 \pm 0.03$ & $0.64 \pm 0.01$ & $0.89 \pm 0.10$ & $8.87\pm\mbox{NA}$\\
			% PBP-4     &$5.96 \pm 0.16$  & $1.18 \pm 0.06$ & $0.08 \pm 0.00$  & $0.00 \pm 0.00$ & $4.08 \pm 0.04$ & $3.97 \pm 0.04$ & $0.64 \pm 0.01$ & $1.71 \pm 0.23$ & $8.93\pm\mbox{NA}$\\
% 			\midrule[0.05pt] % Gal & Ghahramani, 2016: (standard deviations instead of standard errors)
% 			&\citet{pmlr-v48-gal16}
% 			Dropout-TS &$5.23 \pm 0.53$  & $1.66 \pm 0.19$ & $0.10 \pm 0.00$  & $0.01 \pm 0.00$ & $4.02 \pm 0.18$ & $4.36 \pm 0.04$ & $0.62 \pm 0.04$ & $1.11 \pm 0.38$ & $8.85\pm\mbox{NA}$\\
			%                                          (corrected to standard errors in Mukhoti paper):
	        Dropout-TS &$5.23 \pm 0.12$  & $1.66 \pm 0.04$ & $0.10 \pm 0.00$  & $0.01 \pm 0.00$ & $4.02 \pm 0.04$ \\ %& $4.36 \pm 0.01$ & $0.62 \pm 0.01$ & $1.11 \pm 0.09$ & $8.85\pm\mbox{NA}$\\

		Dropout-C   &$4.93 \pm 0.14$  & $1.08 \pm 0.03$ & $0.09 \pm 0.00$  & \resizebox{0.62in}{\height}{$\mathbf{0.00 \pm 0.00}$} & $4.00 \pm 0.04$ \\ %& $4.27 \pm 0.01$ & $0.61 \pm 0.01$ & $0.70 \pm 0.05$ & $--$\\
		Dropout-G  &$4.82 \pm 0.16$  & $0.54 \pm 0.06$ & $0.08 \pm 0.00$  & \resizebox{0.62in}{\height}{$\mathbf{0.00 \pm 0.00}$} & $4.01 \pm 0.04$ \\%& $4.27 \pm 0.02$ & $0.62 \pm 0.01$ & $0.67 \pm 0.05$ & $--$\\
% 		VMG         &$4.89 \pm 0.12$  & $0.54 \pm 0.02$ & $0.08 \pm 0.00$  & \resizebox{0.62in}{\height}{$\mathbf{0.00 \pm 0.00}$} & $4.04 \pm 0.04$ & $4.13 \pm 0.02$ & $0.63 \pm 0.01$ & $0.71 \pm 0.05$ & $--$\\
% 		HS-BNN      &$5.66 \pm 0.41$  & $1.99 \pm 0.34$ & $0.08 \pm 0.00$  & \resizebox{0.62in}{\height}{$\mathbf{0.00 \pm 0.00}$} & $4.03 \pm 0.15$ & $4.39 \pm 0.04$ & $0.63 \pm 0.04$ & $1.58 \pm 0.23$ & $--$\\
% 		PBP-MV      &$5.08 \pm 0.14$  & $0.45 \pm 0.01$ & $0.07 \pm 0.00$  & \resizebox{0.62in}{\height}{$\mathbf{0.00 \pm 0.00}$} & $3.91 \pm 0.04$ & $3.94 \pm 0.02$ & $0.64 \pm 0.01$ & $0.81 \pm 0.06$ & $--$\\

        % \midrule[0.05pt]% Mishkin et al., 2018:
        BBB   &$6.16 \pm 0.13$  & $0.97 \pm 0.09$ & $0.08 \pm 0.00$  & \resizebox{0.62in}{\height}{$\mathbf{0.00 \pm 0.00}$} & $4.21 \pm 0.03$ \\ %& $--$ & $0.64 \pm 0.01$ & $1.13 \pm 0.06$ & $--$\\
		SLANG  &$5.58 \pm 0.19$  & $0.64 \pm 0.03$ & $0.08 \pm 0.00$  & \resizebox{0.62in}{\height}{$\mathbf{0.00 \pm 0.00}$} & $4.16 \pm 0.04$ \\ %& $--$ & $0.65 \pm 0.01$ & $1.08 \pm 0.06$ & $--$\\

        DUN (MLP)   &$4.57 \pm 0.16$  & $0.95 \pm 0.11$ & $0.08 \pm 0.00$  & ${0.00 \pm 0.00}$ & $3.67 \pm 0.06$ \\ %& $3.41 \pm 0.03$ & $0.63 \pm 0.01$ & $2.47 \pm 0.19$ & $--$\\
        Dropout     &$4.61 \pm 0.13$  & $0.57 \pm 0.05$ & $0.07 \pm 0.00$  & ${0.00 \pm 0.00}$ & $3.82 \pm 0.08$ \\ %& $3.43 \pm 0.03$ & $0.64 \pm 0.01$ & $0.88 \pm 0.09$ & $--$\\
        Ensemble    &$4.55 \pm 0.13$  & $0.51 \pm 0.02$ & $0.30 \pm 0.22$  & ${0.00 \pm 0.00}$ & $3.44 \pm 0.05$ \\ %& \resizebox{0.62in}{\height}{$\mathbf{3.26 \pm 0.03}$} & $1.93 \pm 1.28$ & $1.43 \pm 0.11$ & $--$\\
        $\mathbf{\mathcal{L}_{\beta-NLL}(\beta=0.75)}$  & $5.67 \pm 0.16$ &$--$ & $0.08 \pm 0.00$   &$--$ & $4.04 \pm 0.03$ \\ %& $4.28 \pm 0.01$ & $0.64 \pm 0.01$   & $1.97 \pm 0.23$ &  $--$\\
        % $\mathbf{\mathcal{L}_{\beta-NLL}(\beta=1.0)}$   & $5.55 \pm 0.17$ &\cgr $1.54 \pm 0.12$ & $0.08 \pm 0.00$   &\cgr $0.00 \pm 0.00$ & $4.06 \pm 0.04$ & $4.31 \pm 0.02$ & $0.64 \pm 0.01$   & $2.08 \pm 0.25$ &  $--$\\
        % $\mathbf{\mathcal{L}_{MM}}$                     & $6.28 \pm 0.18$ &\cgr $2.19 \pm 0.06$ & $0.08 \pm 0.00$   &\cgr $0.00 \pm 0.00$ & $4.07 \pm 0.04$ & $4.32 \pm 0.03$ & $0.65 \pm 0.01$   & $3.02 \pm 0.31$ &  $--$\\
        $\mathbf{\mathcal{L}_{MSE}}$                    & $4.96 \pm 0.14$ &$--$ & $0.08 \pm 0.00$   & $--$ & $4.01 \pm 0.04$ \\ %& $4.28 \pm 0.03$ & $0.63 \pm 0.01$   & $0.78 \pm 0.06$ &  $--$\\
        % Student-t                                       & $5.82 \pm 0.13$ &\cgr $2.26 \pm 0.08$ & $0.09 \pm 0.00$   &\cgr $0.00 \pm 0.00$ & $4.02 \pm 0.04$ & $4.76 \pm 0.11$ & $0.64 \pm 0.01$   & $1.34 \pm 0.14$ &  $--$\\
        % xVAMP                                           & $5.44 \pm 0.14$ &\cgr $1.87 \pm 0.07$ & $0.08 \pm 0.00$   &\cgr $0.00 \pm 0.00$ & $4.03 \pm 0.04$ & $4.38 \pm 0.02$ & $0.64 \pm 0.01$   & $0.99 \pm 0.10$ &  $--$\\
        % xVAMP*                                          & $5.35 \pm 0.16$ &\cgr $2.00 \pm 0.06$ & $0.08 \pm 0.00$   &\cgr $0.00 \pm 0.00$ & $4.03 \pm 0.04$ & $4.31 \pm 0.01$ & $0.63 \pm 0.01$   & $1.13 \pm 0.15$ &  $--$\\
        % VBEM                                            & $5.21 \pm 0.13$ &\cgr $1.29 \pm 0.07$ & $0.08 \pm 0.00$   &\cgr $0.00 \pm 0.00$ & $4.09 \pm 0.03$ & $4.31 \pm 0.00$ & $0.64 \pm 0.01$   & $1.66 \pm 0.19$ &  $--$\\
        VBEM*                                           & $5.17 \pm 0.13$ &$--$ & $0.08 \pm 0.00$   &$--$ & $4.02 \pm 0.04$ \\ %& $4.35 \pm 0.04$ & $0.63 \pm 0.01$   & \resizebox{0.62in}{\height}{$\mathbf{0.65 \pm 0.04}$} &  $--$\\    

\addlinespace[0.5pt]
% \hdashline
\arrayrulecolor[gray]{0.65}\hline%{lightgray}\hline
\addlinespace[2pt]
% \arrayrulecolor{lightgray}\midrule

			ADA     &$7.53 \pm 0.29$   & $3.61 \pm 1.23$ & \resizebox{0.62in}{\height}{$\mathbf{0.07 \pm 0.00}$}\tnote{1}  & $0.02 \pm 0.00$ & $5.53 \pm 0.58$ \\ %& $5.10 \pm 0.18$ & $0.68 \pm 0.01$ & $1.03 \pm 0.14$\\
			A       &$10.75 \pm 0.29$  & $3.96 \pm 0.36$ & $0.20 \pm 0.00$  & $0.03 \pm 0.00$ & $4.47 \pm 0.08$ \\ %& $5.34 \pm 0.04$ & $0.64 \pm 0.01$ & $9.24 \pm 0.31$  \\ 
			DNN-1    &$7.53 \pm 0.34$  & $4.59 \pm 1.75$ & $0.08 \pm 0.00$  & $0.03 \pm 0.00$ & $4.33 \pm 0.13$ \\ %& $5.08 \pm 0.09$ & $0.72 \pm 0.01$ & $2.33 \pm 0.29$  \\
			DNN-2    &$7.45 \pm 0.31$  & $3.77 \pm 1.34$ & $0.08 \pm 0.00$  & $0.03 \pm 0.00$ & $5.17 \pm 0.33$ \\ %& $5.56 \pm 0.20$ & $0.81 \pm 0.01$ & $3.40 \pm 0.64$ \\
			DNN-3    &$7.44 \pm 0.23$  & $3.93 \pm 1.42$ & $0.08 \pm 0.00$  & $0.03 \pm 0.00$ & $5.67 \pm 0.33$ \\ %& $5.95 \pm 0.21$ & $0.73 \pm 0.01$ & $3.53 \pm 0.59$ \\
			DNN-5    &$7.28 \pm 0.16$  & $3.23 \pm 1.08$ & $0.08 \pm 0.00$  & $0.03 \pm 0.00$ & $5.78 \pm 0.41$ \\ %& $5.85 \pm 0.24$ & $0.74 \pm 0.01$ & $3.29 \pm 0.54$ \\
			DNN-10   &$8.41 \pm 1.05$  & $5.98 \pm 1.42$ & $0.08 \pm 0.00$  & $0.02 \pm 0.00$ & $5.56 \pm 0.45$ \\ %& $6.04 \pm 0.21$ & $0.76 \pm 0.01$ & $3.95 \pm 0.71$ \\
    MAP-1           &$7.79 \pm 0.18$    & \resizebox{0.62in}{\height}{$\mathbf{2.83 \pm 0.99}$}   & $0.09 \pm 0.01$  & $0.02 \pm 0.00$ & \resizebox{0.62in}{\height}{$\mathbf{4.24 \pm 0.12}$} \\ %& $5.16 \pm 0.04$ & \resizebox{0.62in}{\height}{$\mathbf{0.63 \pm 0.01}$} & $1.31 \pm 0.14$ \\
    MAP-2           &$7.78 \pm 0.23$    & $3.70 \pm 1.33$   & $0.08 \pm 0.00$  & $0.03 \pm 0.00$ & $4.33 \pm 0.18$ \\ %& $5.07 \pm 0.06$ & \resizebox{0.62in}{\height}{$\mathbf{0.63 \pm 0.01}$} & $1.05 \pm 0.09$ \\
    % MAP-1 NL        &$7.68 \pm 0.23$    & $3.09 \pm 1.17$   & $0.09 \pm 0.01$  & $0.02 \pm 0.00$ & $4.25 \pm 0.09$ & $5.13 \pm 0.05$ & \resizebox{0.62in}{\height}{$\mathbf{0.63 \pm 0.01}$} & $1.28 \pm 0.14$ \\
    MAP-2 NL        &$7.44 \pm 0.17$    & $3.48 \pm 1.21$   & \resizebox{0.62in}{\height}{$\mathbf{0.07 \pm 0.00}$}  & $0.03 \pm 0.00$ & $4.27 \pm 0.08$ \\ %& $5.08 \pm 0.06$ & \resizebox{0.62in}{\height}{$\mathbf{0.63 \pm 0.01}$} & \resizebox{0.62in}{\height}{$\mathbf{1.01 \pm 0.09}$} \\ 
    % Reg-1 NL        &$8.21 \pm 0.48$    & $4.24 \pm 2.11$   & $0.08 \pm 0.00$  & \resizebox{0.62in}{\height}{$\mathbf{0.01 \pm 0.00}$} & $5.17 \pm 0.60$ & $5.23 \pm 0.12$ & $0.66 \pm 0.02$ & $1.24 \pm 0.11$ \\
    % Reg-2 NL        &$8.27 \pm 0.39$    & $3.83 \pm 1.49$   & \resizebox{0.62in}{\height}{$\mathbf{0.07 \pm 0.00}$}  & \resizebox{0.62in}{\height}{$\mathbf{0.01 \pm 0.00}$} & $5.23 \pm 0.43$ & $5.33 \pm 0.16$ & $0.64 \pm 0.01$ & $1.22 \pm 0.13$ \\
    % BN(ML)-1 NL     &$7.69 \pm 0.51$    & $4.15 \pm 1.64$   & $0.09 \pm 0.00$  & \resizebox{0.62in}{\height}{$\mathbf{0.01 \pm 0.00}$} & $4.49 \pm 0.15$ & $5.27 \pm 0.12$ & \resizebox{0.62in}{\height}{$\mathbf{0.63 \pm 0.01}$} & $1.15 \pm 0.11$ \\ 
    BN(ML)-2 NL     &$7.33 \pm 0.36$    & $4.10 \pm 1.64$   & $0.08 \pm 0.00$  & \resizebox{0.62in}{\height}{$\mathbf{0.01 \pm 0.00}$} & $5.17 \pm 0.28$ \\ %& $5.37 \pm 0.17$ & $0.64 \pm 0.01$ & $1.31 \pm 0.16$ \\
        DUN         &$7.20 \pm 0.18$  & $2.94 \pm 0.67$     & $0.08 \pm 0.00$       & $0.02 \pm 0.00$       & $4.30 \pm 0.09$       \\ %& $5.21 \pm 0.35$       & $0.70 \pm 0.01$   & $1.85 \pm 0.17$ \\
        % DUN (MLP)   &$7.46 \pm 0.21$  & $3.61 \pm 0.88$     & $0.08 \pm 0.00$       & $0.02 \pm 0.00$       & $4.58 \pm 0.08$       & $5.10 \pm 0.24$       & $0.69 \pm 0.01$   & $1.85 \pm 0.14$\\
        Dropout     &$7.06 \pm 0.21$  & $2.87 \pm 0.50$     & $0.07 \pm 0.00$       & $0.03 \pm 0.00$       & $4.69 \pm 0.07$       \\ %& $5.13 \pm 0.28$       & $0.66 \pm 0.01$   & $2.29 \pm 0.47$ \\
        Ensemble    &\resizebox{0.62in}{\height}{$\mathbf{6.85 \pm 0.18}$}  & $3.36 \pm 0.83$     & $1.63 \pm 0.99$       & $0.02 \pm 0.00$       & $4.37 \pm 0.09$       \\ %& \resizebox{0.62in}{\height}{$\mathbf{4.80 \pm 0.27}$}       & $0.67 \pm 0.01$   & $1.84 \pm 0.19$ \\
        MFVI        &$7.55 \pm 0.19$  & $8.61 \pm 2.10$     & $0.10 \pm 0.01$       & $0.03 \pm 0.01$       & $4.68 \pm 0.16$       \\ %& $5.12 \pm 0.13$       & \resizebox{0.62in}{\height}{$\mathbf{0.63 \pm 0.01}$}\tnote{1}   & $1.84 \pm 0.16$ \\
        % SGD         &$7.37 \pm 0.19$  & $3.06 \pm 0.64$     & $0.09 \pm 0.00$       & $0.02 \pm 0.00$       & $4.62 \pm 0.08$       & $5.17 \pm 0.28$       & $0.73 \pm 0.02$   & $2.21 \pm 0.18$ \\
        
		\arrayrulecolor{black}\bottomrule
		\end{tabular}
		\end{adjustbox}
% 		\vspace{0.3cm}
        \begin{tablenotes}\footnotesize
        \item[1] Lowest results in their respective column sections, when compared in higher decimal precision.  \\
        Also bolded are 2-digit decimal literature results that cannot be confirmed as lower or higher \\
        than these lowest results.  
        \end{tablenotes}
    \end{threeparttable}
\end{table}

%%%%%%%%%%%%%%%%%%%%%%%  splitting table in two

\clearpage 

\begin{table}[hb!] 
	\caption{Average test RMSE $\pm$ 1 standard error (best values in bold).}
	\label{tab:experiments_standard_gap2}
	\begin{threeparttable}
	\centering\small

	\begin{adjustbox}{center}	
		\begin{tabular}{@{}>{\bfseries}l*{6}{r}@{}}
			&\multicolumn{6}{@{}c@{}}{\bfseries UCI Standard Splits (Top) and Gap Splits (Bottom)}\\
		    \addlinespace[1pt]
    		\cline{2-7}
    		\addlinespace[3pt]
            &&& \textbf{\textit{Protein}}& \textbf{\textit{Wine Red}} & \textbf{\textit{Yacht}}  & \textbf{\textit{Year}}   \\
			&&& $N_T=45730$ & $N_T=1599$ & $N_T=308$ & $N_T=515345$ \\
			&&& $d_X=9$ & $d_X=11$ & $d_X=6$ & $d_X=90$  \\
			 Model  &&& $d_Y=1$ & $d_Y=1$ & $d_Y=1$ & $d_Y=1$  \\
			\midrule%[0.75pt]
			ADA     &&& $3.33 \pm 0.05$ & \resizebox{0.62in}{\height}{$\mathbf{0.59 \pm 0.01}$} & $0.72 \pm 0.06$ & \resizebox{0.60in}{\height}{$\mathbf{8.84 \pm \mbox{NA}}$}\\
			A        &&& $5.21 \pm 0.02$ & $0.65 \pm 0.01$ & $8.95 \pm 0.27$ & $9.51\pm\mbox{NA}$  \\ 
			DNN-1    &&& $4.35 \pm 0.04$ & $0.66 \pm 0.01$ & $0.95 \pm 0.07$ & $8.96\pm\mbox{NA}$ \\
			DNN-2    &&& $3.79 \pm 0.02$ & $0.73 \pm 0.02$ & $0.92 \pm 0.08$ & $9.78\pm\mbox{NA}$ \\
			DNN-3    &&& $3.83 \pm 0.03$ & $0.66 \pm 0.02$ & $1.27 \pm 0.15$ & $10.73\pm\mbox{NA}$ \\
			DNN-5    &&& $3.70 \pm 0.01$ & $0.64 \pm 0.01$ & $1.22 \pm 0.12$ & $10.20\pm\mbox{NA}$\\
			DNN-10   &&& $4.73 \pm 0.52$ & $0.70 \pm 0.02$ & $3.32 \pm 1.02$ & $10.31\pm\mbox{NA}$\\
			BP-3      &&& $4.01 \pm 0.03$ & $0.65 \pm 0.01$ & $1.11 \pm 0.09$ & $8.93\pm\mbox{NA}$\\
			BP-4      &&& $3.96 \pm 0.01$ & $0.65 \pm 0.02$ & $1.27 \pm 0.13$ & $9.05\pm\mbox{NA}$\\
			PBP-2     &&& $4.25 \pm 0.02$ & $0.64 \pm 0.01$ & $0.85 \pm 0.05$ & $8.92\pm\mbox{NA}$\\
			PBP-3     &&& $4.09 \pm 0.03$ & $0.64 \pm 0.01$ & $0.89 \pm 0.10$ & $8.87\pm\mbox{NA}$\\
	        Dropout-TS &&& $4.36 \pm 0.01$ & $0.62 \pm 0.01$ & $1.11 \pm 0.09$ & $8.85\pm\mbox{NA}$\\

		Dropout-C   &&& $4.27 \pm 0.01$ & $0.61 \pm 0.01$ & $0.70 \pm 0.05$ & $--$\\
		Dropout-G  &&& $4.27 \pm 0.02$ & $0.62 \pm 0.01$ & $0.67 \pm 0.05$ & $--$\\

        BBB   &&& $--$ & $0.64 \pm 0.01$ & $1.13 \pm 0.06$ & $--$\\
		SLANG  &&& $--$ & $0.65 \pm 0.01$ & $1.08 \pm 0.06$ & $--$\\

        DUN (MLP)   &&& $3.41 \pm 0.03$ & $0.63 \pm 0.01$ & $2.47 \pm 0.19$ & $--$\\
        Dropout     &&& $3.43 \pm 0.03$ & $0.64 \pm 0.01$ & $0.88 \pm 0.09$ & $--$\\
        Ensemble    &&& \resizebox{0.62in}{\height}{$\mathbf{3.26 \pm 0.03}$} & $1.93 \pm 1.28$ & $1.43 \pm 0.11$ & $--$\\
    
        $\mathbf{\mathcal{L}_{\beta-NLL}(\beta=0.75)}$  &&& $4.28 \pm 0.01$ & $0.64 \pm 0.01$   & $1.97 \pm 0.23$ &  $--$\\
        $\mathbf{\mathcal{L}_{MSE}}$                    &&& $4.28 \pm 0.03$ & $0.63 \pm 0.01$   & $0.78 \pm 0.06$ &  $--$\\
        VBEM*                                           &&& $4.35 \pm 0.04$ & $0.63 \pm 0.01$   & \resizebox{0.62in}{\height}{$\mathbf{0.65 \pm 0.04}$} &  $--$\\    

\addlinespace[0.5pt]
% \hdashline
\arrayrulecolor[gray]{0.65}\hline%{lightgray}\hline
\addlinespace[2pt]
% \arrayrulecolor{lightgray}\midrule

			ADA     &&& $5.10 \pm 0.18$ & $0.68 \pm 0.01$ & $1.03 \pm 0.14$\\
			A        &&& $5.34 \pm 0.04$ & $0.64 \pm 0.01$ & $9.24 \pm 0.31$  \\ 
			DNN-1    &&& $5.08 \pm 0.09$ & $0.72 \pm 0.01$ & $2.33 \pm 0.29$  \\
			DNN-2    &&& $5.56 \pm 0.20$ & $0.81 \pm 0.01$ & $3.40 \pm 0.64$ \\
			DNN-3    &&& $5.95 \pm 0.21$ & $0.73 \pm 0.01$ & $3.53 \pm 0.59$ \\
			DNN-5    &&& $5.85 \pm 0.24$ & $0.74 \pm 0.01$ & $3.29 \pm 0.54$ \\
			DNN-10   &&& $6.04 \pm 0.21$ & $0.76 \pm 0.01$ & $3.95 \pm 0.71$ \\

    MAP-1           &&& $5.16 \pm 0.04$ & \resizebox{0.62in}{\height}{$\mathbf{0.63 \pm 0.01}$} & $1.31 \pm 0.14$ \\
    MAP-2           &&& $5.07 \pm 0.06$ & \resizebox{0.62in}{\height}{$\mathbf{0.63 \pm 0.01}$} & $1.05 \pm 0.09$ \\
    MAP-2 NL        &&& $5.08 \pm 0.06$ & \resizebox{0.62in}{\height}{$\mathbf{0.63 \pm 0.01}$} & \resizebox{0.62in}{\height}{$\mathbf{1.01 \pm 0.09}$} \\ 
    BN(ML)-2 NL     &&& $5.37 \pm 0.17$ & $0.64 \pm 0.01$ & $1.31 \pm 0.16$ \\
    			%Concrete                   Energy                  Kin8nm                   Naval                  Power                   Protein                 Wine                    Yacht  
%DUN:       7.196 +/- 0.821 (0.18)  2.938 +/- 3.017 (0.67)  0.080 +/- 0.006 (0.00)  0.022 +/- 0.014 (0.00)  4.299 +/- 0.416 (0.09)  5.206 +/- 0.780 (0.35)  0.697 +/- 0.043 (0.01)  1.851 +/- 0.750 (0.17)
%DUNMLP:    7.461 +/- 0.948 (0.21)  3.606 +/- 3.927 (0.88)  0.078 +/- 0.005 (0.00)  0.021 +/- 0.007 (0.00)  4.584 +/- 0.356 (0.08)  5.101 +/- 0.526 (0.24)  0.692 +/- 0.041 (0.01)  1.852 +/- 0.623 (0.14)
%Dropout:   7.064 +/- 0.921 (0.21)  2.874 +/- 2.254 (0.50)  0.071 +/- 0.003 (0.00)  0.034 +/- 0.018 (0.00)  4.688 +/- 0.335 (0.07)  5.133 +/- 0.636 (0.28)  0.660 +/- 0.040 (0.01)  2.290 +/- 2.108 (0.47)
%Ensemble:  6.853 +/- 0.796 (0.18)  3.364 +/- 3.696 (0.83)  1.632 +/- 4.418 (0.99)  0.018 +/- 0.009 (0.00)  4.369 +/- 0.383 (0.09)  4.801 +/- 0.599 (0.27)  0.673 +/- 0.039 (0.01)  1.841 +/- 0.836 (0.19) 
%MFVI:      7.548 +/- 0.865 (0.19)  8.614 +/- 9.390 (2.10)  0.095 +/- 0.025 (0.01)  0.033 +/- 0.041 (0.01)  4.680 +/- 0.703 (0.16)  5.115 +/- 0.298 (0.13)  0.632 +/- 0.029 (0.01)  1.836 +/- 0.712 (0.16)
%SGD:       7.367 +/- 0.866 (0.19)  3.061 +/- 2.880 (0.64)  0.085 +/- 0.007 (0.00)  0.020 +/- 0.009 (0.00)  4.621 +/- 0.339 (0.08)  5.171 +/- 0.632 (0.28)  0.731 +/- 0.070 (0.02)  2.214 +/- 0.793 (0.18)
        DUN         &&& $5.21 \pm 0.35$       & $0.70 \pm 0.01$   & $1.85 \pm 0.17$ \\
        Dropout     &&& $5.13 \pm 0.28$       & $0.66 \pm 0.01$   & $2.29 \pm 0.47$ \\
        Ensemble    &&& \resizebox{0.62in}{\height}{$\mathbf{4.80 \pm 0.27}$}       & $0.67 \pm 0.01$   & $1.84 \pm 0.19$ \\
        MFVI        &&& $5.12 \pm 0.13$       & \resizebox{0.62in}{\height}{$\mathbf{0.63 \pm 0.01}$}\tnote{1}   & $1.84 \pm 0.16$ \\
        
		\arrayrulecolor{black}\bottomrule
		\end{tabular}
		\end{adjustbox}
% 		\vspace{0.3cm}
        \begin{tablenotes}\footnotesize
        \item[1] Lowest results in their respective column sections, when compared in higher \\
        decimal precision.  Also bolded are 2-digit decimal literature results that cannot \\ be 
        confirmed as lower or higher
        than these lowest results.  
        \end{tablenotes}
    \end{threeparttable}
\end{table}

% \clearpage

% \newpage

\clearpage

\newpage

\section{Discussion}\label{sec:discussion}

In comparison with all models tested on the same UCI standard splits in Tables \ref{tab:experiments_standard_gap1} and \ref{tab:experiments_standard_gap2}, our baseline ADA model performs well in general, ranking lowest in average test RMSE for five datasets ({\it Kin8nm}, {\it Naval}\,\footnote{Lowest result in higher decimal precision, with the exception of Dropout-C, Dropout-G, BBB, and SLANG, which are presented in the literature to 2-digits.}, \textit{Power, Wine Red}, and {\it Year}), second lowest for {\it Protein}, third lowest for {\it Energy}, and fourth lowest for {\it Yacht}.  Additionally, the use of subset training and dimensionality reduction for the {\it Year} experiment ensures experiment tractability and significantly reduces its running time.  The DNN-3 model outperforms all models for the {\it Concrete} and {\it Energy} datasets, and Ensemble and VBEM* have the lowest average RMSEs for {\it Protein} and {\it Yacht}, respectively. In comparison with all models tested on different standard splits in Tables \ref{tab:all_literature_experiments_standard1} and \ref{tab:all_literature_experiments_standard2}, our ADA model maintains a similar performance level, as those literature model experiments---with the exception of the {\it Year} dataset on which PBP-MV performs best---primarily show improved performance for the smaller {\it Energy} and {\it Yacht} datasets on which ADA did not have top performance.   
Sample
% reshaped sequences 
D$_1$ deformation sequences  
in Figures \ref{fig:Kin8nm}-\ref{fig:Energy} from these experiments demonstrate the smooth invertibility of data transformations through the D modules of our model.%, an advantage our models over ``black-box" or non-invertible machine learning constructs. 

While the standard splits are useful for testing a model's ability to fit data, the gap splits can test in a sense how well a model generalizes to out-of-distribution data.  A robust model will simultaneously perform well on the standard splits and not critically fail on the gap splits.  Our ADA model 
demonstrates above average performance overall in ranking comparisons with all models tested on the UCI gap splits in Tables \ref{tab:experiments_standard_gap1} and \ref{tab:experiments_standard_gap2}, with no excessively poor predictions.  Specifically, our model's average test RMSE ranks lowest for {\it Kin8nm}\footnote{Lowest result in higher decimal precision, with the exception of MAP-2 NL, presented in the literature to 2-digits.}, second lowest for {\it Yacht}, fifth lowest for {\it Protein}, twelfth lowest (fourth highest) for {\it Power}, and approximately center ranking performance for {\it Concrete}, {\it Energy}, {\it Naval}, and {\it Wine Red}.

\section{Beyond the Baseline}\label{sec:complexity}

We test more complex model architectures beyond the ADA baseline 
in Table \ref{tab:additional_sequences}, using the {\it Airfoil} dataset in the UCI repository for the experiments. Performance is compared with DNNs, as the {\it Airfoil} dataset has similar size and dimensions to the {\it Concrete} and {\it Energy} datasets on which the DNN models perform best in Table \ref{tab:experiments_standard_gap1}.
We generate 10 randomized train-test splits (90\% train, 10\% test) and again evaluate prediction performance of the test splits. The  parameters of the diffeomorphic regression models 
in Table \ref{tab:additional_sequences} follow those in the experiments in Section \ref{sec:experiments}, including 
%the more complex diffeomorphic regression models which use 
the same A and D modules used in our ADA model, with two exceptions.  
First, for (AD)${\mbox{\textsuperscript{x}}}$A models with x sequential AD module pairs,
the affine costs of the inner A modules are the functions
\begin{equation}
U_q({A}) =  \left\| M - \mathrm{I}_{d}\right\|^2 = \mathrm{trace}((M - \mathrm{I}_{d})^T(M - \mathrm{I}_{d})),\quad q=1,\dots,m-1 \nonumber
\end{equation} 
for affine transformations from $\mathbb{R}^{{d_q}}$ to $\mathbb{R}^{{d_q}}$ (here, $d_{q+1} = d_q$) of the form
${A}(x) = M x + b,$ ${M\in\mathcal M_{d_q}(\mathbb{R}),}$ ${b\in\mathbb{R}^{d_q}.}$ 
%where $d_{q+1}=d_q,$ $q=1,\dots,m-1.$  
The initialization of these inner A modules is
\[
M_q = \mathrm{I}_{d_q}+\mathrm{diag}(w), w\sim\mathcal{N}(0,0.01^2)\in\mathbb{R}^{d_{q}},\hspace{0.1cm} b_q=0\in\mathbb{R}^{d_{q}},\quad q=1,\dots,m-1.
\] 
Second, for AD${\mbox{\textsuperscript{x}}}$A models with x sequential D modules, while the D modules have the same dimension, kernel type, and number of discretized time points as in our ADA model, the kernel widths of this sequence of $m$ D modules increase ($h_q\uparrow$), with respective values $\frac{1}{m+1},\frac{2}{m+1},\dots,\frac{m}{m+1}$, or decrease ($h_q\downarrow$), with respective values $\frac{m}{m+1},\frac{m-1}{m+1},\dots,\frac{1}{m+1}.$  All other experimental parameters in Table \ref{tab:additional_sequences} are the same as in Section \ref{sec:experiments}. %, with the exception of the use of float32 precision for all PyKeOps kernel computations.  
Starting with the original ADA model, the results show improved performance with increased model complexity, with the AD$^4$A ($h_q\downarrow$) 
model %, with four sequential D modules of decreasing kernel widths,
outperforming all models, including the DNNs.  
% The AD$^4$A ($h_q\downarrow$) model, with four sequential D modules of decreasing kernel widths, had the best result of all the models.  % why?
% , with an architecture reminiscent of the decreasing kernel sizes in CNNs for extracting low- to high-level patterns in the data.  
Note that to ensure these improved results are not the result of the increased number of time steps inherent in increasing the number of D modules, the ADA model is also run with increased $T_1$, as shown in Table \ref{tab:additional_sequences}.    

\setcounter{table}{3}
\renewcommand{\thetable}{\arabic{table}}
\begin{table}[ht!] 
	\caption{Average test RMSE $\pm$ 1 standard error (best values in bold).}
	\label{tab:additional_sequences}
	\begin{threeparttable}
    \centering\small
  
    \begin{adjustbox}{center}	
		\begin{tabular}{@{}>{\bfseries}l*{1}{r}@{}}
			 % Model & \textbf{Airfoil} \\
            & \textbf{\textit{Airfoil}} \\
            & $N_T=1503$ \\
            & $d_X = 5$ \\
            Model & $d_Y = 1$ \\
			\midrule%[0.75pt]
            ADA  &              $1.53 \pm 0.05$ \\        %  1.52763182  +/-  0.05482153   %   1.53797721  +/-  0.0534776
            % ADADA &             $1.56 \pm 0.06$ \\      %   1.56152464  +/-  0.05842726    ('reg','reg','reg')
            ADA ($T_1=20)$ &      $1.40 \pm 0.06$ \\      %  1.39602327  +/-  0.05744697   %   
            ADA ($T_1=30)$ &      $1.43 \pm 0.06$ \\      %  1.42934185  +/-  0.05639888   %   
            ADA ($T_1=40)$ &      $1.44 \pm 0.05$ \\      %  1.44009326  +/-  0.05298719   %   
            
            AD$^2$A ($h_q\uparrow$)  &   $1.38 \pm 0.05$ \\      %  1.38448486  +/-  0.05095385
            AD$^3$A ($h_q\uparrow$) &   $1.33 \pm 0.05$ \\      %  1.33313336  +/-  0.05145175
            AD$^4$A ($h_q\uparrow$) &  $1.32 \pm 0.05$ \\      %  1.31688033  +/-  0.04887550
            AD$^2$A ($h_q\downarrow$) &  $1.25 \pm 0.04$ \\      %  1.24543933  +/-  0.04419609
            AD$^3$A ($h_q\downarrow$) & $1.18 \pm 0.03$ \\      %  1.18052542  +/-  0.03214253
            AD$^4$A ($h_q\downarrow$) & \resizebox{0.62in}{\height}{$\mathbf{1.08 \pm 0.03}$} \\      %  1.08090947  +/-  0.02569199
            (AD)$^2$A  &             $1.41 \pm 0.05$ \\      %  1.41355553  +/-  0.05068450   
            (AD)$^3$A  &           $1.28 \pm 0.04$ \\      %  1.27757814  +/-  0.04247031 
            (AD)$^4$A &         $1.21 \pm 0.05$ \\      %  1.21343135  +/-  0.05103045  
            (AD)$^5$A &       $1.25 \pm 0.05$ \\      %  1.24880523  +/-  0.05385626
            % A &                 $4.85 \pm 0.08$ \\      %  4.84914988  +/-  0.07994997
            DNN-1 &             $2.02 \pm 0.06$ \\      %  2.01672911  +/-  0.06003411
            DNN-2 &             $1.36 \pm 0.05$ \\      %  1.35941270  +/-  0.04721632
            DNN-3 &             $1.25 \pm 0.05$ \\      %  1.24967696  +/-  0.04884235
            DNN-5 &             $1.16 \pm 0.04$ \\      %  1.15768015  +/-  0.03500883
            DNN-10 &            $1.79 \pm 0.60$ \\      %  1.78742693  +/-  0.59702501
            \bottomrule%[2pt]
		\end{tabular}
	\end{adjustbox}
    \end{threeparttable}
\end{table} 

We test the AD$^4$A ($h_q\downarrow$) model further in Table \ref{tab:ADDDDAexperiments_standard} on the standard and gap splits of six of the UCI datasets,  
using the same Section \ref{sec:experiments} experimental parameters. %, with the exception of float32 precision for the PyKeOps kernel computations for {\it Kin8nm, Power}, and {\it Wine Red}.
In comparison with the ADA results,  the \mbox{AD$^4$A ($h_q\downarrow$)} model  
shows improved ({\it Concrete}, {\it Kin8nm}, {\it Power}) and similar ({\it Energy}, {\it Wine Red}) results for the standard split experiments, and improved ({\it Concrete}, {\it Power}) and similar ({\it Kin8nm}, {\it Yacht}) results for the gap split experiments, with %the exception of 
slightly worse performance on average %overall 
on the remaining experiments---{\it Yacht} standard splits and {\it Energy} and {\it Wine Red} gap splits. 
% on which the model overfits. 
In particular, for the standard splits, the improved {\it Concrete} result is now comparable with the lowest result in DNN-3, and the improved {\it Kin8nm} result is the lowest\footnote{No caveats needed regarding lower decimal precision of literature results.} in both Tables \ref{tab:experiments_standard_gap1} and \ref{tab:all_literature_experiments_standard1}.  For the gap splits, the improved {\it Concrete} result does not improve its ranking in Table \ref{tab:experiments_standard_gap1} but the improved {\it Power} result does, bringing its performance up to approximately center ranking.

\begin{table}[ht!] 
	\caption{Average test RMSE $\pm$ 1 standard error (best values in bold).}
	\label{tab:ADDDDAexperiments_standard}
	\begin{threeparttable}
	\centering\small

	\begin{adjustbox}{center}	
		\begin{tabular}{@{}>{\bfseries}l*{6}{r}@{}}
			&\multicolumn{6}{@{}c@{}}{\bfseries UCI Standard Splits (Top) and Gap Splits (Bottom)}\\
		    % \addlinespace[1pt]
    		% \cline{2-7}
    		% \addlinespace[3pt]
            \addlinespace[3pt]
    		\cline{2-7}
    		\addlinespace[4pt]
			 % & \textbf{Concrete}& \textbf{Energy} & \textbf{Kin8nm}  & \textbf{Power} & \textbf{Wine Red} & \textbf{Yacht}   \\
			 %       & $N_T=1030$ & $N_T=768$ & $N_T=8192$ & $N_T=9568$ & $N_T=1599$ & $N_T=308$  \\
			 %       & $d_X=8$ & $d_X=8$ & $d_X=8$ & $d_X=4$ & $d_X=11$ & $d_X=6$  \\
			 % Model  & $d_Y=1$ & $d_Y=1$ & $d_Y=1$ & $d_Y=1$ & $d_Y=1$ & $d_Y=1$  \\
             Model  & \textbf{\textit{Concrete}}& \textbf{\textit{Energy}} & \textbf{\textit{Kin8nm}}  & \textbf{\textit{Power}} & \textbf{\textit{Wine Red}} & \textbf{\textit{Yacht}}   \\
			\midrule%[0.75pt]
			ADA     &$4.93 \pm 0.13$  & \resizebox{0.62in}{\height}{$\mathbf{0.50 \pm 0.01}$} & $0.07 \pm 0.00$   & $3.36 \pm 0.05$  & \resizebox{0.62in}{\height}{$\mathbf{0.59 \pm 0.01}$} & \resizebox{0.62in}{\height}{$\mathbf{0.72 \pm 0.06}$} \\  
			AD$^4$A ($h_q\downarrow$)    & \resizebox{0.62in}{\height}{$\mathbf{4.49 \pm 0.12}$}  & \resizebox{0.62in}{\height}{$\mathbf{0.50 \pm 0.01}$} & \resizebox{0.62in}{\height}{$\mathbf{0.06 \pm 0.00}$} & \resizebox{0.62in}{\height}{$\mathbf{3.24 \pm 0.05}$} & \resizebox{0.62in}{\height}{$\mathbf{0.59 \pm 0.01}$} & $0.78 \pm 0.05$ \\  
% \addlinespace[1pt]
% \hdashline
% \addlinespace[2pt]
\addlinespace[0.5pt]
\arrayrulecolor[gray]{0.65}\hline%{lightgray}\hline
\addlinespace[2pt]
           ADA     &$7.53 \pm 0.29$   & \resizebox{0.62in}{\height}{$\mathbf{3.61 \pm 1.23}$} & \resizebox{0.62in}{\height}{$\mathbf{0.07 \pm 0.00}$}  & $5.53 \pm 0.58$  & \resizebox{0.62in}{\height}{$\mathbf{0.68 \pm 0.01}$} & \resizebox{0.62in}{\height}{$\mathbf{1.03 \pm 0.14}$}\\   
   			AD$^4$A ($h_q\downarrow$)   & \resizebox{0.62in}{\height}{$\mathbf{7.47 \pm 0.29}$}  & $3.66 \pm 1.15$ & \resizebox{0.62in}{\height}{$\mathbf{0.07 \pm 0.00}$} & \resizebox{0.62in}{\height}{$\mathbf{4.60 \pm 0.23}$} & $0.69 \pm 0.01$ & \resizebox{0.62in}{\height}{$\mathbf{1.03 \pm 0.18}$}\\
            \arrayrulecolor{black}\bottomrule%[2pt]
	   \end{tabular}
    \end{adjustbox}
    \end{threeparttable}
\end{table} 
%                   Concrete                            Energy                          Kin8nm                          Power                               Wine Red                    Yacht
% Standard      4.48862588  +/-  0.12341351     0.50456005  +/-  0.01319319     0.06269159  +/-  0.00036410     3.23641957  +/-  0.05343320     0.58913344  +/-  0.01119520     0.77618960  +/-  0.05386197
% Gap           7.47213403  +/-  0.29052018     3.66420490  +/-  1.15052649     0.06613975  +/-  0.00100299     4.59702401  +/-  0.23435334     0.68522935  +/-  0.01050685     1.03275493  +/-  0.18123702           

\section{Summary}

We present a layered approach to multivariate regression using FineMorphs, 
a sequence model of affine and diffeomorphic transformations. 
Optimal control concepts from shape analysis
%foundational to both shape analysis and backpropagation 
are leveraged to optimally ``reshape" model states while learning. 
Diffeomorphisms (the model states) are generated by RKHS time-dependent vector fields (the control) calculated by Hamiltonian control theory, minimizing---along with the optimal affine parameters---a learning objective functional consisting of 
a %total 
kinetic energy term and affine and endpoint costs.  
% deformation, affine, and endpoint costs.
% As a kernel method, with its roots in kernels and shape generation and analysis, this approach provides a natural generative counterpoint
% to traditional deep CNNs but with the benefit of ``infinite depth" as well as smooth invertibility and thus ``explainability."  
In our setting, any arbitrary number and order of arbitrary affine and diffeomorphic transformations is possible, and diffeomorphisms can be generated as flows of sub-optimal vector fields to reduce dataset size and model complexity, while affine modules can scale data prior to diffeomorphic transforms as well as reduce (or increase) dimensionality.  
For both the optimal and sub-optimal vector fields cases,
a proof of the existence of a solution to the variational problem and a derivation of the necessary conditions for optimality are provided.
On standard UCI benchmark experiments, our baseline diffeomorphic regression model---ADA---performs favorably overall against state-of-the-art, hyperparameter-tuned deep BNNs and other models in the literature as well as DNNs in TensorFlow.  The computational intractability of the largest dataset in the experiments is successfully addressed with our model's dimensionality and dataset reduction capabilities, with good performance.  
A general trend of improved performance with increased model complexity is observed, in particular with ``coarse-to-fine" models containing multiple sequential diffeomorphisms of decreasing kernel sizes. 
% We test more complex model architectures beyond our ADA baseline, with multiple sequential diffeomorphisms of decreasing kernel sizes showing notable improvement on six of the UCI benchmarks, with the exception of the two smallest datasets on which overfitting is more likely to occur.  
% In particular, the improved performance of this model with a seemingly low- to high-level feature ``generation" architecture reinforces potential parallels with the low- to high-level feature extraction architectures of CNNs.  
% Our baseline model and, to a greater extent, our complex model, additionally demonstrate out-of-distribution robustness with reasonable performances in the experiments using the custom gap splits, in which the ``middle regions" of the data are assigned to the test sets.  
Additionally, our models  demonstrate out-of-distribution robustness with reasonable performances in experiments using custom gap splits, in which the ``middle regions" of the data are assigned to the test sets. 
In general, our diffeomorphic regression models provide an important degree of explainability and interpretability, even for the more complex architectures, as each diffeomorphic module in the model is a smooth invertible transformation of the data.   
% An advantage of our diffeomorphic regression models
% is that each D module is a smooth invertible transformation of the data, as illustrated by the D$_1$ deformation sequences from the baseline experiments in Figures \ref{fig:Kin8nm}-\ref{fig:Energy}. 
% Our ADA and AD${\mbox{\textsuperscript{x}}}$A models 
% % in Tables \ref{tab:additional_sequences} and \ref{tab:ADDDDAexperiments_standard} 
% in particular 
% % which maintain the same dimension throughout the inner model sequences, 
% have even greater advantages over ``black box" or non-invertible machine learning constructs such as the DNNs as they are smooth invertible transformations of the data throughout the entire inner model sequence, backwards traceable from the final D$_m$ module to the starting A$_0$ module.  
Future work includes further understanding of the theoretical basis, limitations, and advantages of our models; investigating dimensionality reduction and transformer architectures; and improving model run-time through adaptation of L-BFGS or stochastic optimization approaches.   

% \clearpage 

% \newpage 

\bibliography{diffeo_reg.bib}

\newpage 

\appendix
\setcounter{equation}{0}
\renewcommand{\theequation}{A.\arabic{equation}}%\arabic{table}}

\section{Existence of Solution to the FineMorph Variational Problem}\label{sec:existence}

% \begin{align}
% G(v_1,\dots,v_m,A_0,\dots,A_m) &= \sum_{q=1}^m \|v_q\|_{\mathcal{H}_q}^2 + \lambda \sum_{q=0}^{m}U_q({A}_q)\nonumber \\
% + \frac{1}{\sigma^2}\sum_{k=1}^N \Gamma_k(&\pi_r({A}_m\circ\bvarphi_m(1)\circ{A}_{m-1}\circ\cdots\circ\bvarphi_1(1)\circ{A}_0(\iota_n(x_k)))) \nonumber
% \end{align}
% minimized
% % over $v_1(\cdot),\dots,v_m(\cdot),$ and ${A}_0,\dots,{A}_m,$  
% over ${A}_0,\dots,{A}_m,$ and $v_q\in \mathcal{H}_q,$ $q=1,\dots,m,$
% such that $\bvarphi_q(t)$ satisfies  
% \begin{align}
% \partial_t\bvarphi_q(t)&=v_q(t)\circ\bvarphi_q(t),\quad t\in[0,1] \nonumber \\
% \bvarphi_q(0)&=\mbox{id}.\nonumber
% \end{align}

%with initial condition $\bvarphi_q(0)=\mbox{id}.$  
% A minimizer of this objective function 
% % in $v_1(\cdot),\dots,v_m(\cdot)$ 
% % $v_1,\dots,v_m$ 
% % with $v_q\in L^2([0,1],V_q)$ always exists for fixed ${A}_0,\dots,{A}_m$ 
% always exists, 
% as demonstrated in Appendix \ref{sec:existence}. 
% %under mild regularity conditions on the dependency of $\Gamma$ with respect to $\mathcal{T}_0$ (continuity in $x_1,\dots,x_N$ suffices). 

% Let $b_q=0\in \mathbb{R}^{d_q},$ $q=1,\dots,m.$
% \begin{align}
% G(v_1,\dots,v_m,M_0,\dots,M_m)&=\sum_{q=1}^m \|v_q\|_{\mathcal{H}_q}^2 + \lambda \sum_{q=0}^{m}\| M_q \|^2 \nonumber \\
% + \frac{1}{\sigma^2}\sum_{k=1}^N \Gamma_k(&\pi_r({M}_m\circ\bvarphi_m(1)\circ{M}_{m-1}\circ\cdots\circ\bvarphi_1(1)\circ{M}_0(\iota_n(x_k)))) \nonumber
% \end{align}
The variational problem in \eqref{eq:obj2} is to minimize 
\begin{align}
\begin{split}
G(v_1,\dots,v_m,M_0,\dots,M_m,b_0,\dots,b_m)= &\sum_{q=1}^m \|v_q\|_{\mathcal{H}_q}^2 + \lambda \sum_{q=0}^{m}\| M_q \|^2  \\
&+ \frac{1}{\sigma^2}\sum_{k=1}^N \Gamma_k(\pi_r(\zeta_k^{m+1}))
\label{eq:A1}
% & \qquad + \frac{1}{\sigma^2}\sum_{k=1}^N \Gamma_k(\pi_r({M}_m\bvarphi_m(1)({M}_{m-1}\bvarphi_{m-1}(1)(\cdots \nonumber \\
% & \qquad\qquad\qquad\qquad\qquad\qquad\qquad \cdots M_1\bvarphi_1(1)({M}_0(\iota_s(x_k))+b_0)\cdots)+b_{m-1})+b_m)) \nonumber 
\end{split}
\end{align}
over $v_q\in \mathcal{H}_q,$ $q=1,\dots,m,$ and ${M_q\in \mathcal{M}_{d_{q+1},d_q}(\mathbb{R}),}$ ${b_q\in \mathbb{R}^{d_{q+1}}},$ $q=0,\dots,m,$ 
subject to 
\begin{equation}
\label{eq:appendix:1.1}
\left\{
\begin{aligned}
\partial_t\bvarphi_{v_q}(t)&=v_q(t)\circ\bvarphi_{v_q}(t),\quad t\in[0,1] \\
\zeta_k^{q+1} &= M_q \bvarphi_{v_q}(1,\zeta_k^q) + b_q\\
\bvarphi_{v_q}(0)&=\mbox{id},\quad \zeta_k^{1} = M_0\iota_s(x_k) + b_0.
\end{aligned}
\right.
\end{equation}
% Forward states are ${\xi^0,\zeta^1,\xi^1,\zeta^2,\dots,\xi^m,\zeta^{m+1},}$ with ${\xi_k^q=\bvarphi_{v_q}(1)(\zeta_k^q),}$ ${\zeta_k^{q+1}={A}_{q}(\xi_k^{q}),}$ and initialization ${\xi^0_k = \iota_s(x_k).}$ Backpropagation states are ${\rho^1,\eta^1,\rho^2,\dots,\eta^m,\rho^{m+1},}$ with ${\eta_k^q=M_q^T\rho_k^{q+1},}$ ${\rho_k^q= \mathcal{F}_q(\eta_k^q),}$ and initialization ${\rho^{m+1}_k=-\frac{1}{\sigma^2}\iota_r(\nabla\Gamma_k(\pi_r(\zeta^{m+1}_k))).}$

$G$ is bounded from below and thus has an infimum $G_{\mathrm{min}}.$ 
%over the spaces $\mathcal{H}_q,$ ${q=1,\dots,m,}$ and $\mathcal{M}_{d_{q+1},d_q}(\mathbb{R}),$ ${\mathbb{R}^{d_{q+1}},}$ ${q=0,\dots,m,}$ respectively.  
%$v_1,\dots,v_m,M_0,\dots,M_m,b_0,\dots,b_m$ 
We want to show that $G_{\mathrm{min}}$ is also a minimum, i.e., there exists 
$v_q^{(*)}\in \mathcal{H}_q,$ $q=1,\dots,m,$ and ${M_q^{(*)}\in \mathcal{M}_{d_{q+1},d_q}(\mathbb{R}),}$ ${b_q^{(*)}\in \mathbb{R}^{d_{q+1}}},$ ${q=0,\dots,m,}$ 
such that
\[
G(v_1^{(*)},\dots,v_m^{(*)},M_0^{(*)},\dots,M_m^{(*)},b_0^{(*)},\dots,b_m^{(*)})=G_{\mathrm{min}}. 
\]

\bigskip

We prove this under the following weak assumptions which are satisfied in all practical cases within our problem space. We introduce the following action of translation on diffeomorphisms: $b\cdot \varphi_q: x \mapsto \varphi(x+b) - b$ and corresponding infinitesimal action on vector fields $b\cdot f: x \mapsto f(x+b)$.  (As is customary, we use the same notation for action and infinitesimal action.)
\begin{enumerate}[label = (H\arabic*), wide]
    \item The Hilbert norms on $V_q$ (recall that ${\mathcal H_q = L^2([0,1], V_q)}$) are translation invariant: for any $f \in V_q$ and $b\in \mathbb R^{d_{q}}$, the vector field $b\cdot f$ belongs to $V_q$ with $\|b\cdot f\|_{V_q} = \|f\|_{V_q}$.
    \item The functions $\Gamma_k$ are continuous, non-negative and satisfy $\Gamma_k(\zeta) \to \infty$ when $\|\zeta\|\to \infty$.
    \end{enumerate}
The existence proof is detailed below, first in the unconstrained case of equation \ref{eq:A1}, then in the ``sub-Riemannian'' case introduced in Section \ref{sec:distill}.

\bigskip

% \noindent{\bf Notation.}
% For any diffeomorphism $\psi$ and translation vector $b$ in the domain of $\psi$, 
% we define the diffeomorphism group action 
% \[
% b\cdot \psi(x)=\psi(x+b)-b.
% \]
% Let $\psi$ be associated with vector field $f$
% \[
% \partial_t \psi = f(\psi).
% \]
% Applying this action to $\psi$ 
% \[
% \partial_t (b\cdot \psi(x)) = f(\psi(x+b)) = f(b \cdot \psi(x) + b) = b\cdot f(b \cdot \psi(x))
% \]
% results in the group action on $f$
% \[
% b \cdot f(x)=f(x+b).
% \]

% \bigskip

\noindent{\bf Existence: unconstrained case.}
% For $v\in \mathcal H_q$, applying this action to the associated flow $\bvarphi_v$ 
% \[
% \partial_t (b\cdot \bvarphi_v(t,x)) = v(t)(\bvarphi_v(t,x+b)) = v(t)(b \cdot \bvarphi_v(t,x) + b) = b\cdot v(t)(b \cdot \bvarphi_v(t,x))
% \]
% results in the group action on $v$ 
% \[
% b \cdot v(t)(x)=v(t)(x+b).
% \]
If $v\in \mathcal H_q$ and $b \in \mathbb R^{d_{q}}$, we will denote by $b\cdot v$ the time-dependent vector field ${t\mapsto b\cdot v(t)}$.  Importantly, the associated flow
$\bvarphi_{b\cdot v}$ (defined by ${\partial_t \bvarphi_{b\cdot v}(t,x) = b\cdot v(t)(\bvarphi_{b\cdot v}(t,x))}$ and $\bvarphi_{b\cdot v}(0,x) = x$) satisfies $\bvarphi_{b\cdot v} = b\cdot \bvarphi_{v}$.  

% We check that $\bvarphi_{b\cdot v}(t,x) =\bvarphi_v(t,x+b) - b$, where $\bvarphi_v$ is the flow associated with $v$. Indeed, we have
% \[
% \partial_t (\bvarphi_{b\cdot v}(t,x)) = v(t)(\bvarphi_v(t,x+b)) = v(t)(\bvarphi_{b\cdot v}(t,x) + b) = b\cdot v(t)(\bvarphi_{b\cdot v}(t,x)).
% \]
% Since $\bvarphi_{b\cdot v}$ and $b \cdot \bvarphi_v$ coincide at $t=0$ and satisfy the same differential equation, they coincide for all $t$.

Indeed $b \cdot \bvarphi_v(0,x) = \bvarphi_v(0,x+b) - b = x$, and 
% where $\bvarphi_v$ is the flow associated with $v$.
% We check that 
% \[
% \bvarphi_{b\cdot v}(t,x) = b \cdot \bvarphi_v(t,x) \iff b\cdot v(t)(x) = v(t)(x+b). 
% \]
% Indeed, we have for $(\impliedby)$,
\begin{align*}
\partial_t (b \cdot \bvarphi_v)(t,x) &= \partial_t (\bvarphi_v(t,x+b)-b) \\
&= v(t)(\bvarphi_v(t,x+b)) \\
&= b\cdot v(t)(\bvarphi_v(t,x+b)-b) = b\cdot v(t)((b\cdot \bvarphi_v)(t,x)).
\end{align*}

% Since $\bvarphi_{b\cdot v}$ and $\bvarphi_v(t,x+b) - b$ coincide at $t=0$ and satisfy the same differential equation, they coincide for all $t$.  For $(\implies)$, we have
% \[
% b\cdot v(t)(\bvarphi_{b\cdot v}(t,x))=\partial_t \bvarphi_{b\cdot v}(t,x) = v(t)(\bvarphi_v(t,x+b)) = v(t)(\bvarphi_{b\cdot v}(t,x) + b).
% \]

As a consequence, if (H1) is true, one can assume, without changing the value of the infimum, that $b_0 = \cdots = b_{m-1} = 0$. Indeed, given $v_1, \ldots, v_m$, $M_0, \ldots, M_m$, $b_0, \ldots, b_m$, one can define
\[
c_q = \begin{cases} b_0, & \text{if $q = 0$}\\
 M_q c_{q-1} + b_q, & \text{if $1\le q \le m,$} 
 \end{cases}
\]
and, one has, letting $\tilde v_q = c_{q-1} \cdot v_q$, $q\geq 1$, $\tilde b_0 = \cdots = \tilde b_{m-1} = 0$, $\tilde b_m = c_m$,
\begin{align}
\begin{split}\label{eq:b}
    &G(\tilde v_1,\dots, \tilde v_m, M_0,\dots,M_m, 0,\cdots, 0, \tilde  b_m) \\
    &\qquad\qquad\qquad\qquad\qquad\quad= G(v_1,\dots,v_m,M_0,\dots,M_m,b_0,\dots,b_m).
\end{split}
\end{align}
To see this, let $\bvarphi_{\tilde v_q}$ and $\tilde \zeta_k^q$ be defined by \eqref{eq:appendix:1.1}, i.e.,
\[
\left\{
\begin{aligned}
\partial_t\bvarphi_{\tilde v_q}(t)&=\tilde v_q(t)\circ \bvarphi_{\tilde v_q}(t),\quad t\in[0,1] \\
\tilde \zeta_k^{q+1} &= M_q \bvarphi_{\tilde v_q}(1,\tilde \zeta_k^q) + \tilde b_q\\
\bvarphi_{\tilde v_q}(0)&=\mbox{id},\quad \tilde \zeta_k^1 = M_0\iota_s(x_k)+ \tilde b_0.
\end{aligned}
\right.
\]
As we just saw, we have $\bvarphi_{\tilde v_q} = c_{q-1}\cdot \bvarphi_{v_q}$. 
Moreover, for $q\leq m-1$ (so that $\tilde b_q=0$), we have
\[
\tilde \zeta_k^{q+1} = M_q (c_{q-1}\cdot\bvarphi_{ v_q})(1,\tilde \zeta_k^q) = M_q \bvarphi_{v_q}(1,\tilde \zeta_k^q + c_{q-1}) - M_q c_{q-1}, 
\]
yielding
\[
\tilde \zeta_k^{q+1} + c_q  = M_q \bvarphi_{v_q}(1,\tilde \zeta_k^q + c_{q-1}) + b_q.
\]
So $\tilde \zeta_k^{q+1} + c_q$ satisfy the same iterations as  $\zeta_k^{q+1}$, with same initial condition 
\[
\tilde \zeta_k^1 + c_0 = M_0 \iota_s(x_k) + c_0 = \zeta_k^1
\]
(since $c_0=b_0$). This shows that $\tilde \zeta_k^{q+1} + c_q = \zeta_k^{q+1}$, $q=0, \ldots, m-1$. 
Finally, 
\[
\tilde \zeta_k^{m+1} = M_m \bvarphi_{v_m}(1,\tilde \zeta_k^m + c_{m-1}) - M_mc_{m-1} + c_m = M_m \bvarphi_{v_m}(1,\zeta_k^m) + b_m = \zeta_k^{m+1}
.\]
Since $\|\tilde v_q \|_{\mathcal H_q} = \|v_q \|_{\mathcal H_q}$, \eqref{eq:b} is satisfied.

% the objective function can be equivalently expressed as
% \begin{align}
% &G(\hat v_1,\dots,\hat v_m,M_0,\dots,M_m,b_0,\dots,b_m)=\sum_{q=1}^m \|\hat v_q\|_{\mathcal{H}_q}^2 + \lambda \sum_{q=0}^{m}\| M_q \|^2 \nonumber \\
% & \quad + \frac{1}{\sigma^2}\sum_{k=1}^N \Gamma_k(\pi_r({M}_m\hat \bvarphi_m(1)({M}_{m-1}\hat \bvarphi_{m-1}(1)(\cdots M_1\hat \bvarphi_1(1)({M}_0(\iota_s(x_k)))\cdots))+c_m)), \nonumber  
% \end{align}
\bigskip

We now conclude the argument by considering a minimizing sequence for $G$ in the form 
$v_q^{(n)}\in \mathcal{H}_q,$ $q=1,\dots,m,$ and ${M_q^{(n)}\in \mathcal{M}_{d_{q+1},d_q}(\mathbb{R}),}$ ${b_q^{(n)}\in \mathbb{R}^{d_{q+1}}},$ $q=0,\dots,m,$ with $b^{(n)}_0 = \cdots = b^{(n)}_{q-1} = 0$, satisfying
\[
% G(v_1^{(n)},\dots,v_m^{(n)},M_0^{(n)},\dots,M_m^{(n)})\rightarrow G_{min}.
\lim_{n\rightarrow \infty}{G(v_1^{(n)},\dots,v_m^{(n)},M_0^{(n)},\dots,M_m^{(n)},b_0^{(n)},\dots,b_m^{(n)})}= G_{\mathrm{min}}.
\]
Denote by $\bvarphi_{v_q^{(n)}}$ and ${\zeta_k^q}^{(n)}$ the diffeomorphisms and vectors  defined in \eqref{eq:appendix:1.1} for each $n$.

Each $M_q^{(n)}$ sequence is bounded in $\mathcal{M}_{d_{q+1},d_q}(\mathbb{R}),$ and each $v_q^{(n)}$ is bounded in $\mathcal H_q$. There is therefore no loss of generality (just using a subsequence) in assuming that $M_q^{(n)}$ converges to some ${M_q^{(*)}\in \mathcal{M}_{d_{q+1},d_q}(\mathbb{R})}$, and that $v_q^{(n)}$ converges weakly in $\mathcal H_q$ to some $v_q^{(*)}$ that satisfies
% \begin{equation}\label{eq:appendix.eq1}
% \| M_q^{n_j} \|^2\rightarrow \| M_q^* \|^2\quad (j\rightarrow \infty). 
% \end{equation}
% % By definition, 
% Each $v_q^n$ sequence is bounded in the corresponding Hilbert space $\mathcal{H}_q,$   
% with a subsequence (denoted $v_q^{n_j}$) that weakly converges to some $v_q^*\in \mathcal{H}_q.$ Therefore,   
\begin{equation}\label{eq:appendix.eq2}
\liminf_{n\rightarrow \infty} {\| v_q^{(n)}\|_{\mathcal{H}_q}^2} \ge \| v_q^{(*)}\|_{\mathcal{H}_q}^2. 
\end{equation}

Let $\bvarphi_{v_q^{(*)}}(t)$ be the flow associated with $v_q^{(*)}$.   Weak convergence in $\mathcal{H}_q$ 
implies, at each fixed $t\in[0,1],$ uniform convergence of $\bvarphi_{v_q^{(n)}}(t)$ to ${\bvarphi_{v_q^{(*)}}(t)}$ on $\mathbb{R}^{d_q}$ \citep{younes2010shapes}. As a consequence, for all $q\leq m$, the sequence $({\zeta_k^q}^{(n)}, n\geq 0)$ also converges to a limit ${\zeta_k^q}^{(*)}$ that satisfies \eqref{eq:appendix:1.1}. 

Each $\Gamma_k(\pi_r({\zeta_k^{m+1}}^{(n)}))$ must be bounded independently of $n$, since we have a minimizing sequence. Assumption (H2) then implies that  $\pi_r({\zeta_k^{m+1}}^{(n)})$ is also bounded, with $\pi_r({\zeta_k^{m+1}}^{(n)}) = \pi_r(M_m^{(n)}\bvarphi_{v_m^{(n)}}(1,{\zeta_k^m}^{(n)})) + \pi_r(b_m^{(n)})$.  Since the first term in this sum converges, we see that $\pi_r(b_m^{(n)})$ is also bounded, so that,  taking a subsequence if needed, we  can assume that $\pi_r(b_m^{(n)})$ converges to some $b_m^{(*)}$ (with $\pi_r(b_m^{(*)}) = b_m^{(*)}$).  Using \eqref{eq:appendix.eq2}, we obtain 
\[
G(v_1^{(*)},\dots,v_m^{(*)},M_0^{(*)},\dots,M_m^{(*)},0,\dots,0, b_m^{(*)})= G_{\mathrm{min}},
\]
which concludes the proof.
\bigskip

\noindent{\bf Existence: sub-Riemannian case.}
The situation in which the vector fields $v_q$ are restricted to sub-optimal finite-dimensional spaces, as considered in  Section \ref{sec:distill}, is handled similarly, and follows arguments previously made in \citet{Younes2012,arguillere2015shape,gris2018sub,younes:hal-02386227}.
Here, we associate a closed subspace of $V_q$ with a
% with every 
diffeomorphism $\psi$ on $\mathbb{R}^{d_q}$. This subspace will also depend on the configuration (denoted $\zeta^q$) that comes as input to the diffeomorphic module.
%and family $\zeta$ of  ``objects" on $\mathbb{R}^{d_q}$. 
%to which $\psi$ is applied to generate the subspace.
% that define $\psi$.
% a closed subspace, say  $W(\bvarphi,\zeta)$, of $V_q$
We denote this subspace as $W_q(\psi,\zeta)$, with $\psi \in \mathrm{Diff}_{V_q}$ and $\zeta \in (\mathbb{R}^{d_q})^N$. We also denote the orthonormal projection of $f\in V_q$ onto $W_q(\psi,\zeta)$ as $P_{W_q(\psi,\zeta)}(f)$. 
% by $P_{W(\bvarphi,\zeta)}(f)$ the orthonormal projection of $f\in V_q$ onto $W(\bvarphi,\zeta)$.

We will make the following hypotheses on the spaces $W_q(\psi,\zeta)$, which form a ``distribution'' in the terminology of sub-Riemannian geometry.
\begin{enumerate}[label = (HS\arabic*), wide]
    \item For $b\in \mathbb{R}^{d_q}$, let ${b\cdot W_q(\psi,\zeta) = \{b\cdot f: f\in W_q(\psi,\zeta)\}}$.  We assume $b\cdot W_q(\psi,\zeta) = W_q(b\cdot \psi,\zeta-b)$.
    \item The spaces $W_q(\psi,\zeta)$ depend continuously on $\psi$ and $\zeta$, in the sense that the mapping ${\psi\mapsto P_{W_q(\psi,\zeta)}}$, 
which takes values in the space of linear operators on $V_q$, 
is continuous in $\psi$ (for uniform convergence) and $\zeta$.
    \end{enumerate}
    
%     We require   , i.e., 
% \[
% f \in W_q(\psi,\zeta) \iff b \cdot f \in W_q(b\cdot \psi,\zeta+b)\,.
% \]
% We still assume $\|b\cdot f\|_{V_q} = \|f\|_{V_q}$, i.e., (H1) is satisfied.
% and $\|b\cdot f\|_{V_q} = \|f\|_{V_q}$ with $b\cdot f: x \mapsto f(x+b)$.

In the setting of equation \eqref{eq:A1}, we now add to the minimization the requirement that each $v_q$ belongs to the space
\[
\mathcal W_q(\bvarphi_{v_q}(\cdot),\zeta^q) = \left\{ v\in \mathcal H_q: v(t) \in W_q(\bvarphi_{v_q}(t),\zeta^q) \text{ for almost all } t\in [0,1]\right\}.
\] 
Then, assuming (H1), (H2), (HS1) and (HS2), there exists a solution to this minimization problem.

\medskip
The proof starts by repeating the argument made in the unconstrained case. The combination of (H1) and (HS1) allows us to claim that there is no loss of generality in restricting the minimization to $b_1 = \cdots = b_{m-1} = 0$. Then,  given any minimizing sequence $v_q^{(n)}\in \mathcal{H}_q,$ $q=1,\dots,m$, ${M_q^{(n)}\in \mathcal{M}_{d_{q+1},d_q}(\mathbb{R})}$, ${b_q^{(n)}\in \mathbb{R}^{d_{q+1}}}$, $q=0,\dots,m,$ with $b^{(n)}_0 = \cdots = b^{(n)}_{m-1} = 0$, one can find a subsequence such that each $v_q^{(n)}$ converges weakly to $v_q^{(*)}\in \mathcal H_q$, and $M_q^{(n)}, b_q^{(n)}$ converge to $M_q^{(*)}, b_q^{(*)}$, and such that the limit achieves the minimum of the objective function in \eqref{eq:A1}, with the additional property that $\bvarphi_{v_q^{(n)}}$ converges uniformly to $\bvarphi_{v_q^{(*)}}$ (which also ensures that the sequence ${\zeta^{q}}^{(n)}$ converges to a limit ${\zeta^{q}}^{(*)}$).

The only point that remains to be shown in the sub-Riemannian context is that $v_q^{(*)}$ satisfies the constraints, i.e., that $v_q^{(*)}\in \mathcal W_q(\bvarphi_{v_q^{(*)}}(\cdot),{\zeta^q}^{(*)})$, $q=1,\dots,m$. We now proceed with the argument.

% with ${{z^q_k}^{(*)}(t) = \bvarphi_{v^{(*)}_q}(t)({\zeta_k^q}^{(*)})}$. 

Given a continuous function $\bvarphi: t \mapsto \bvarphi(t)$ and $\zeta\in (\mathbb R^{d_q})^N$ , let $\boldsymbol P_{q,\bvarphi, \zeta}$ be defined on $\mathcal H_q$ by ${\boldsymbol P_{q,\bvarphi, \zeta}(v)(t) =P_{W_q(\bvarphi(t),\zeta)}(v(t))}$. Clearly, $\boldsymbol P_{q,\bvarphi, \zeta}$ is bounded,  maps $\mathcal H_q$ to $\mathcal W_q(\bvarphi(\cdot),\zeta)$, and $\boldsymbol P_{q,\bvarphi, \zeta}(v) = v$ if and only if $v \in \mathcal W_q(\bvarphi(\cdot),\zeta)$, showing that this set is closed and that $\boldsymbol P_{q,\bvarphi, \zeta}$ is its orthogonal projection. Moreover, if $\bvarphi^{(n)}$ converges to $\bvarphi$ and $\zeta^{(n)}$ to $\zeta$, then $\boldsymbol P_{q,\bvarphi^{(n)}, \zeta^{(n)}}$ converges to $\boldsymbol P_{q,\bvarphi, \zeta}$, as can be deduced by dominated convergence and the hypotheses made on $P_{W_q(\bvarphi(t),\zeta)}$.

Returning to $v_q^{(*)}$, assume that $v\in \mathcal H_q$ is perpendicular to $\mathcal W_q(\bvarphi_{v_q^{(*)}}(\cdot),{\zeta^q}^{(*)})$. Then 
\[
|\langle v, v_q^{(n)}\rangle_{\mathcal H_q}| = |\langle \boldsymbol P_{q, \bvarphi_{v_q^{(n)}}, {\zeta^q}^{(n)}}(v), v_q^{(n)}\rangle_{\mathcal H_q}| \leq \|\boldsymbol P_{q,\bvarphi_{v_q^{(n)}}, {\zeta^q}^{(n)}}(v)\|_{\mathcal H_q} \|v_q^{(n)}\|_{\mathcal H_q}.
\]
Since $\boldsymbol P_{q, \bvarphi_{v_q^{(n)}}, {\zeta^q}^{(n)}}(v) $ converges to $\boldsymbol P_{q,\bvarphi_{v_q^{(*)}}, {\zeta^q}^{(*)}}(v) = 0$, we find that 
$\langle v, v_q^{(n)}\rangle_{\mathcal H_q}$ tends to 0. By weak convergence, this quantity also converges to $\langle v, v_q^{(*)}\rangle_{\mathcal H_q}$, which must therefore also vanish. Since this is true for all $v\in \mathcal W_q({\bvarphi_{v_q^{(*)}}}(\cdot),{\zeta^q}^{(*)})^\perp$, we find that ${v_q^{(*)} \in (\mathcal W_q({\bvarphi_{v_q^{(*)}}}(\cdot),{\zeta^q}^{(*)})^\perp)^\perp = \mathcal W_q({\bvarphi_{v_q^{(*)}}}(\cdot),{\zeta^q}^{(*)})}$ (since the space is closed). This concludes the proof in the sub-Riemannian case.
\medskip

To conclude, we check that (HS1) and (HS2) hold 
in the context of Section \ref{sec:distill} (assuming that (H1) is true). In that section,
the finite-dimensional space is generated by the columns of the matrices $K_q(\cdot, z_l)$, $l=1, \ldots, N_S$, which leads us to define 
\[
W_q(\psi,\zeta) = \left\{\sum_{l=1}^{N_S} K_q(\cdot, z_l) w_l: w_1, \ldots, w_{N_S}\in \mathbb R^{d_q}, z_l = \psi(\zeta_l)\right\},
\] 
for a diffeomorphism $\psi$ and $\zeta \in (\mathbb R^{d_q})^N$.  We have $f \in b\cdot W_q(\psi,\zeta)$ if and only if there exists $w_1,\ldots, w_{N_S} $ such that, for all $x\in \mathbb R^{d_q}$, 
\[
    f(x) = \sum_{l=1}^{N_S} K_q(x+b, z_l) w_l
\]
with $z_l = \psi(\zeta_l)$. 
By translation invariance of the norm in $V_q$, this is equivalent to 
\[
  f(x)  = \sum_{l=1}^{N_S} K_q(x, z_l-b) w_l.
\]
We have $z_l - b = \psi(\zeta_l - b + b) - b = (b\cdot \psi)(\zeta_l - b)$ showing that $f \in b\cdot W_q(\psi,\zeta)$ is equivalent to $f \in W_q(b\cdot\psi, \zeta - b)$, proving (HS1). 

Continuity of the projections is true %in the context of Section \ref{sec:distill} 
because $P_{W(\psi,\zeta^q)}(f)$ for $f\in V_q$ takes the form 
\[
\sum_{l=1}^{N_S} K_q(\cdot, z_l) w_{l}(f),
\]
where $w_{1}(f), \ldots, w_{N_S}(f)$ satisfy the linear system
\[
\sum_{l=1}^{N_S} K_q(z_k, z_l) w_{l}(f) = f(z_k), \quad k=1, \ldots, N_S,
\]
which has a unique solution, continuous in $z$ (over the set of $N_S$ distinct points in $\mathbb R^{d_q}$) and thus in $\psi$ and $\zeta$. 

\newpage

\section{Necessary Conditions for Optimality}\label{sec:PMP}
% \subsection{Optimal Vector Fields}\label{sec:PMPoptimal}
% For vector fields
% \[
% v_q(t)(\cdot) = \sum_{l=1}^{N}K_q(\cdot,z_l^q(t))a_l^q(t),
% \]
% define the Lagrangian or running cost function of each D$_q$ module as
% \[
% L_q(\vect{u},\vect{w})=\sum_{k,l=1}^{N}w_k^TK_q(u_k,u_l)w_l,\quad \vect{u},\vect{w}\in(\mathbb{R}^{d_q})^N.
% \]
% \[
% L(\vect{z}^q(t),\vect{a}^q(t)):=\sum_{k,l=1}^{N}a_k^q(t)^TK_q(z_k^q(t),z_l^q(t))a_l^q(t),
% \]
Recall the notation for the general case of sub-optimal vector fields, where a training data subset of size $N_S\le N$ is chosen, the training data renumbered such that the first $N_S$ elements coincide with this subset, and ${(z_1^q(\cdot),\dots,z_{N_S}^q(\cdot))}$ and ${\vect{a}^q(\cdot)=(a_1^q(\cdot),\dots,a_{N_S}^q(\cdot))}$ represent the states corresponding to this subset and the controls, respectively.     
We now let $G$ denote the general reduced objective function in \eqref{eq:obj4}, namely
%The objective function $G(\vect{a}^1(\cdot),\dots,\vect{a}^m(\cdot),{A}_0,\dots,{A}_m)$ can be expressed as
\begin{multline*}
% &G(\vect{a}^1(\cdot),\dots,\vect{a}^m(\cdot),{A}_0,\dots,{A}_m)\nonumber \\
% &\qquad\qquad\qquad =
G(\vect{a}^1(\cdot),\dots,\vect{a}^m(\cdot),{A}_0,\dots,{A}_m) =
\sum_{q=1}^{m} \int_{0}^{1}L_q(\vect{z}^q(t),\vect{a}^q(t))dt 
\\ + \lambda \sum_{q=0}^{m}U_q({A}_q) + \frac{1}{\sigma^2}\sum_{k=1}^N \Gamma_k(\pi_r(\zeta_k^{m+1})), 
\end{multline*}
where
the Lagrangian or running cost functions $L_q$ are 
\begin{align}
L_q: (\mathbb{R}^{d_q})^{N_S}\times(\mathbb{R}^{d_q})^{N_S}&\rightarrow \mathbb{R}\nonumber \\
(\vect{u},\vect{w})&\mapsto \sum_{k,l=1}^{N_S} w_k^TK_q(u_k,u_l)w_l. \nonumber
\end{align}
% with $S\subset \{1, \ldots, N\},$ $|S|=N_S\le N.$
The dynamical system constraints are, for $k=1, \ldots, N$,
\[
\left\{
\begin{aligned}
\xi_k^0&=\iota_s(x_k)\\
\zeta_k^q&={A}_{q-1}(\xi_k^{q-1})\\
z_k^q(0)&=\zeta_k^q \\
\partial_t {z}_k^q(t)&= v_q(t)(z_k^q(t))\\
\xi_k^q&=z_k^q(1),\\
\end{aligned}
\right.
\]
where 
\[
v_q(t)(\cdot) = \sum_{l=1}^{N_S} K_q(\cdot,z_l^q(t))a_l^q(t).
\]
  
Adjoin these constraints to $G$
by the Lagrange multipliers ${\rho_1^q,\dots,\rho_N^q\in\mathbb{R}^{d_q}}$ and  ${\vect{p}^q(\cdot)=(p_1^q(\cdot),\dots,p_N^q(\cdot)),}$ ${q=1,\dots,m,}$ respectively, 
% \begin{equation}
% \sum_{q=1}^{m} \left[\sum_{k=1}^N \rho_k^q(z_k^q(0)-\zeta_k^q)+\int_{0}^{1}\bigg(L(\vect{z}^q,\vect{a}^q)+\sum_{k=1}^N {p_k^q}^T(\dot{z}_k^q-v_q(t,z_k^q))\bigg)dt \right] + \lambda \sum_{q=0}^{m}U_q({A}_q) + \frac{1}{\sigma^2}\sum_{k=1}^N \Gamma_k(\pi_r(\zeta_k^{m+1})), \nonumber
% \end{equation}
% \begin{align}
% &\sum_{q=1}^{m} \left[\sum_{k=1}^N {\rho_k^q}^T(z_k^q(0)-\zeta_k^q)+\int_{0}^{1}\bigg(L_q(\vect{z}^q,\vect{a}^q)+\sum_{k=1}^N {p_k^q}^T \bigg(\dot{z}_k^q - \sum_{l=1}^{N}K_q(z_k^q,z_l^q)a_l^q\bigg)\bigg)dt \right] \nonumber \\
% &\qquad\qquad\qquad\qquad\qquad + \lambda \sum_{q=0}^{m}U_q({A}_q) + \frac{1}{\sigma^2}\sum_{k=1}^N \Gamma_k(\pi_r(\zeta_k^{m+1})). \nonumber
% \end{align}
\begin{align}
&\sum_{q=1}^{m} \left[\sum_{k=1}^N {\rho_k^q}^T(z_k^q(0)-\zeta_k^q)+\int_{0}^{1}\bigg(L_q(\vect{z}^q,\vect{a}^q)+\sum_{k=1}^N {p_k^q}^T(\partial_t{z}_k^q-v_q(t)(z_k^q))\bigg)dt \right] \nonumber\\
&\qquad\qquad\qquad\qquad\qquad\qquad\qquad\qquad\qquad + \lambda \sum_{q=0}^{m}U_q({A}_q) + \frac{1}{\sigma^2}\sum_{k=1}^N \Gamma_k(\pi_r(\zeta_k^{m+1})), \nonumber
\end{align}
where $p_k^q(\cdot)$ is a time-dependent vector in $\mathbb{R}^{d_q}$. 
In the Hamiltonian formulation, the adjoined objective function becomes
\begin{align}
&\sum_{q=1}^{m} \left[\sum_{k=1}^N {\rho_k^q}^T(z_k^q(0)-\zeta_k^q)+\int_{0}^{1}\bigg(\sum_{k=1}^N {p_k^q}^T \partial_t {z}_k^q - H^q_{\vect{a}^q}(\vect{z}^q,\vect{p}^q) \bigg)dt \right] \nonumber\\
&\qquad\qquad\qquad\qquad\qquad\qquad\qquad\qquad\qquad + \lambda \sum_{q=0}^{m}U_q({A}_q) + \frac{1}{\sigma^2}\sum_{k=1}^N \Gamma_k(\pi_r(\zeta_k^{m+1})), \nonumber
\end{align}
with Hamiltonian functions $H^q_{\vect{w}},$ $\vect{w}\in(\mathbb{R}^{d_q})^{N_S},$ 
\begin{align}
H^q_{\vect{w}}:(\mathbb{R}^{d_q})^N\times(\mathbb{R}^{d_q})^N&\rightarrow \mathbb{R}\nonumber \\
                  (\vect{u},\vect{r})&\mapsto \sum_{k=1}^N {r_k}^T \sum_{l=1}^{N_S} K_q(u_k,u_l)w_l -L_q(\vect{u}_S,\vect{w})\nonumber\\
&\phantom{\mapsto } =\sum_{k=1}^N \sum_{l=1}^{N_S} {r_k}^T  K_q(u_k,u_l)w_l - \sum_{k,l=1}^{N_S} w_k^T K_q(u_k,u_l)w_l, \nonumber
\end{align} 
where $\vect{u}_S=( u_i\in\vect{u}, i=\{1,\dots,N_S\}).$
Apply the calculus of variations: %to the adjoined objective function   
\begin{align}
&\sum_{q=1}^{m} \left[\sum_{k=1}^N \right. ((z_k^q(0)-\zeta_k^q)\delta \rho_k^q + \rho_k^q \delta z_k^q\rvert_0  \nonumber\\
& \qquad\qquad\qquad\qquad - \rho_k^q \partial_{M_{q-1}}\zeta_k^{q} \delta M_{q-1} - \rho_k^q \partial_{b_{q-1}}\zeta_k^{q} \delta b_{q-1} - \rho_k^q \partial_{\xi_k^{q-1}}\zeta_k^q \delta \xi_k^{q-1})  \nonumber \\
&\quad\quad\quad + \left.\int_{0}^{1}\bigg(\sum_{k=1}^N (\partial_t {z}_k^q\delta p_k^q + p_k^q \partial_t \delta{z}_k^q - \partial_{p_k^q}H^q_{\vect{a}^q}\delta p_k^q - \partial_{z_k^q}H^q_{\vect{a}^q}\delta z_k^q) - \sum_{k=1}^{N_S}(\partial_{a_k^q}H^q_{\vect{a}^q}\delta a_k^q) \bigg)dt \right] \nonumber \\
&\quad + \lambda \sum_{q=0}^{m}(\partial_{M_q}U_q({A}_q)\delta M_q + \partial_{b_q}U_q({A}_q)\delta b_q)  \nonumber \\
&\quad +  \frac{1}{\sigma^2}\sum_{k=1}^N (\partial_{M_m}\Gamma_k(\pi_r(\zeta_k^{m+1}))\delta M_m + \partial_{b_m}\Gamma_k(\pi_r(\zeta_k^{m+1}))\delta b_m +
% \nonumber \\
% &\qquad\qquad\qquad\qquad\qquad\qquad\qquad\qquad\qquad +
\partial_{\xi_m^q}\Gamma_k(\pi_r(\zeta_k^{m+1}))\delta \xi_m^q). \nonumber
\end{align}
Substitute 
% \begin{equation}
% \int_{0}^{1}p_k^q\delta\dot{z}_k^q dt = \cancelto{p_k^q(1)\delta\xi_k^q}{\vphantom{\sum}(p_k^q\delta z_k^q)\rvert_1} - (p_k^q\delta z_k^q)\rvert_0 - \int_0^1\dot{p}_k^q\delta z_k^q dt \nonumber
% \end{equation}
\begin{align}
\int_{0}^{1}p_k^q \partial_t \delta{z}_k^q dt &= (p_k^q\delta z_k^q)\rvert_1 - (p_k^q\delta z_k^q)\rvert_0 - \int_0^1\partial_t {p}_k^q\delta z_k^q dt \nonumber \\
&= p_k^q(1)\delta\xi_k^q - (p_k^q\delta z_k^q)\rvert_0 - \int_0^1\dot{p}_k^q\delta z_k^q dt \nonumber
\end{align}
and $\partial_{b_q}U_q({A}_q)=0\in\mathbb{R}^{d_{q+1}}$:
\begin{align}
&\sum_{q=1}^{m} \left[\sum_{k=1}^N \right. ((z_k^q(0)-\zeta_k^q)\delta \rho_k^q + (\rho_k^q \delta z_k^q-p_k^q\delta z_k^q)\rvert_0  \nonumber\\
& \qquad\qquad\qquad\qquad - \rho_k^q (\xi_k^{q-1})^T \delta M_{q-1} - \rho_k^q \delta b_{q-1} - M_{q-1}^T\rho_k^q  \delta \xi_k^{q-1} +p_k^q(1)\delta \xi_k^q )  \nonumber\\
&\quad\quad\quad + \left.\int_{0}^{1}\bigg(\sum_{k=1}^N (\partial_t{z}_k^q\delta p_k^q - \partial_t {p}_k^q \delta z_k^q - \partial_{p_k^q}H^q_{\vect{a}^q}\delta p_k^q - \partial_{z_k^q}H^q_{\vect{a}^q}\delta z_k^q) - \sum_{k=1}^{N_S}(\partial_{a_k^q}H^q_{\vect{a}^q}\delta a_k^q) \bigg)dt \right] \nonumber \\
&\quad + \lambda \sum_{q=0}^{m}\partial_{M_q}U_q({A}_q)\delta M_q  \nonumber\\
&\quad + \frac{1}{\sigma^2}\sum_{k=1}^N \big(\iota_r(\nabla\Gamma_k(\pi_r(\zeta_k^{m+1}))){\xi_k^{m}}^T\delta M_m + \iota_r(\nabla\Gamma_k(\pi_r(\zeta_k^{m+1})))\delta b_m \nonumber \\ 
&\qquad\qquad\qquad\qquad\qquad\qquad\qquad\qquad\qquad + M_m^T\iota_r(\nabla\Gamma_k(\pi_r(\zeta_k^{m+1})))\delta \xi_k^m\big). \nonumber
\end{align}
Group terms by variation:
% \begin{align}
% &\sum_{k=1}^N \left[\sum_{q=1}^{m} \right. ((z_k^q(0)-\zeta_k^q)\delta \rho_k^q + (\rho_k^q -p_k^q(0))\delta z_k^q\rvert_0 - \rho_k^q (\xi_k^{q-1})^T \delta M_{q-1} - \rho_k^q \delta b_{q-1}) \nonumber\\
% &\quad\quad\quad + \sum_{q=1}^{m-1} (p_k^q(1) - M_{q}^T\rho_k^{q+1})\delta \xi_k^q  \nonumber\\
% &\quad\quad\quad + \left.\int_{0}^{1}\bigg(\sum_{q=1}^{m} ((\dot{z}_k^q - \partial_{p_k^q}H_{\vect{a}^q})\delta p_k^q - (\dot{p}_k^q + \partial_{z_k^q}H_{\vect{a}^q})\delta z_k^q - \partial_{a_k^q}H_{\vect{a}^q}\delta a_k^q) \bigg)dt \right] \nonumber \\
% &\quad + \lambda \sum_{q=0}^{m}(\partial_{M_q}U_q({A}_q)\delta M_q) \nonumber\\
% &\quad + \frac{1}{\sigma^2}\sum_{k=1}^N (\iota_r(\nabla\Gamma_k(\pi_r(\zeta_k^{m+1}))){\xi_k^{m}}^T\delta M_m + \iota_r(\nabla\Gamma_k(\pi_r(\zeta_k^{m+1})))\delta b_m + (M_m^T\iota_r(\nabla\Gamma_k(\pi_r(\zeta_k^{m+1}))) +\sigma^2 p_k^m(1))\delta \xi_k^m ), \nonumber
% \end{align}
\begin{align}
&\sum_{q=1}^m \left[\sum_{k=1}^{N} \right. ((z_k^q(0)-\zeta_k^q)\delta \rho_k^q + (\rho_k^q -p_k^q(0))\delta z_k^q\rvert_0) \nonumber\\
&\quad\quad\quad + \left.\int_{0}^{1}\bigg(\sum_{k=1}^{N} ((\partial_t{z}_k^q - \partial_{p_k^q}H^q_{\vect{a}^q})\delta p_k^q - (\partial_t {p}_k^q + \partial_{z_k^q}H^q_{\vect{a}^q})\delta z_k^q) - \sum_{k=1}^{N_S}(\partial_{a_k^q}H^q_{\vect{a}^q}\delta a_k^q) \bigg)dt \right] \nonumber \\
&\quad + \sum_{q=1}^{m-1}\sum_{k=1}^N  (p_k^q(1) - M_{q}^T\rho_k^{q+1})\delta \xi_k^q + \sum_{k=1}^N \bigg(\frac{1}{\sigma^2}M_m^T\iota_r(\nabla\Gamma_k(\pi_r(\zeta_k^{m+1}))) + p_k^m(1) \bigg)\delta \xi_k^m\nonumber \\
&\quad + \sum_{q=0}^{m-1}\bigg(\lambda\partial_{M_q}U_q({A}_q)-\sum_{k=1}^N\rho_k^{q+1} {\xi_k^{q}}^T \bigg)\delta M_q - \sum_{q=0}^{m-1}\sum_{k=1}^N \rho_k^{q+1}\delta b_q \nonumber\\
&\quad + \bigg(\frac{1}{\sigma^2}\sum_{k=1}^N \iota_r(\nabla\Gamma_k(\pi_r(\zeta_k^{m+1}))){\xi_k^{m}}^T + \lambda\partial_{M_m}U_m({A}_m) \bigg)\delta M_m \nonumber \\
&\quad + \bigg(\frac{1}{\sigma^2}\sum_{k=1}^N \iota_r(\nabla\Gamma_k(\pi_r(\zeta_k^{m+1})))\bigg)\delta b_m.  \nonumber 
\end{align}
% and setting the coefficients of the variations to zero, we have 
Setting the coefficients of the variations with respect to the forward states 
and boundary conditions 
to zero, we have the following backpropagation states 
and boundary conditions:
\begin{align}
% &z_k^q(0)=\zeta_k^q,\quad q=1,\dots,m \nonumber \\
% &\partial_t z_k^q(t) = \partial_{p_k^q(t)}H^q_{\vect{a}^q}(\vect{z}^q,\vect{p}^q) =\sum_{l=1}^{N}K_q(z_k^q(t),z_l^q(t))a_l^q(t),\quad q=1,\dots,m \nonumber \\ 
&p_k^m(1) = -\frac{1}{\sigma^2}M_m^T\iota_r(\nabla\Gamma_k(\pi_r(\zeta_k^{m+1}))) \nonumber \\
&p_k^q(1) = M_q^T\rho_k^{q+1},\quad q=m-1,\dots,1 \nonumber 
\end{align}
which imply
\begin{align}
&\rho_k^{m+1} = -\frac{1}{\sigma^2}\iota_r(\nabla\Gamma_k(\pi_r(\zeta_k^{m+1}))) \nonumber \\
&p_k^q(1) = M_q^T\rho_k^{q+1},\quad q=m,\dots,1, \nonumber 
\end{align}
and
\begin{align}
&\partial_t p_k^q(t) = -\partial_{z_k^q(t)}H_{\vect{a}^q}^q(\vect{z}^q,\vect{p}^q),\quad q=m,\dots,1 \nonumber \\
&\phantom{\partial_t p_k^q(t)} = -\sum\limits_{l=1}^{N_S}\nabla_1 K_q(z_k^q(t), z_l^q(t))p_k^q(t)^Ta_l^q(t)  \nonumber\\
&\qquad\qquad\qquad\quad -
\begin{cases}
    \sum_{l=1}^N\nabla_1K_q( z_k^q(t),z_l^q(t))a_k^q(t)^Tp_l^q(t)&  \\
    \qquad -2\sum_{l=1}^{N_S}\nabla_1K_q( z_k^q(t), z_l^q(t))a_k^q(t)^Ta_l^q(t),& \text{if } k\le N_S\\
    0,              &  \text{if } k> N_S 
\end{cases}  \vphantom{\sum\limits_{l=1}^{N_S}} \nonumber \\
&\rho_k^q = p_k^q(0),\quad q=m,\dots,1. \nonumber 
\end{align}
The coefficients of the variations with respect to our parameters are the gradients
\begin{align}
&\partial_{a_k^q(t)}G = - \partial_{a_k^q(t)}H^q_{\vect{a}^q}(\vect{z}^q,\vect{p}^q),\quad k=1,\dots,N_S, \hspace{0.2cm} q=1,\dots,m   \nonumber \\
&\phantom{\partial_{a_k^q(t)}G}=2\sum_{l=1}^{N_S} K_q(z_k^q(t), z_l^q(t))a_l^q(t) - \sum_{l=1}^{N} K_q( z_k^q(t),z_l^q(t))p_l^q(t) \nonumber \\
&\partial_{M_q} G = \lambda\partial_{M_q}U_q({A}_q) - \sum_{k=1}^{N} \rho_k^{q+1}{\xi_k^{q}}^T,\quad q=0,\dots,m\nonumber \\
&\partial_{b_q} G = - \sum_{k=1}^{N} \rho_k^{q+1},\quad q=0,\dots,m,\nonumber 
\end{align} 
which are calculated using the backpropagation states.
For the optimal vector fields case of $N_S = N,$ the expressions for the costates and gradient with respect to the controls simplify to
\[
\partial_t p_k^q(t) = -\sum_{l=1}^{N}\nabla_1 K_q(z_k^q(t),z_l^q(t))(p_k^q(t)^Ta_l^q(t)+a_k^q(t)^Tp_l^q(t) -2a_k^q(t)^Ta_l^q(t))  
\]
and
\[
    \partial_{a_k^q(t)}G = \sum_{l=1}^N K_q(z_k^q(t),z_l^q(t))(2a_l^q(t)-p_l^q(t)),  
\]
respectively.

By the PMP,
our optimal controls $\vect{a}^q(\cdot)$ and state trajectories $\vect{z}^q(\cdot)$ must also solve these Hamiltonian systems with corresponding costates $\vect{p}^q(\cdot)$ and stationarity conditions  
\begin{equation}
    \vect{a}^q(t) = \underset{\vect{a}'(t)}{\operatorname{argmax}}\hspace{0.07cm}H^q_{\vect{a}'(t)}(\vect{z}^q(t),\vect{p}^q(t)).\nonumber
\end{equation}
% for each fixed $t.$ 
Therefore, an optimal minimizer of our learning problem sets the above gradients to zero. %can be calculated by gradient descent methods using the gradients above.    

\newpage

\section{Literature Results}\label{sec:all_literature_results}

\setcounter{table}{0}
\renewcommand{\thetable}{C.1\Alph{table}}%\arabic{table}}
\begin{table}[ht!] 
	\caption{Average test RMSE $\pm$ 1 standard error.}
	\label{tab:all_literature_experiments_standard1}
	\begin{threeparttable}
	\centering\small

	\begin{adjustbox}{center}	
		\begin{tabular}{@{}>{\bfseries}l*{5}{r}@{}}
			&\multicolumn{5}{@{}c@{}}{\bfseries UCI Standard Splits (Different Splits in Gray)}\\
   		 %    \addlinespace[1pt]
    		% \cline{2-10}
    		% \addlinespace[3pt]
		    \addlinespace[3pt]
    		\cline{2-6}
    		\addlinespace[4pt]
			 % & \textbf{Concrete}& \textbf{Energy} & \textbf{Kin8nm}  & \textbf{Naval} & \textbf{Power} & \textbf{Protein}& \textbf{Wine Red} & \textbf{Yacht}  & \textbf{Year}   \\
			 %       & $N_T=1030$ & $N_T=768$ & $N_T=8192$ & $N_T=11934$ & $N_T=9568$ & $N_T=45730$ & $N_T=1599$ & $N_T=308$ & $N_T=515345$ \\
			 %       & $d_X=8$ & $d_X=8$ & $d_X=8$ & $d_X=16$ & $d_X=4$ & $d_X=9$ & $d_X=11$ & $d_X=6$ & $d_X=90$  \\
			 % Model  & $d_Y=1$ & $d_Y=1$ & $d_Y=1$ & $d_Y=1$ & $d_Y=1$ & $d_Y=1$ & $d_Y=1$ & $d_Y=1$ & $d_Y=1$  \\
            Model  & \textbf{\textit{Concrete}}& \textbf{\textit{Energy}} & \textbf{\textit{Kin8nm}}  & \textbf{\textit{Naval}} & \textbf{\textit{Power}} \\ %& \textbf{Protein}& \textbf{Wine Red} & \textbf{Yacht}  & \textbf{Year}   \\
			\midrule%[0.75pt]
			VI      &$7.13 \pm 0.12$  & $2.65 \pm 0.08$ & $0.10 \pm 0.00$  & $0.01 \pm 0.00$ & $4.33 \pm 0.04$ \\ %& $4.84 \pm 0.03$ & $0.65 \pm 0.01$ & $6.89 \pm 0.67$ & $9.03 \pm \mbox{NA}$\\
			BP      &$5.98 \pm 0.22$  & $1.10 \pm 0.07$ & $0.09 \pm 0.00$  & $0.00 \pm 0.00$ & $4.18 \pm 0.04$ \\ %& $4.54 \pm 0.03$ & $0.65 \pm 0.01$ & $1.18 \pm 0.16$ & $8.93\pm\mbox{NA}$\\
			BP-2      &$5.40 \pm 0.13$  & $0.68 \pm 0.04$ & $0.07 \pm 0.00$  & $0.00 \pm 0.00$ & $4.22 \pm 0.07$ \\ %& $4.19 \pm 0.03$ & $0.65 \pm 0.01$ & $1.54 \pm 0.19$ & $8.98\pm\mbox{NA}$\\
			BP-3      &$5.57 \pm 0.13$  & $0.63 \pm 0.03$ & $0.07 \pm 0.00$  & $0.00 \pm 0.00$ & $4.11 \pm 0.04$ \\ %& $4.01 \pm 0.03$ & $0.65 \pm 0.01$ & $1.11 \pm 0.09$ & $8.93\pm\mbox{NA}$\\
			BP-4      &$5.53 \pm 0.14$  & $0.67 \pm 0.03$ & $0.07 \pm 0.00$  & $0.00 \pm 0.00$ & $4.18 \pm 0.06$ \\ %& $3.96 \pm 0.01$ & $0.65 \pm 0.02$ & $1.27 \pm 0.13$ & $9.05\pm\mbox{NA}$\\
			PBP     &$5.67 \pm 0.09$  & $1.80 \pm 0.05$ & $0.10 \pm 0.00$  & $0.01 \pm 0.00$ & $4.12 \pm 0.03$ \\ %& $4.73 \pm 0.01$ & $0.64 \pm 0.01$ & $1.02 \pm 0.05$ & $8.88\pm\mbox{NA}$\\
			PBP-2     &$5.24 \pm 0.12$  & $0.90 \pm 0.05$ & $0.07 \pm 0.00$  & $0.00 \pm 0.00$ & $4.03 \pm 0.03$ \\ %& $4.25 \pm 0.02$ & $0.64 \pm 0.01$ & $0.85 \pm 0.05$ & $8.92\pm\mbox{NA}$\\
			PBP-3     &$5.73 \pm 0.11$  & $1.24 \pm 0.06$ & $0.07 \pm 0.00$  & $0.01 \pm 0.00$ & $4.07 \pm 0.04$ \\ %& $4.09 \pm 0.03$ & $0.64 \pm 0.01$ & $0.89 \pm 0.10$ & $8.87\pm\mbox{NA}$\\
			PBP-4     &$5.96 \pm 0.16$  & $1.18 \pm 0.06$ & $0.08 \pm 0.00$  & $0.00 \pm 0.00$ & $4.08 \pm 0.04$ \\ %& $3.97 \pm 0.04$ & $0.64 \pm 0.01$ & $1.71 \pm 0.23$ & $8.93\pm\mbox{NA}$\\
% 			\midrule[0.05pt] % Gal & Ghahramani, 2016: (standard deviations instead of standard errors)
% 			&\citet{pmlr-v48-gal16}
% 			Dropout-TS &$5.23 \pm 0.53$  & $1.66 \pm 0.19$ & $0.10 \pm 0.00$  & $0.01 \pm 0.00$ & $4.02 \pm 0.18$ & $4.36 \pm 0.04$ & $0.62 \pm 0.04$ & $1.11 \pm 0.38$ & $8.85\pm\mbox{NA}$\\
			%                                          (corrected to standard errors in Mukhoti paper):
	        Dropout-TS &$5.23 \pm 0.12$  & $1.66 \pm 0.04$ & $0.10 \pm 0.00$  & $0.01 \pm 0.00$ & $4.02 \pm 0.04$ \\ %& $4.36 \pm 0.01$ & $0.62 \pm 0.01$ & $1.11 \pm 0.09$ & $8.85\pm\mbox{NA}$\\

            VMG  &\cgr $4.70 \pm 0.14$  &\cgr $1.16 \pm 0.03$ &\cgr $0.08 \pm 0.00$  &\cgr $0.00 \pm 0.00$ &\cgr $3.88 \pm 0.03$ \\ %&\cgr $4.14 \pm 0.01$ &\cgr $0.61 \pm 0.01$ &\cgr $0.77 \pm 0.06$ &\cgr $8.78\pm\mbox{NA}$\\

            % \midrule[0.05pt]% Ghosh, 2017:    (different dataset splits)
            % \citet{2017arXiv170510388G}   (standard deviations instead of standard errors)
            % HS-BNN  &$5.66 \pm 0.41$  & $1.99 \pm 0.34$ & $ 0.08 \pm 0.00$  & \resizebox{0.62in}{\height}{$\mathbf{0.00 \pm 0.00}$} & $4.03 \pm 0.15$ & $4.39 \pm 0.04$ & $0.63 \pm 0.04$ & $1.58 \pm 0.23$ & $9.26\pm\mbox{NA}$\\
		%                                          (corrected to standard errors):
            HS-BNN  &\cgr $5.66 \pm 0.09$  &\cgr $1.99 \pm 0.08$ &\cgr $ 0.08 \pm 0.00$  &\cgr $0.00 \pm 0.00$ &\cgr $4.03 \pm 0.03$ \\ %&\cgr $4.39 \pm 0.02$ &\cgr $0.63 \pm 0.01$ &\cgr $1.58 \pm 0.05$ &\cgr $9.26\pm\mbox{NA}$\\
            
            % \midrule[0.05pt]% Sun, 2017:   (different dataset splits)
            % \citet{pmlr-v54-sun17b}
            PBP-MV  &\cgr $5.08 \pm 0.14$  &\cgr $0.45 \pm 0.01$ &\cgr $0.07 \pm 0.00$  &\cgr $0.00 \pm 0.00$ &\cgr $3.91 \pm 0.04$ \\ %&\cgr $3.94 \pm 0.02$ &\cgr $0.64 \pm 0.01$ &\cgr $0.81 \pm 0.06$ &\cgr $8.72\pm\mbox{NA}$\\

        % \midrule[0.05pt]% Mukhoti et al., 2018:   
        % \citet{https://doi.org/10.48550/arxiv.1811.09385}
        % Dropout-TS  &$5.23 \pm 0.12$  & $1.66 \pm 0.04$ & $0.10 \pm 0.00$  & $0.01 \pm 0.00$ & $4.02 \pm 0.04$ & $4.36 \pm 0.01$ & $0.62 \pm 0.01$ & $1.11 \pm 0.09$ & $--$\\
		Dropout-C   &$4.93 \pm 0.14$  & $1.08 \pm 0.03$ & $0.09 \pm 0.00$  & $0.00 \pm 0.00$ & $4.00 \pm 0.04$ \\ %& $4.27 \pm 0.01$ & $0.61 \pm 0.01$ & $0.70 \pm 0.05$ & $--$\\
		Dropout-G  &$4.82 \pm 0.16$  & $0.54 \pm 0.06$ & $0.08 \pm 0.00$  & $0.00 \pm 0.00$ & $4.01 \pm 0.04$ \\ %& $4.27 \pm 0.02$ & $0.62 \pm 0.01$ & $0.67 \pm 0.05$ & $--$\\
% 		VMG         &$4.89 \pm 0.12$  & $0.54 \pm 0.02$ & $0.08 \pm 0.00$  & \resizebox{0.62in}{\height}{$\mathbf{0.00 \pm 0.00}$} & $4.04 \pm 0.04$ & $4.13 \pm 0.02$ & $0.63 \pm 0.01$ & $0.71 \pm 0.05$ & $--$\\
% 		HS-BNN      &$5.66 \pm 0.41$  & $1.99 \pm 0.34$ & $0.08 \pm 0.00$  & \resizebox{0.62in}{\height}{$\mathbf{0.00 \pm 0.00}$} & $4.03 \pm 0.15$ & $4.39 \pm 0.04$ & $0.63 \pm 0.04$ & $1.58 \pm 0.23$ & $--$\\
% 		PBP-MV      &$5.08 \pm 0.14$  & $0.45 \pm 0.01$ & $0.07 \pm 0.00$  & \resizebox{0.62in}{\height}{$\mathbf{0.00 \pm 0.00}$} & $3.91 \pm 0.04$ & $3.94 \pm 0.02$ & $0.64 \pm 0.01$ & $0.81 \pm 0.06$ & $--$\\

        % \midrule[0.05pt]% Mishkin et al., 2018:
        BBB   &$6.16 \pm 0.13$  & $0.97 \pm 0.09$ & $0.08 \pm 0.00$  & $0.00 \pm 0.00$ & $4.21 \pm 0.03$ \\ %& $--$ & $0.64 \pm 0.01$ & $1.13 \pm 0.06$ & $--$\\
		SLANG  &$5.58 \pm 0.19$  & $0.64 \pm 0.03$ & $0.08 \pm 0.00$  & $0.00 \pm 0.00$ & $4.16 \pm 0.04$ \\ %& $--$ & $0.65 \pm 0.01$ & $1.08 \pm 0.06$ & $--$\\

		MAP-1  &\cgr $5.41 \pm 0.12$  &\cgr $0.52 \pm 0.02$ &\cgr $0.08 \pm 0.00$  &\cgr $0.00 \pm 0.00$ &\cgr $4.11 \pm 0.04$ \\ %&\cgr $4.67 \pm 0.03$ &\cgr $0.64 \pm 0.01$ &\cgr $0.73 \pm 0.06$ & $--$\\
		MAP-2  &\cgr $5.13 \pm 0.12$  &\cgr $0.47 \pm 0.02$ &\cgr $0.07 \pm 0.00$  &\cgr $0.00 \pm 0.00$ &\cgr $3.99 \pm 0.03$ \\ %&\cgr $4.33 \pm 0.01$ &\cgr $0.63 \pm 0.01$ &\cgr $0.66 \pm 0.06$ & $--$\\
		MAP-1 NL   &\cgr $5.14 \pm 0.13$  &\cgr $0.44 \pm 0.01$ &\cgr $0.08 \pm 0.00$  &\cgr $0.00 \pm 0.00$ &\cgr $4.01 \pm 0.04$ \\ %&\cgr $4.56 \pm 0.01$ &\cgr $0.64 \pm 0.01$ &\cgr $0.61 \pm 0.05$ & $--$\\
		MAP-2 NL   &\cgr $5.05 \pm 0.11$  &\cgr $0.42 \pm 0.02$ &\cgr $0.07 \pm 0.00$  &\cgr $0.00 \pm 0.00$ &\cgr $3.90 \pm 0.04$ \\ %&\cgr $4.24 \pm 0.01$ &\cgr $0.63 \pm 0.01$ &\cgr $0.63 \pm 0.05$ & $--$\\
		Reg-1 NL   &\cgr $5.03 \pm 0.16$  &\cgr $0.46 \pm 0.01$ &\cgr $0.08 \pm 0.00$  &\cgr $0.00 \pm 0.00$ &\cgr $3.91 \pm 0.04$ \\ %&\cgr $4.25 \pm 0.02$ &\cgr $0.64 \pm 0.01$ &\cgr $0.64 \pm 0.04$ & $--$\\
		Reg-2 NL    &\cgr $4.82 \pm 0.14$  &\cgr $0.43 \pm 0.02$ &\cgr $0.07 \pm 0.00$  &\cgr $0.00 \pm 0.00$ &\cgr $3.74 \pm 0.04$ \\ %&\cgr $3.94 \pm 0.02$ &\cgr $0.63 \pm 0.01$ &\cgr $0.58 \pm 0.06$ & $--$\\
		BN(ML)-1 NL &\cgr $5.08 \pm 0.13$  &\cgr $0.46 \pm 0.01$ &\cgr $0.08 \pm 0.00$  &\cgr $0.00 \pm 0.00$ &\cgr $3.94 \pm 0.04$ \\ %&\cgr $4.24 \pm 0.01$ &\cgr $0.63 \pm 0.01$ &\cgr $0.79 \pm 0.06$ & $--$\\
		BN(ML)-2 NL &\cgr $5.17 \pm 0.12$  &\cgr $0.42 \pm 0.01$ &\cgr $0.07 \pm 0.00$  &\cgr $0.00 \pm 0.00$ &\cgr $3.73 \pm 0.04$ \\ %&\cgr $3.94 \pm 0.02$ &\cgr $0.63 \pm 0.01$ &\cgr $0.55 \pm 0.05$ & $--$\\
		BN(BO)-1 NL &\cgr $4.96 \pm 0.15$  &\cgr $0.48 \pm 0.01$ &\cgr $0.08 \pm 0.00$  &\cgr $0.00 \pm 0.00$ &\cgr $3.94 \pm 0.04$ \\ %&\cgr $4.25 \pm 0.01$ &\cgr $0.63 \pm 0.01$ &\cgr $0.77 \pm 0.06$ & $--$\\
		BN(BO)-2 NL &\cgr $4.78 \pm 0.19$  &\cgr $0.40 \pm 0.01$ &\cgr $0.07 \pm 0.00$  &\cgr $0.00 \pm 0.00$ &\cgr $3.70 \pm 0.04$ \\ %&\cgr $3.88 \pm 0.02$ &\cgr $0.63 \pm 0.01$ &\cgr $0.66 \pm 0.06$ & $--$\\

        % \midrule[0.05pt]% Antoran et al, 2020: (standard deviations instead of standard errors)
        % \citet{NEURIPS2020_781877bd}
        			%Concrete                   Energy                  Kin8nm                   Naval                  Power                   Protein                 Wine                    Yacht  
%DUN:       4.613 +/- 0.607 (0.14)  0.612 +/- 0.157 (0.04)  0.076 +/- 0.005 (0.00)  0.003 +/- 0.002 (0.00)  3.573 +/- 0.254 (0.06)  3.402 +/- 0.058 (0.03)  0.659 +/- 0.061 (0.01)  2.514 +/- 1.985 (0.44)
%DUNMLP:    4.571 +/- 0.703 (0.16)  0.948 +/- 0.474 (0.11)  0.077 +/- 0.006 (0.00)  0.002 +/- 0.001 (0.00)  3.671 +/- 0.247 (0.06)  3.412 +/- 0.076 (0.03)  0.629 +/- 0.047 (0.01)  2.465 +/- 0.841 (0.19)
%Dropout:   4.610 +/- 0.572 (0.13)  0.571 +/- 0.204 (0.05)  0.070 +/- 0.005 (0.00)  0.001 +/- 0.001 (0.00)  3.823 +/- 0.350 (0.08)  3.425 +/- 0.070 (0.03)  0.642 +/- 0.049 (0.01)  0.876 +/- 0.411 (0.09)
%Ensemble:  4.552 +/- 0.582 (0.13)  0.507 +/- 0.110 (0.02)  0.304 +/- 0.991 (0.22)  0.001 +/- 0.000 (0.00)  3.444 +/- 0.238 (0.05)  3.260 +/- 0.074 (0.03)  1.934 +/- 5.708 (1.28)  1.429 +/- 0.483 (0.11) 
%MFVI:      5.894 +/- 0.742 (0.17)  1.686 +/- 1.016 (0.23)  0.084 +/- 0.007 (0.00)  0.005 +/- 0.005 (0.00)  4.286 +/- 0.179 (0.04)  4.511 +/- 0.145 (0.06)  0.660 +/- 0.040 (0.01)  3.419 +/- 7.333 (1.64)
%SGD:       4.983 +/- 0.914 (0.20)  0.797 +/- 0.283 (0.06)  0.202 +/- 0.544 (0.12)  0.002 +/- 0.001 (0.00)  3.697 +/- 0.272 (0.06)  3.589 +/- 0.174 (0.08)  0.652 +/- 0.054 (0.01)  2.352 +/- 0.905 (0.20)
			%                                          (corrected to standard errors):
        DUN         &$4.61 \pm 0.14$  & $0.61 \pm 0.04$ & $0.08 \pm 0.00$  & ${0.00 \pm 0.00}$ & $3.57 \pm 0.06$ \\ %& $3.40 \pm 0.03$ & $0.66 \pm 0.01$ & $2.51 \pm 0.44$ & $--$\\
        DUN (MLP)   &$4.57 \pm 0.16$  & $0.95 \pm 0.11$ & $0.08 \pm 0.00$  & ${0.00 \pm 0.00}$ & $3.67 \pm 0.06$ \\ %& $3.41 \pm 0.03$ & $0.63 \pm 0.01$ & $2.47 \pm 0.19$ & $--$\\
        Dropout     &$4.61 \pm 0.13$  & $0.57 \pm 0.05$ & $0.07 \pm 0.00$  & ${0.00 \pm 0.00}$ & $3.82 \pm 0.08$ \\ %& $3.43 \pm 0.03$ & $0.64 \pm 0.01$ & $0.88 \pm 0.09$ & $--$\\
        Ensemble    &$4.55 \pm 0.13$  & $0.51 \pm 0.02$ & $0.30 \pm 0.22$  & ${0.00 \pm 0.00}$ & $3.44 \pm 0.05$ \\ %& $3.26 \pm 0.03$ & $1.93 \pm 1.28$ & $1.43 \pm 0.11$ & $--$\\
        MFVI        &$5.89 \pm 0.17$  & $1.69 \pm 0.23$ & $0.08 \pm 0.00$  & ${0.01 \pm 0.00}$ & $4.29 \pm 0.04$ \\ %& $4.51 \pm 0.06$ & $0.66 \pm 0.01$ & $3.42 \pm 1.64$ & $--$\\
        SGD         &$4.98 \pm 0.20$  & $0.80 \pm 0.06$ & $0.20 \pm 0.12$  & ${0.00 \pm 0.00}$ & $3.70 \pm 0.06$ \\ %& $3.59 \pm 0.08$ & $0.65 \pm 0.01$ & $2.35 \pm 0.20$ & $--$\\

        $\mathbf{\mathcal{L}_{\beta-NLL}(\beta=0)}$     & $6.08 \pm 0.15$ &\cgr $2.25 \pm 0.08$ & $0.09 \pm 0.00$   &\cgr $0.00 \pm 0.00$ & $4.06 \pm 0.04$ \\ %& $4.49 \pm 0.05$ & $0.64 \pm 0.01$   & $1.22 \pm 0.11$ &  $--$\\
        $\mathbf{\mathcal{L}_{\beta-NLL}(\beta=0.25)}$  & $5.79 \pm 0.17$ &\cgr $1.81 \pm 0.07$ & $0.08 \pm 0.00$   &\cgr $0.00 \pm 0.00$ & $4.04 \pm 0.04$ \\ %& $4.35 \pm 0.02$ & $0.64 \pm 0.01$   & $1.73 \pm 0.22$ &  $--$\\
        $\mathbf{\mathcal{L}_{\beta-NLL}(\beta=0.5)}$   & $5.61 \pm 0.15$ &\cgr $1.12 \pm 0.06$ & $0.08 \pm 0.00$   &\cgr $0.00 \pm 0.00$ & $4.04 \pm 0.04$ \\ %& $4.31 \pm 0.01$ & $0.64 \pm 0.01$   & $2.35 \pm 0.32$ &  $--$\\
        $\mathbf{\mathcal{L}_{\beta-NLL}(\beta=0.75)}$  & $5.67 \pm 0.16$ &\cgr $1.31 \pm 0.10$ & $0.08 \pm 0.00$   &\cgr $0.00 \pm 0.00$ & $4.04 \pm 0.03$ \\ %& $4.28 \pm 0.01$ & $0.64 \pm 0.01$   & $1.97 \pm 0.23$ &  $--$\\
        $\mathbf{\mathcal{L}_{\beta-NLL}(\beta=1.0)}$   & $5.55 \pm 0.17$ &\cgr $1.54 \pm 0.12$ & $0.08 \pm 0.00$   &\cgr $0.00 \pm 0.00$ & $4.06 \pm 0.04$ \\ %& $4.31 \pm 0.02$ & $0.64 \pm 0.01$   & $2.08 \pm 0.25$ &  $--$\\
        $\mathbf{\mathcal{L}_{MM}}$                     & $6.28 \pm 0.18$ &\cgr $2.19 \pm 0.06$ & $0.08 \pm 0.00$   &\cgr $0.00 \pm 0.00$ & $4.07 \pm 0.04$ \\ %& $4.32 \pm 0.03$ & $0.65 \pm 0.01$   & $3.02 \pm 0.31$ &  $--$\\
        $\mathbf{\mathcal{L}_{MSE}}$                    & $4.96 \pm 0.14$ &\cgr $0.92 \pm 0.02$ & $0.08 \pm 0.00$   &\cgr $0.00 \pm 0.00$ & $4.01 \pm 0.04$ \\ %& $4.28 \pm 0.03$ & $0.63 \pm 0.01$   & $0.78 \pm 0.06$ &  $--$\\
        Student-t                                       & $5.82 \pm 0.13$ &\cgr $2.26 \pm 0.08$ & $0.09 \pm 0.00$   &\cgr $0.00 \pm 0.00$ & $4.02 \pm 0.04$ \\ %& $4.76 \pm 0.11$ & $0.64 \pm 0.01$   & $1.34 \pm 0.14$ &  $--$\\
        xVAMP                                           & $5.44 \pm 0.14$ &\cgr $1.87 \pm 0.07$ & $0.08 \pm 0.00$   &\cgr $0.00 \pm 0.00$ & $4.03 \pm 0.04$ \\ %& $4.38 \pm 0.02$ & $0.64 \pm 0.01$   & $0.99 \pm 0.10$ &  $--$\\
        xVAMP*                                          & $5.35 \pm 0.16$ &\cgr $2.00 \pm 0.06$ & $0.08 \pm 0.00$   &\cgr $0.00 \pm 0.00$ & $4.03 \pm 0.04$ \\ %& $4.31 \pm 0.01$ & $0.63 \pm 0.01$   & $1.13 \pm 0.15$ &  $--$\\
        VBEM                                            & $5.21 \pm 0.13$ &\cgr $1.29 \pm 0.07$ & $0.08 \pm 0.00$   &\cgr $0.00 \pm 0.00$ & $4.09 \pm 0.03$ \\ %& $4.31 \pm 0.00$ & $0.64 \pm 0.01$   & $1.66 \pm 0.19$ &  $--$\\
        VBEM*                                           & $5.17 \pm 0.13$ &\cgr $1.08 \pm 0.04$ & $0.08 \pm 0.00$   &\cgr $0.00 \pm 0.00$ & $4.02 \pm 0.04$ \\ %& $4.35 \pm 0.04$ & $0.63 \pm 0.01$   & $0.65 \pm 0.04$ &  $--$\\        
			\bottomrule
		\end{tabular}
		\end{adjustbox}
% 		\vspace{0.3cm}
        % \begin{tablenotes}\footnotesize
        % \item[1]Different UCI standard splits shaded gray.  
        % \end{tablenotes}
    \end{threeparttable}
\end{table}

%%%%%%%%%%%%%%%%%%%%%%%  splitting table in two

\clearpage 

\begin{table}[ht!] 
	\caption{Average test RMSE $\pm$ 1 standard error.}
	\label{tab:all_literature_experiments_standard2}
	\begin{threeparttable}
	\centering\small

	\begin{adjustbox}{center}	
		\begin{tabular}{@{}>{\bfseries}l*{5}{r}@{}}
			&\multicolumn{5}{@{}c@{}}{\bfseries UCI Standard Splits (Different Splits in Gray)}\\
   		 %    \addlinespace[1pt]
    		% \cline{2-10}
    		% \addlinespace[3pt]
		    \addlinespace[3pt]
    		\cline{2-6}
    		\addlinespace[4pt]
			 % & \textbf{Concrete}& \textbf{Energy} & \textbf{Kin8nm}  & \textbf{Naval} & \textbf{Power} & \textbf{Protein}& \textbf{Wine Red} & \textbf{Yacht}  & \textbf{Year}   \\
			 %       & $N_T=1030$ & $N_T=768$ & $N_T=8192$ & $N_T=11934$ & $N_T=9568$ & $N_T=45730$ & $N_T=1599$ & $N_T=308$ & $N_T=515345$ \\
			 %       & $d_X=8$ & $d_X=8$ & $d_X=8$ & $d_X=16$ & $d_X=4$ & $d_X=9$ & $d_X=11$ & $d_X=6$ & $d_X=90$  \\
			 % Model  & $d_Y=1$ & $d_Y=1$ & $d_Y=1$ & $d_Y=1$ & $d_Y=1$ & $d_Y=1$ & $d_Y=1$ & $d_Y=1$ & $d_Y=1$  \\
            Model  && \textbf{\textit{Protein}}& \textbf{\textit{Wine Red}} & \textbf{\textit{Yacht}}  & \textbf{\textit{Year}}   \\
			\midrule%[0.75pt]
			VI      && $4.84 \pm 0.03$ & $0.65 \pm 0.01$ & $6.89 \pm 0.67$ & $9.03 \pm \mbox{NA}$\\
			BP      && $4.54 \pm 0.03$ & $0.65 \pm 0.01$ & $1.18 \pm 0.16$ & $8.93\pm\mbox{NA}$\\
			BP-2    && $4.19 \pm 0.03$ & $0.65 \pm 0.01$ & $1.54 \pm 0.19$ & $8.98\pm\mbox{NA}$\\
			BP-3    && $4.01 \pm 0.03$ & $0.65 \pm 0.01$ & $1.11 \pm 0.09$ & $8.93\pm\mbox{NA}$\\
			BP-4    && $3.96 \pm 0.01$ & $0.65 \pm 0.02$ & $1.27 \pm 0.13$ & $9.05\pm\mbox{NA}$\\
			PBP     && $4.73 \pm 0.01$ & $0.64 \pm 0.01$ & $1.02 \pm 0.05$ & $8.88\pm\mbox{NA}$\\
			PBP-2   && $4.25 \pm 0.02$ & $0.64 \pm 0.01$ & $0.85 \pm 0.05$ & $8.92\pm\mbox{NA}$\\
			PBP-3   && $4.09 \pm 0.03$ & $0.64 \pm 0.01$ & $0.89 \pm 0.10$ & $8.87\pm\mbox{NA}$\\
			PBP-4   && $3.97 \pm 0.04$ & $0.64 \pm 0.01$ & $1.71 \pm 0.23$ & $8.93\pm\mbox{NA}$\\
	        Dropout-TS && $4.36 \pm 0.01$ & $0.62 \pm 0.01$ & $1.11 \pm 0.09$ & $8.85\pm\mbox{NA}$\\

            VMG  &&\cgr $4.14 \pm 0.01$ &\cgr $0.61 \pm 0.01$ &\cgr $0.77 \pm 0.06$ &\cgr $8.78\pm\mbox{NA}$\\
            HS-BNN  &&\cgr $4.39 \pm 0.02$ &\cgr $0.63 \pm 0.01$ &\cgr $1.58 \pm 0.05$ &\cgr $9.26\pm\mbox{NA}$\\
            PBP-MV  &&\cgr $3.94 \pm 0.02$ &\cgr $0.64 \pm 0.01$ &\cgr $0.81 \pm 0.06$ &\cgr $8.72\pm\mbox{NA}$\\
		Dropout-C  && $4.27 \pm 0.01$ & $0.61 \pm 0.01$ & $0.70 \pm 0.05$ & $--$\\
		Dropout-G  && $4.27 \pm 0.02$ & $0.62 \pm 0.01$ & $0.67 \pm 0.05$ & $--$\\
        BBB   && $--$ & $0.64 \pm 0.01$ & $1.13 \pm 0.06$ & $--$\\
		SLANG && $--$ & $0.65 \pm 0.01$ & $1.08 \pm 0.06$ & $--$\\
		MAP-1  &&\cgr $4.67 \pm 0.03$ &\cgr $0.64 \pm 0.01$ &\cgr $0.73 \pm 0.06$ & $--$\\
		MAP-2  &&\cgr $4.33 \pm 0.01$ &\cgr $0.63 \pm 0.01$ &\cgr $0.66 \pm 0.06$ & $--$\\
		MAP-1 NL   &&\cgr $4.56 \pm 0.01$ &\cgr $0.64 \pm 0.01$ &\cgr $0.61 \pm 0.05$ & $--$\\
		MAP-2 NL   &&\cgr $4.24 \pm 0.01$ &\cgr $0.63 \pm 0.01$ &\cgr $0.63 \pm 0.05$ & $--$\\
		Reg-1 NL   &&\cgr $4.25 \pm 0.02$ &\cgr $0.64 \pm 0.01$ &\cgr $0.64 \pm 0.04$ & $--$\\
		Reg-2 NL   &&\cgr $3.94 \pm 0.02$ &\cgr $0.63 \pm 0.01$ &\cgr $0.58 \pm 0.06$ & $--$\\
		BN(ML)-1 NL &&\cgr $4.24 \pm 0.01$ &\cgr $0.63 \pm 0.01$ &\cgr $0.79 \pm 0.06$ & $--$\\
		BN(ML)-2 NL &&\cgr $3.94 \pm 0.02$ &\cgr $0.63 \pm 0.01$ &\cgr $0.55 \pm 0.05$ & $--$\\
		BN(BO)-1 NL &&\cgr $4.25 \pm 0.01$ &\cgr $0.63 \pm 0.01$ &\cgr $0.77 \pm 0.06$ & $--$\\
		BN(BO)-2 NL &&\cgr $3.88 \pm 0.02$ &\cgr $0.63 \pm 0.01$ &\cgr $0.66 \pm 0.06$ & $--$\\

        			%Concrete                   Energy                  Kin8nm                   Naval                  Power                   Protein                 Wine                    Yacht  
%DUN:       4.613 +/- 0.607 (0.14)  0.612 +/- 0.157 (0.04)  0.076 +/- 0.005 (0.00)  0.003 +/- 0.002 (0.00)  3.573 +/- 0.254 (0.06)  3.402 +/- 0.058 (0.03)  0.659 +/- 0.061 (0.01)  2.514 +/- 1.985 (0.44)
%DUNMLP:    4.571 +/- 0.703 (0.16)  0.948 +/- 0.474 (0.11)  0.077 +/- 0.006 (0.00)  0.002 +/- 0.001 (0.00)  3.671 +/- 0.247 (0.06)  3.412 +/- 0.076 (0.03)  0.629 +/- 0.047 (0.01)  2.465 +/- 0.841 (0.19)
%Dropout:   4.610 +/- 0.572 (0.13)  0.571 +/- 0.204 (0.05)  0.070 +/- 0.005 (0.00)  0.001 +/- 0.001 (0.00)  3.823 +/- 0.350 (0.08)  3.425 +/- 0.070 (0.03)  0.642 +/- 0.049 (0.01)  0.876 +/- 0.411 (0.09)
%Ensemble:  4.552 +/- 0.582 (0.13)  0.507 +/- 0.110 (0.02)  0.304 +/- 0.991 (0.22)  0.001 +/- 0.000 (0.00)  3.444 +/- 0.238 (0.05)  3.260 +/- 0.074 (0.03)  1.934 +/- 5.708 (1.28)  1.429 +/- 0.483 (0.11) 
%MFVI:      5.894 +/- 0.742 (0.17)  1.686 +/- 1.016 (0.23)  0.084 +/- 0.007 (0.00)  0.005 +/- 0.005 (0.00)  4.286 +/- 0.179 (0.04)  4.511 +/- 0.145 (0.06)  0.660 +/- 0.040 (0.01)  3.419 +/- 7.333 (1.64)
%SGD:       4.983 +/- 0.914 (0.20)  0.797 +/- 0.283 (0.06)  0.202 +/- 0.544 (0.12)  0.002 +/- 0.001 (0.00)  3.697 +/- 0.272 (0.06)  3.589 +/- 0.174 (0.08)  0.652 +/- 0.054 (0.01)  2.352 +/- 0.905 (0.20)
			%                                          (corrected to standard errors):
        DUN         && $3.40 \pm 0.03$ & $0.66 \pm 0.01$ & $2.51 \pm 0.44$ & $--$\\
        DUN (MLP)   && $3.41 \pm 0.03$ & $0.63 \pm 0.01$ & $2.47 \pm 0.19$ & $--$\\
        Dropout     && $3.43 \pm 0.03$ & $0.64 \pm 0.01$ & $0.88 \pm 0.09$ & $--$\\
        Ensemble    && $3.26 \pm 0.03$ & $1.93 \pm 1.28$ & $1.43 \pm 0.11$ & $--$\\
        MFVI        && $4.51 \pm 0.06$ & $0.66 \pm 0.01$ & $3.42 \pm 1.64$ & $--$\\
        SGD         && $3.59 \pm 0.08$ & $0.65 \pm 0.01$ & $2.35 \pm 0.20$ & $--$\\
        $\mathbf{\mathcal{L}_{\beta-NLL}(\beta=0)}$     && $4.49 \pm 0.05$ & $0.64 \pm 0.01$   & $1.22 \pm 0.11$ &  $--$\\
        $\mathbf{\mathcal{L}_{\beta-NLL}(\beta=0.25)}$  && $4.35 \pm 0.02$ & $0.64 \pm 0.01$   & $1.73 \pm 0.22$ &  $--$\\
        $\mathbf{\mathcal{L}_{\beta-NLL}(\beta=0.5)}$   && $4.31 \pm 0.01$ & $0.64 \pm 0.01$   & $2.35 \pm 0.32$ &  $--$\\
        $\mathbf{\mathcal{L}_{\beta-NLL}(\beta=0.75)}$  && $4.28 \pm 0.01$ & $0.64 \pm 0.01$   & $1.97 \pm 0.23$ &  $--$\\
        $\mathbf{\mathcal{L}_{\beta-NLL}(\beta=1.0)}$   && $4.31 \pm 0.02$ & $0.64 \pm 0.01$   & $2.08 \pm 0.25$ &  $--$\\
        $\mathbf{\mathcal{L}_{MM}}$                     && $4.32 \pm 0.03$ & $0.65 \pm 0.01$   & $3.02 \pm 0.31$ &  $--$\\
        $\mathbf{\mathcal{L}_{MSE}}$                    && $4.28 \pm 0.03$ & $0.63 \pm 0.01$   & $0.78 \pm 0.06$ &  $--$\\
        Student-t                                       && $4.76 \pm 0.11$ & $0.64 \pm 0.01$   & $1.34 \pm 0.14$ &  $--$\\
        xVAMP                                           && $4.38 \pm 0.02$ & $0.64 \pm 0.01$   & $0.99 \pm 0.10$ &  $--$\\
        xVAMP*                                          && $4.31 \pm 0.01$ & $0.63 \pm 0.01$   & $1.13 \pm 0.15$ &  $--$\\
        VBEM                                            && $4.31 \pm 0.00$ & $0.64 \pm 0.01$   & $1.66 \pm 0.19$ &  $--$\\
        VBEM*                                           && $4.35 \pm 0.04$ & $0.63 \pm 0.01$   & $0.65 \pm 0.04$ &  $--$\\        
			\bottomrule
		\end{tabular}
		\end{adjustbox}
% 		\vspace{0.3cm}
        % \begin{tablenotes}\footnotesize
        % \item[1]Different UCI standard splits shaded gray.  
        % \end{tablenotes}
    \end{threeparttable}
\end{table}

\clearpage 

\newpage

\setcounter{table}{0}
\renewcommand{\thetable}{C.2\Alph{table}}%\arabic{table}}
\begin{table}[ht!] 
	\caption{Average test RMSE $\pm$ 1 standard error.}
	\label{tab:all_literature_experiments_gap1}
	\begin{threeparttable}
	\centering\small

		\begin{adjustbox}{center}	
		\begin{tabular}{@{}>{\bfseries}l*{5}{r}@{}}
			&\multicolumn{5}{@{}c@{}}{\bfseries UCI Gap Splits}\\
   		 %    \addlinespace[1pt]
    		% \cline{2-9}
    		% \addlinespace[3pt]
            \addlinespace[3pt]
    		\cline{2-6}
    		\addlinespace[4pt]
			 % & \textbf{Concrete}& \textbf{Energy} & \textbf{Kin8nm}  & \textbf{Naval} & \textbf{Power} & \textbf{Protein}& \textbf{Wine Red} & \textbf{Yacht}   \\
			 %       & $N_T=1030$ & $N_T=768$ & $N_T=8192$ & $N_T=11934$ & $N_T=9568$ & $N_T=45730$ & $N_T=1599$ & $N_T=308$ \\
			 %       & $d_X=8$ & $d_X=8$ & $d_X=8$ & $d_X=16$ & $d_X=4$ & $d_X=9$ & $d_X=11$ & $d_X=6$  \\
			 % Model  & $d_Y=1$ & $d_Y=1$ & $d_Y=1$ & $d_Y=1$ & $d_Y=1$ & $d_Y=1$ & $d_Y=1$ & $d_Y=1$ \\
            Model  & \textbf{\textit{Concrete}}& \textbf{\textit{Energy}} & \textbf{\textit{Kin8nm}}  & \textbf{\textit{Naval}} & \textbf{\textit{Power}} \\ %& \textbf{Protein}& \textbf{Wine Red} & \textbf{Yacht}  \\
			\midrule%[0.75pt]
    MAP-1           &$7.79 \pm 0.18$    & $2.83 \pm 0.99$   & $0.09 \pm 0.01$  & $0.02 \pm 0.00$ & $4.24 \pm 0.12$ \\ %& $5.16 \pm 0.04$ & $0.63 \pm 0.01$ & $1.31 \pm 0.14$ \\
    MAP-2           &$7.78 \pm 0.23$    & $3.70 \pm 1.33$   & $0.08 \pm 0.00$  & $0.03 \pm 0.00$ & $4.33 \pm 0.18$ \\ %& $5.07 \pm 0.06$ & $0.63 \pm 0.01$ & $1.05 \pm 0.09$ \\
    MAP-1 NL        &$7.68 \pm 0.23$    & $3.09 \pm 1.17$   & $0.09 \pm 0.01$  & $0.02 \pm 0.00$ & $4.25 \pm 0.09$ \\ %& $5.13 \pm 0.05$ & $0.63 \pm 0.01$ & $1.28 \pm 0.14$ \\
    MAP-2 NL        &$7.44 \pm 0.17$    & $3.48 \pm 1.21$   & $0.07 \pm 0.00$  & $0.03 \pm 0.00$ & $4.27 \pm 0.08$ \\ %& $5.08 \pm 0.06$ & $0.63 \pm 0.01$ & $1.01 \pm 0.09$ \\ 
    Reg-1 NL        &$8.21 \pm 0.48$    & $4.24 \pm 2.11$   & $0.08 \pm 0.00$  & $0.01 \pm 0.00$ & $5.17 \pm 0.60$ \\ %& $5.23 \pm 0.12$ & $0.66 \pm 0.02$ & $1.24 \pm 0.11$ \\
    Reg-2 NL        &$8.27 \pm 0.39$    & $3.83 \pm 1.49$   & $0.07 \pm 0.00$  & $0.01 \pm 0.00$ & $5.23 \pm 0.43$ \\ %& $5.33 \pm 0.16$ & $0.64 \pm 0.01$ & $1.22 \pm 0.13$ \\
    BN(ML)-1 NL     &$7.69 \pm 0.51$    & $4.15 \pm 1.64$   & $0.09 \pm 0.00$  & $0.01 \pm 0.00$ & $4.49 \pm 0.15$ \\ %& $5.27 \pm 0.12$ & $0.63 \pm 0.01$ & $1.15 \pm 0.11$ \\ 
    BN(ML)-2 NL     &$7.33 \pm 0.36$    & $4.10 \pm 1.64$   & $0.08 \pm 0.00$  & $0.01 \pm 0.00$ & $5.17 \pm 0.28$ \\ %& $5.37 \pm 0.17$ & $0.64 \pm 0.01$ & $1.31 \pm 0.16$ \\
    BN(BO)-1 NL     &$7.74 \pm 0.31$    & $4.76 \pm 1.98$   & $0.08 \pm 0.00$  & $0.01 \pm 0.00$ & $4.66 \pm 0.21$ \\ %& $5.14 \pm 0.10$ & $0.65 \pm 0.01$ & $1.37 \pm 0.15$ \\
    BN(BO)-2 NL     &$9.20 \pm 0.55$    & $4.58 \pm 1.87$   & $0.07 \pm 0.00$  & $0.01 \pm 0.00$ & $5.27 \pm 0.36$ \\ %& $5.46 \pm 0.17$ & $0.64 \pm 0.01$ & $1.59 \pm 0.23$ \\		
		
        % \midrule[0.05pt]% Antoran et al, 2020: (standard deviations instead of standard errors)
        % \citet{NEURIPS2020_781877bd}
    			%Concrete                   Energy                  Kin8nm                   Naval                  Power                   Protein                 Wine                    Yacht  
%DUN:       7.196 +/- 0.821 (0.18)  2.938 +/- 3.017 (0.67)  0.080 +/- 0.006 (0.00)  0.022 +/- 0.014 (0.00)  4.299 +/- 0.416 (0.09)  5.206 +/- 0.780 (0.35)  0.697 +/- 0.043 (0.01)  1.851 +/- 0.750 (0.17)
%DUNMLP:    7.461 +/- 0.948 (0.21)  3.606 +/- 3.927 (0.88)  0.078 +/- 0.005 (0.00)  0.021 +/- 0.007 (0.00)  4.584 +/- 0.356 (0.08)  5.101 +/- 0.526 (0.24)  0.692 +/- 0.041 (0.01)  1.852 +/- 0.623 (0.14)
%Dropout:   7.064 +/- 0.921 (0.21)  2.874 +/- 2.254 (0.50)  0.071 +/- 0.003 (0.00)  0.034 +/- 0.018 (0.00)  4.688 +/- 0.335 (0.07)  5.133 +/- 0.636 (0.28)  0.660 +/- 0.040 (0.01)  2.290 +/- 2.108 (0.47)
%Ensemble:  6.853 +/- 0.796 (0.18)  3.364 +/- 3.696 (0.83)  1.632 +/- 4.418 (0.99)  0.018 +/- 0.009 (0.00)  4.369 +/- 0.383 (0.09)  4.801 +/- 0.599 (0.27)  0.673 +/- 0.039 (0.01)  1.841 +/- 0.836 (0.19) 
%MFVI:      7.548 +/- 0.865 (0.19)  8.614 +/- 9.390 (2.10)  0.095 +/- 0.025 (0.01)  0.033 +/- 0.041 (0.01)  4.680 +/- 0.703 (0.16)  5.115 +/- 0.298 (0.13)  0.632 +/- 0.029 (0.01)  1.836 +/- 0.712 (0.16)
%SGD:       7.367 +/- 0.866 (0.19)  3.061 +/- 2.880 (0.64)  0.085 +/- 0.007 (0.00)  0.020 +/- 0.009 (0.00)  4.621 +/- 0.339 (0.08)  5.171 +/- 0.632 (0.28)  0.731 +/- 0.070 (0.02)  2.214 +/- 0.793 (0.18)
			%                                          (corrected to standard errors):
        DUN         &$7.20 \pm 0.18$  & $2.94 \pm 0.67$     & $0.08 \pm 0.00$       & $0.02 \pm 0.00$       & $4.30 \pm 0.09$       \\ %& $5.21 \pm 0.35$       & $0.70 \pm 0.01$   & $1.85 \pm 0.17$ \\
        DUN (MLP)   &$7.46 \pm 0.21$  & $3.61 \pm 0.88$     & $0.08 \pm 0.00$       & $0.02 \pm 0.00$       & $4.58 \pm 0.08$       \\ %& $5.10 \pm 0.24$       & $0.69 \pm 0.01$   & $1.85 \pm 0.14$\\
        Dropout     &$7.06 \pm 0.21$  & $2.87 \pm 0.50$     & $0.07 \pm 0.00$       & $0.03 \pm 0.00$       & $4.69 \pm 0.07$       \\ %& $5.13 \pm 0.28$       & $0.66 \pm 0.01$   & $2.29 \pm 0.47$ \\
        Ensemble    & $6.85 \pm 0.18$  & $3.36 \pm 0.83$     & $1.63 \pm 0.99$       & $0.02 \pm 0.00$       & $4.37 \pm 0.09$       \\ %& $4.80 \pm 0.27$       & $0.67 \pm 0.01$   & $1.84 \pm 0.19$ \\
        MFVI        &$7.55 \pm 0.19$  & $8.61 \pm 2.10$     & $0.10 \pm 0.01$       & $0.03 \pm 0.01$       & $4.68 \pm 0.16$       \\ %& $5.12 \pm 0.13$       & $0.63 \pm 0.01$   & $1.84 \pm 0.16$ \\
        SGD         &$7.37 \pm 0.19$  & $3.06 \pm 0.64$     & $0.09 \pm 0.00$       & $0.02 \pm 0.00$       & $4.62 \pm 0.08$       \\ %& $5.17 \pm 0.28$       & $0.73 \pm 0.02$   & $2.21 \pm 0.18$ \\
        
			\bottomrule
		\end{tabular}
	\end{adjustbox}
    \end{threeparttable}
\end{table}

%%%%%%%%%%%%%%%%%%%%%%%  splitting table in two

\begin{table}[ht!] 
	\caption{Average test RMSE $\pm$ 1 standard error.}
	\label{tab:all_literature_experiments_gap2}
	\begin{threeparttable}
	\centering\small

		\begin{adjustbox}{center}	
		\begin{tabular}{@{}>{\bfseries}l*{3}{r}@{}}
			&\multicolumn{3}{@{}c@{}}{\bfseries UCI Gap Splits}\\
   		 %    \addlinespace[1pt]
    		% \cline{2-9}
    		% \addlinespace[3pt]
            \addlinespace[3pt]
    		\cline{2-4}
    		\addlinespace[4pt]
			 % & \textbf{Concrete}& \textbf{Energy} & \textbf{Kin8nm}  & \textbf{Naval} & \textbf{Power} & \textbf{Protein}& \textbf{Wine Red} & \textbf{Yacht}   \\
			 %       & $N_T=1030$ & $N_T=768$ & $N_T=8192$ & $N_T=11934$ & $N_T=9568$ & $N_T=45730$ & $N_T=1599$ & $N_T=308$ \\
			 %       & $d_X=8$ & $d_X=8$ & $d_X=8$ & $d_X=16$ & $d_X=4$ & $d_X=9$ & $d_X=11$ & $d_X=6$  \\
			 % Model  & $d_Y=1$ & $d_Y=1$ & $d_Y=1$ & $d_Y=1$ & $d_Y=1$ & $d_Y=1$ & $d_Y=1$ & $d_Y=1$ \\
            Model  & \textbf{\textit{Protein}}& \textbf{\textit{Wine Red}} & \textbf{\textit{Yacht}}  \\
			\midrule%[0.75pt]

    MAP-1          & $5.16 \pm 0.04$ & $0.63 \pm 0.01$ & $1.31 \pm 0.14$ \\
    MAP-2          & $5.07 \pm 0.06$ & $0.63 \pm 0.01$ & $1.05 \pm 0.09$ \\
    MAP-1 NL       & $5.13 \pm 0.05$ & $0.63 \pm 0.01$ & $1.28 \pm 0.14$ \\
    MAP-2 NL       & $5.08 \pm 0.06$ & $0.63 \pm 0.01$ & $1.01 \pm 0.09$ \\ 
    Reg-1 NL       & $5.23 \pm 0.12$ & $0.66 \pm 0.02$ & $1.24 \pm 0.11$ \\
    Reg-2 NL       & $5.33 \pm 0.16$ & $0.64 \pm 0.01$ & $1.22 \pm 0.13$ \\
    BN(ML)-1 NL    & $5.27 \pm 0.12$ & $0.63 \pm 0.01$ & $1.15 \pm 0.11$ \\ 
    BN(ML)-2 NL    & $5.37 \pm 0.17$ & $0.64 \pm 0.01$ & $1.31 \pm 0.16$ \\
    BN(BO)-1 NL    & $5.14 \pm 0.10$ & $0.65 \pm 0.01$ & $1.37 \pm 0.15$ \\
    BN(BO)-2 NL    & $5.46 \pm 0.17$ & $0.64 \pm 0.01$ & $1.59 \pm 0.23$ \\		
    			%Concrete                   Energy                  Kin8nm                   Naval                  Power                   Protein                 Wine                    Yacht  
%DUN:       7.196 +/- 0.821 (0.18)  2.938 +/- 3.017 (0.67)  0.080 +/- 0.006 (0.00)  0.022 +/- 0.014 (0.00)  4.299 +/- 0.416 (0.09)  5.206 +/- 0.780 (0.35)  0.697 +/- 0.043 (0.01)  1.851 +/- 0.750 (0.17)
%DUNMLP:    7.461 +/- 0.948 (0.21)  3.606 +/- 3.927 (0.88)  0.078 +/- 0.005 (0.00)  0.021 +/- 0.007 (0.00)  4.584 +/- 0.356 (0.08)  5.101 +/- 0.526 (0.24)  0.692 +/- 0.041 (0.01)  1.852 +/- 0.623 (0.14)
%Dropout:   7.064 +/- 0.921 (0.21)  2.874 +/- 2.254 (0.50)  0.071 +/- 0.003 (0.00)  0.034 +/- 0.018 (0.00)  4.688 +/- 0.335 (0.07)  5.133 +/- 0.636 (0.28)  0.660 +/- 0.040 (0.01)  2.290 +/- 2.108 (0.47)
%Ensemble:  6.853 +/- 0.796 (0.18)  3.364 +/- 3.696 (0.83)  1.632 +/- 4.418 (0.99)  0.018 +/- 0.009 (0.00)  4.369 +/- 0.383 (0.09)  4.801 +/- 0.599 (0.27)  0.673 +/- 0.039 (0.01)  1.841 +/- 0.836 (0.19) 
%MFVI:      7.548 +/- 0.865 (0.19)  8.614 +/- 9.390 (2.10)  0.095 +/- 0.025 (0.01)  0.033 +/- 0.041 (0.01)  4.680 +/- 0.703 (0.16)  5.115 +/- 0.298 (0.13)  0.632 +/- 0.029 (0.01)  1.836 +/- 0.712 (0.16)
%SGD:       7.367 +/- 0.866 (0.19)  3.061 +/- 2.880 (0.64)  0.085 +/- 0.007 (0.00)  0.020 +/- 0.009 (0.00)  4.621 +/- 0.339 (0.08)  5.171 +/- 0.632 (0.28)  0.731 +/- 0.070 (0.02)  2.214 +/- 0.793 (0.18)
			%                                          (corrected to standard errors):
        DUN         & $5.21 \pm 0.35$       & $0.70 \pm 0.01$   & $1.85 \pm 0.17$ \\
        DUN (MLP)   & $5.10 \pm 0.24$       & $0.69 \pm 0.01$   & $1.85 \pm 0.14$\\
        Dropout     & $5.13 \pm 0.28$       & $0.66 \pm 0.01$   & $2.29 \pm 0.47$ \\
        Ensemble    & $4.80 \pm 0.27$       & $0.67 \pm 0.01$   & $1.84 \pm 0.19$ \\
        MFVI        & $5.12 \pm 0.13$       & $0.63 \pm 0.01$   & $1.84 \pm 0.16$ \\
        SGD         & $5.17 \pm 0.28$       & $0.73 \pm 0.02$   & $2.21 \pm 0.18$ \\
        
			\bottomrule
		\end{tabular}
	\end{adjustbox}
    \end{threeparttable}
\end{table}

\clearpage 

\newpage

\end{document}